\documentclass{article}

\usepackage[dblblindworkshop, final]{neurips_2026}

\usepackage[utf8]{inputenc} 
\usepackage[T1]{fontenc}    
\usepackage{hyperref}       
\hypersetup{hidelinks}
\usepackage{url}            
\usepackage{booktabs}       
\usepackage{amsfonts}       
\usepackage{nicefrac}       
\usepackage{microtype}      
\usepackage{xcolor}         
\usepackage{graphicx}
\usepackage{subcaption}
\usepackage{multirow}
\usepackage{amsmath}
\usepackage{placeins}

\title{Overwhelmed by Choice: Studying LLM Decision Making at Scale}
\workshoptitle{TAE (Trust-AI-Eval): Can We Trust AI Evaluation?}

\author{
Yu-Chi Lin$^{1}$ \quad
Aryan Seth$^{1}$ \quad
Anshul Aravind$^{1}$ \quad
Eugene Lee$^{2}$ \\
Tanmay Parekh$^{1}$ \quad
Nanyun Peng$^{1}$ \quad
Kai-Wei Chang$^{1}$ \\
$^{1}$University of California, Los Angeles \quad
$^{2}$University of Cincinnati \\
\texttt{yclin0177@g.ucla.edu, anshularavind21@g.ucla.edu,}\\
\texttt{violetpeng@cs.ucla.edu, kwchang@cs.ucla.edu}
}

\begin{document}

\maketitle

\begin{abstract}
    Multiple-choice and candidate-selection evaluations are widely used to assess LLM reasoning and decision-making, yet most benchmarks contain relatively small candidate sets. It remains unclear whether conclusions drawn from these settings remain valid as the candidate space scales. We systematically evaluate LLMs as the number of competing candidates increases and find substantial accuracy degradation across tasks, prompting strategies, and model scales. Controlled analyses show that standard long-context retrieval explanations cannot fully account for this degradation. Instead, we identify two systematic failure patterns. First, \textit{gold-margin collapse}: the score gap between the correct answer and the strongest distractor progressively shrinks, driven primarily by weakening confidence in the correct answer. Second, earlier candidate preferences become increasingly difficult to overturn, with later candidates exerting progressively weaker influence on the final prediction. Motivated by these findings, we evaluate hierarchical partitioning and permutation-based inference, which improve accuracy by roughly 20 percentage points at $N=160$ on both HotpotQA and MIMIC. Overall, our results identify candidate-set scale as an important evaluation-protocol variable and show that strong small-option performance does not necessarily imply robust large-scale candidate comparison.
\end{abstract}

\section{Introduction}

Large language models (LLMs) are increasingly deployed in applications that require decisions over large candidate spaces, including reranking\citep{DBLP:journals/corr/abs-2308-07107}, medical coding and diagnosis \citep{johnson2016mimic,singh2025max}, taxonomy prediction, and tool selection in agentic systems \citep{DBLP:journals/corr/abs-2508-16260}. Yet most MCQ evaluations contain fewer than ten options \citep{mmlu,arc,gpqa}, potentially overlooking the challenges of scalable decision-making \citep{zhang2025rethinking}, where models must distinguish
among hundreds of competing candidates.

\begin{figure*}[t]
    \centering
    \includegraphics[width=\linewidth]{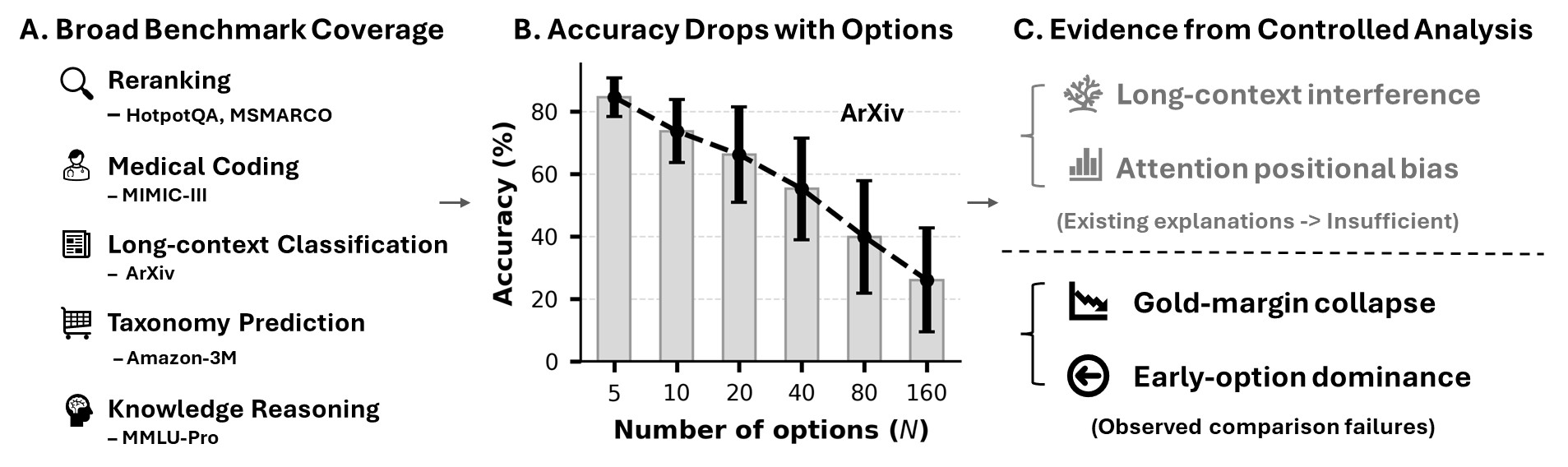}
    \caption{
    Overview of large-option degradation under scalable candidate comparison.
    Controlled analyses suggest that standard long-context explanations alone cannot fully explain the observed degradation, revealing additional comparison-level failure patterns under large candidate sets.
    }
    \label{fig:overview}
\end{figure*}

We systematically evaluate LLMs as candidate sets scale across diverse decision-making tasks (Figure~\ref{fig:overview}), including reranking, medical diagnosis, long-context classification, and information extraction \citep{thakur2021beir,johnson2016mimic,arxiv-dataset,wang-etal-2020-maven}. Across datasets, prompting strategies, and model scales, accuracy consistently declines as the number of candidates increases.

A natural hypothesis is that this degradation primarily arises from standard long-context inference failures, including noise and interference effects in long-context reasoning~\citep{hong2025contextrot,shi2023distracted} and positional biases in transformer attention~\citep{hsieh2024found,liu-etal-2024-lost,wu2025positionbias}. However, these studies largely focus on retrieval, where the primary challenge is locating relevant information within long contexts. Large-option decision-making additionally requires comparison and prioritization across competing candidates. Through controlled analysis, we find that existing long-context interference and transformer positional-bias explanations alone cannot fully account for the degradation observed under large candidate settings.

Instead, we identify two systematic comparison-level failure patterns. First, we observe \textit{gold-margin collapse}, where the score gap between the correct answer and the strongest distractor shrinks as candidate sets scale. Importantly, this collapse is driven primarily by weakening confidence in the correct answer rather than increasingly competitive distractors. Second, model predictions become increasingly dominated by earlier candidate positions under large option sets, consistent with controlled gold-insertion experiments showing that candidate influence progressively decreases with later positions. Together, these findings suggest that large-option degradation reflects not only long-context retrieval challenges, but also systematic weaknesses in large-scale candidate comparison.

Motivated by these findings, we investigate two inference-time interventions. Hierarchical partitioning decomposes large candidate spaces into smaller local comparison stages, reducing effective comparison complexity, while permutation-based inference mitigates early-option dominance by aggregating predictions across shuffled candidate orders. Despite their simplicity, both methods substantially improve robustness under large candidate settings. At \(N=160\), they improve accuracy by roughly 20 percentage points over standard single-pass inference on both HotpotQA and MIMIC.

Overall, our contributions are summarized as follows:
\begin{itemize}
    \item We systematically characterize large-scale candidate comparison failures in LLM decision-making across diverse domains.

    \item We show that large-option degradation cannot be fully explained by standard long-context retrieval failures, highlighting an important distinction between retrieval-oriented reasoning and large-scale candidate comparison.

    \item We identify two consistent failure patterns in large-scale candidate comparison: \textit{gold-margin collapse}, where confidence in the correct answer progressively weakens as candidate sets scale, and \textit{early-option dominance}, where predictions become increasingly dominated by earlier candidate evaluations.

    \item We demonstrate that restructuring strategies, including hierarchical partitioning and permutation inference, substantially improve robustness under large candidate settings.

\end{itemize}

\section{Large-Scale Candidate Comparison}

\subsection{Experimental Setup}

\paragraph{Task Setup}
We study single-answer multiple-choice decision-making, where a model
receives an instruction $I$, a query $q$, and an indexed candidate set
\[
\mathcal{O}=\{(1,o_1),(2,o_2),\ldots,(N,o_N)\},
\]
containing exactly one correct candidate. We evaluate diverse tasks
spanning reranking, scientific classification, medical diagnosis, and
long-context reasoning; dataset statistics are provided in
Appendix Table~\ref{tab:dataset_stats}.

We use publicly available datasets and pretrained models under their original research licenses and usage terms.

\paragraph{Output Generation}
We evaluate open-source models from the Qwen3, Gemma, Llama, and Phi families under varying candidate set sizes
\[
N \in \{5,10,20,40,80,160\}.
\]
Unless otherwise specified, models are evaluated under standard zero-shot prompting. We additionally explore several inference-time reasoning settings to examine whether large-option degradation persists under stronger reasoning configurations.

\paragraph{Evaluation}
We evaluate decision accuracy under varying candidate set sizes. To investigate the mechanisms underlying large-option degradation, we further analyze model confidence dynamics, candidate interaction behavior, and the potential roles of context length and attention-based positional effects under large candidate spaces.

\subsection{Accuracy Degradation at Large N}

\begin{figure*}[t]
    \centering
    \begin{subfigure}[t]{0.31\linewidth}
        \centering
        \includegraphics[width=\linewidth]{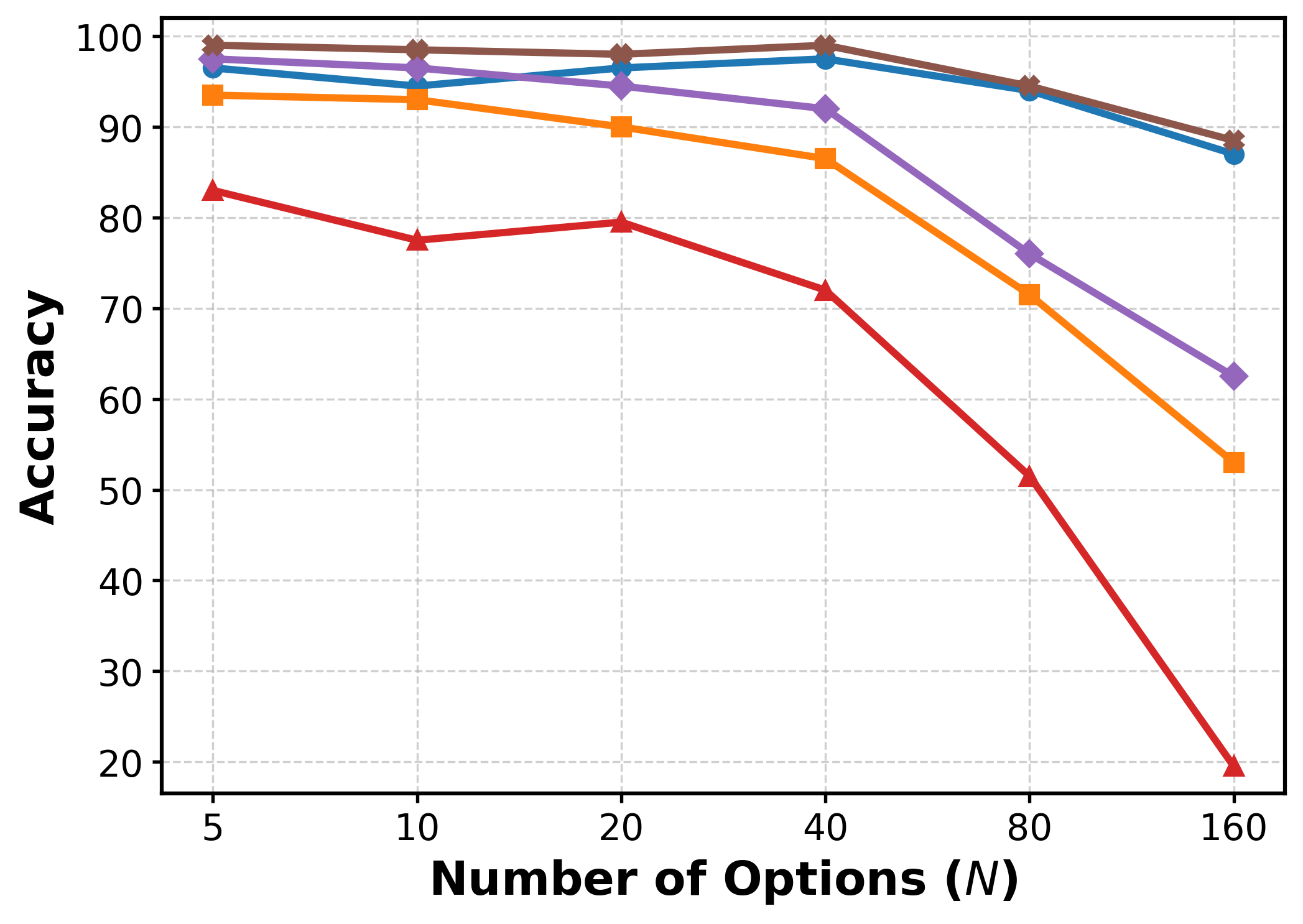}
        \caption{HotpotQA}
    \end{subfigure}
    \hfill
    \begin{subfigure}[t]{0.31\linewidth}
        \centering
        \includegraphics[width=\linewidth]{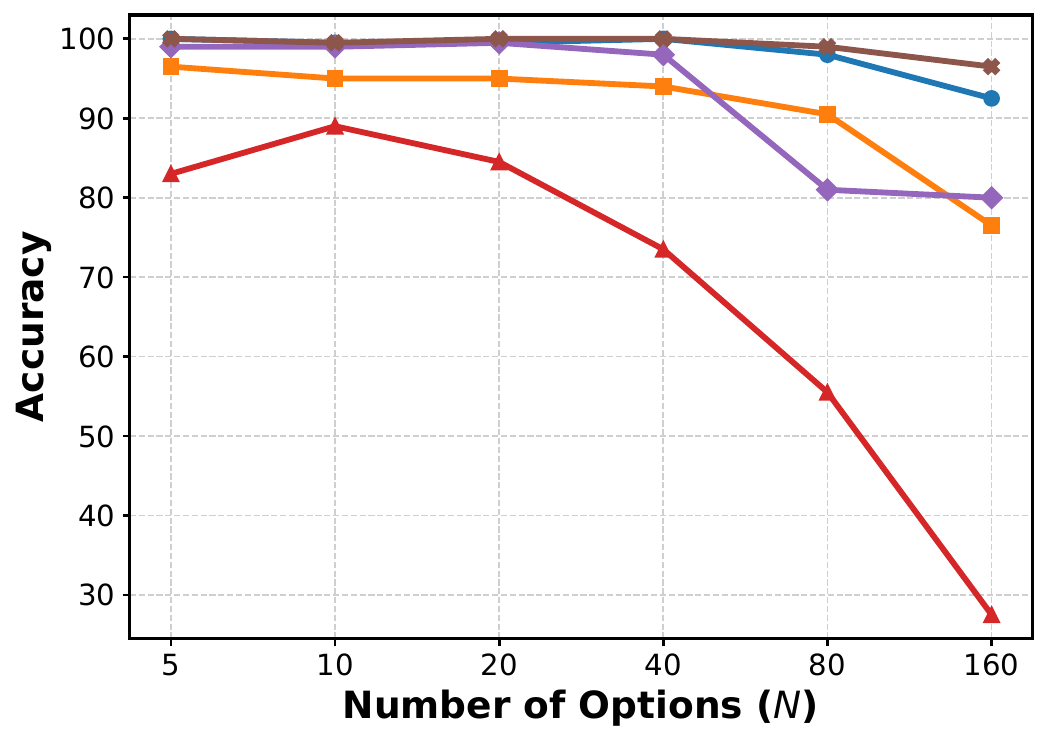}
        \caption{MSMARCO}
    \end{subfigure}
    \hfill
    \begin{subfigure}[t]{0.31\linewidth}
        \centering
        \includegraphics[width=\linewidth]{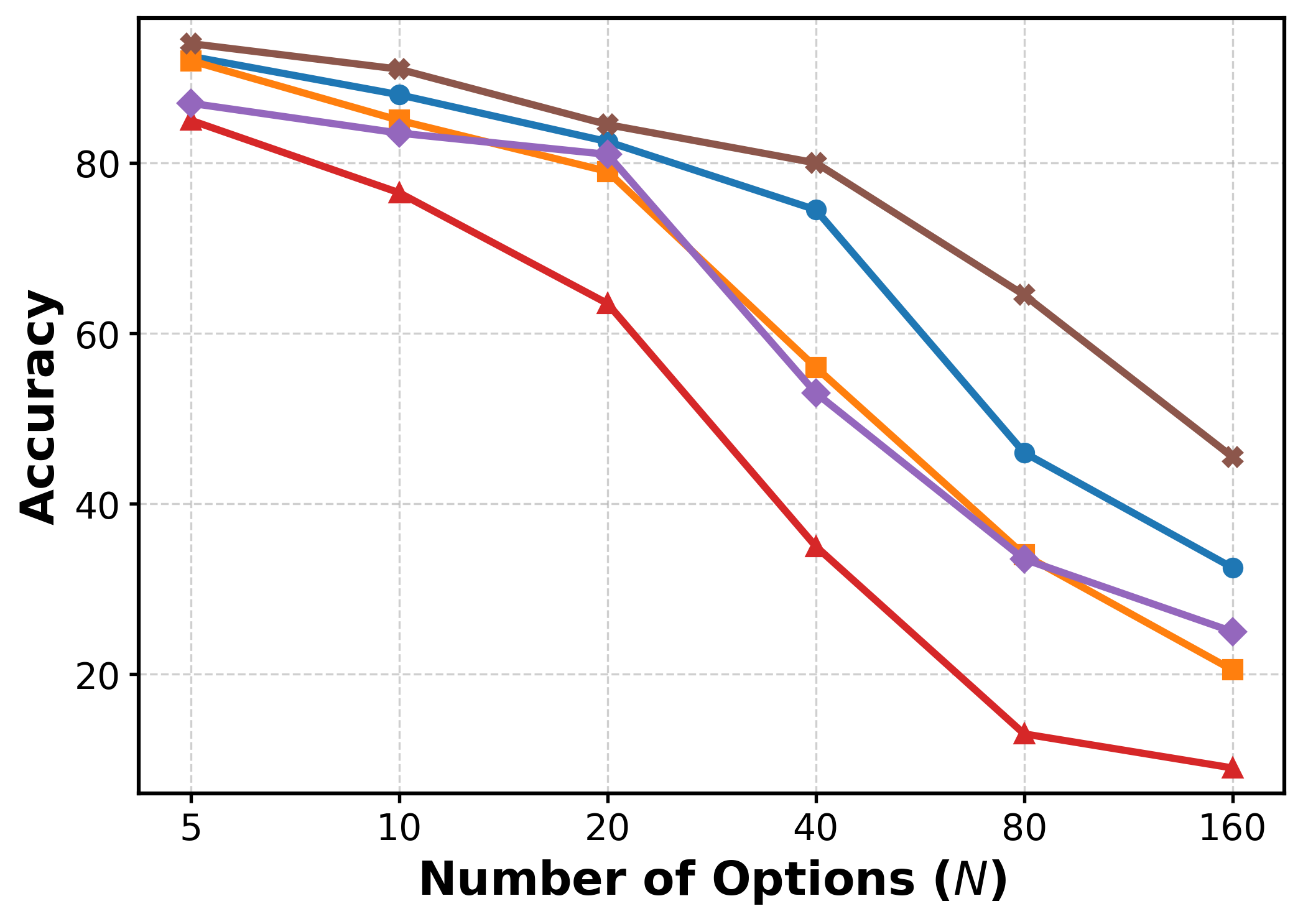}
        \caption{MIMIC}
    \end{subfigure}

    \vspace{0.35em}

    \begin{subfigure}[t]{0.31\linewidth}
        \centering
        \includegraphics[width=\linewidth]{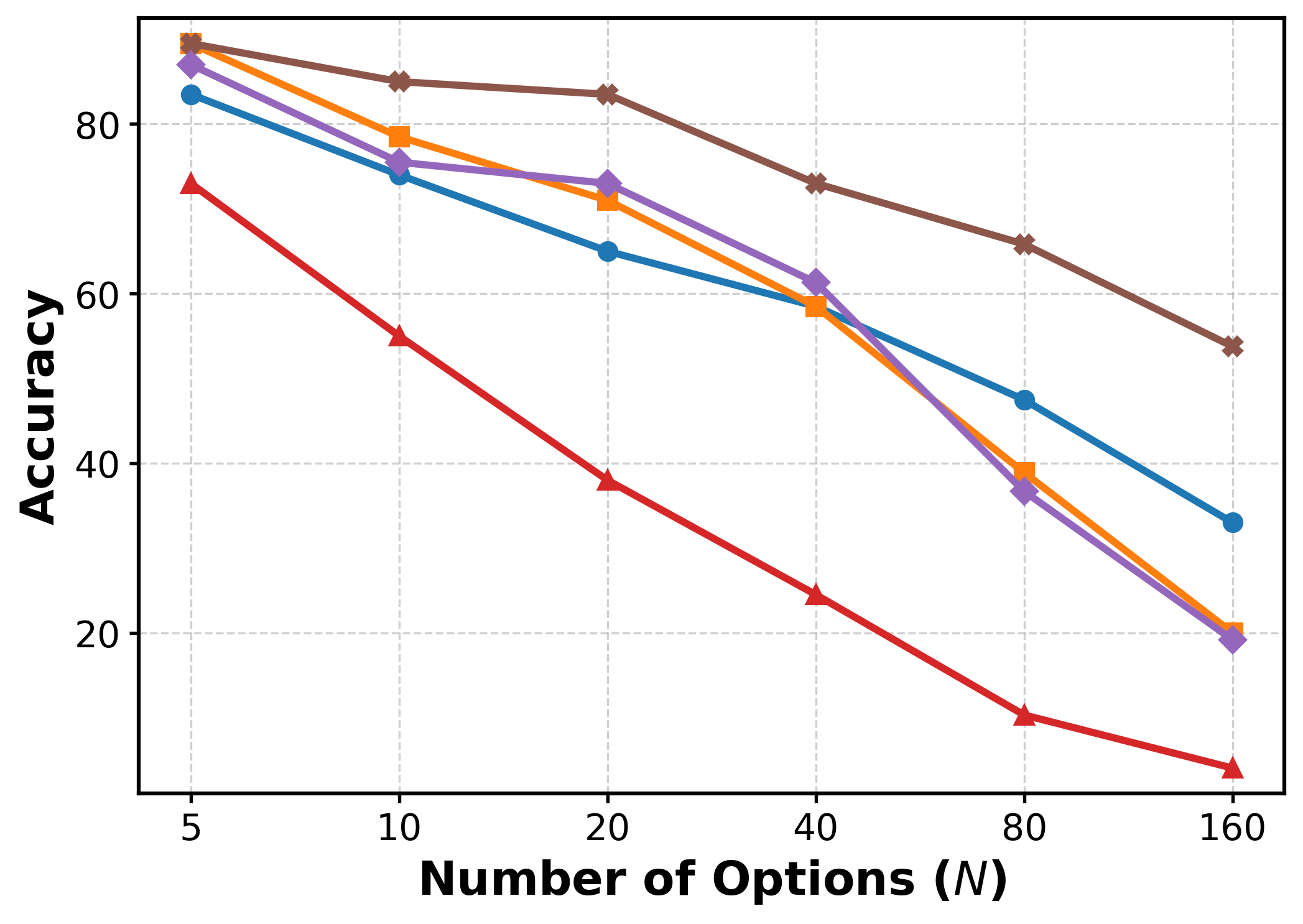}
        \caption{ArXiv}
    \end{subfigure}
    \hfill
    \begin{subfigure}[t]{0.31\linewidth}
        \centering
        \includegraphics[width=\linewidth]{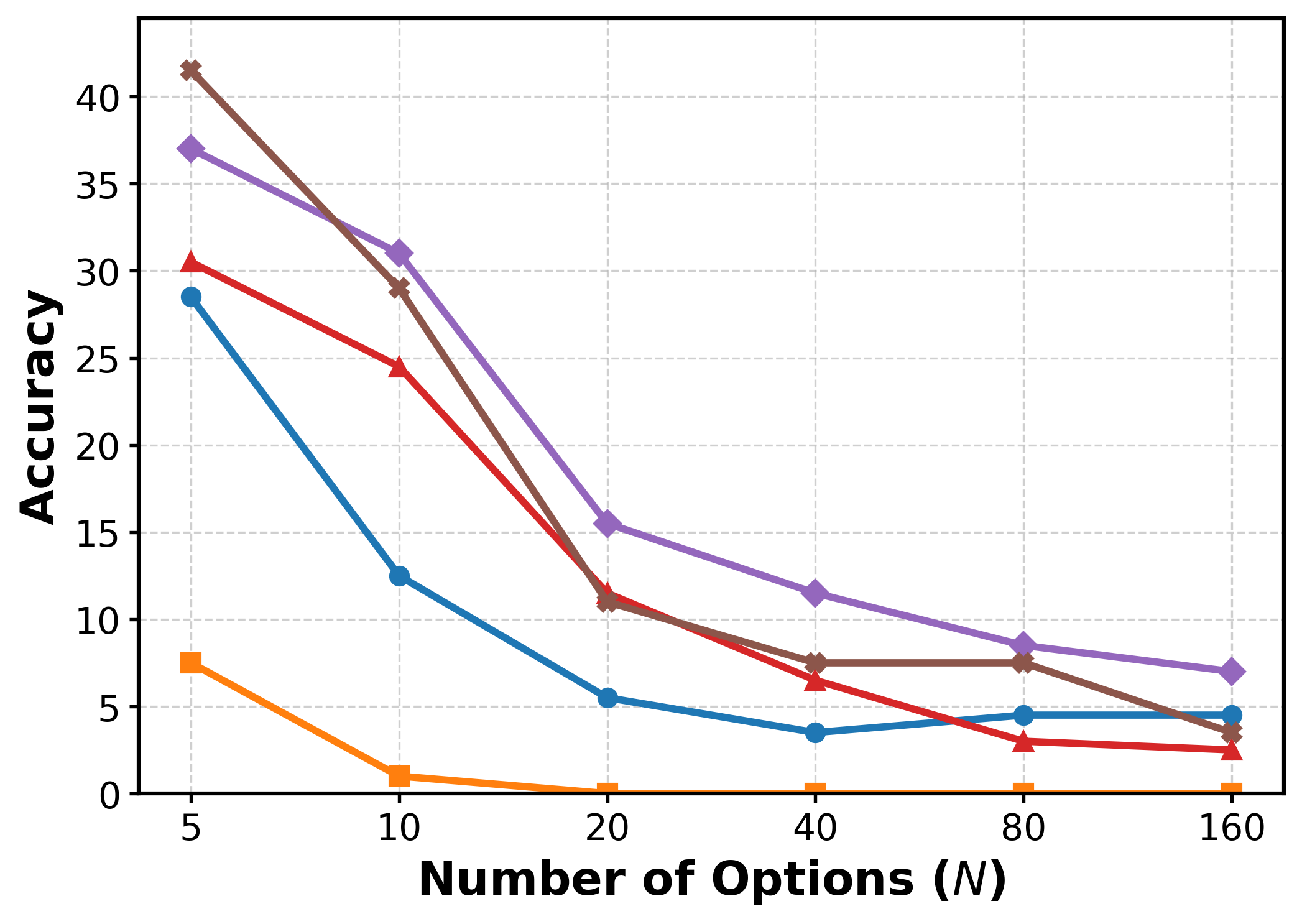}
        \caption{Amazon-3M}
    \end{subfigure}
    \hfill
    \begin{subfigure}[t]{0.31\linewidth}
        \centering
        \includegraphics[width=\linewidth]{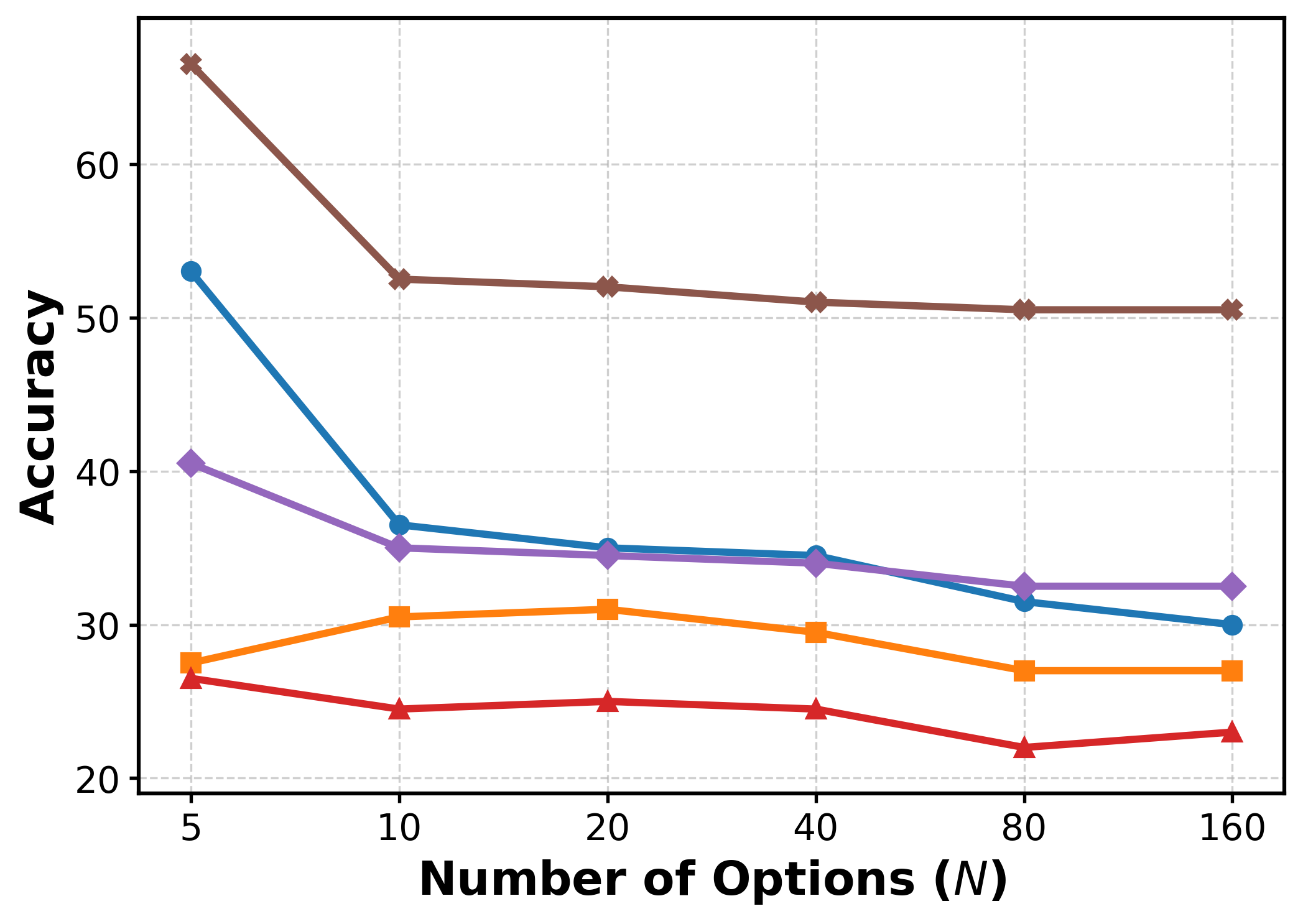}
        \caption{MMLU Pro}
    \end{subfigure}

    \vspace{0.35em}
    \includegraphics[width=0.55\linewidth]{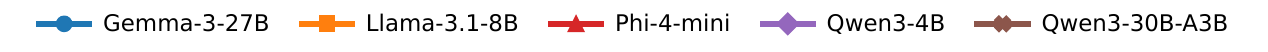}

    \caption{Accuracy degradation in the large-$N$ regime.
    As the number of options increases, accuracy consistently declines across all datasets.}
    \label{fig:accuracy_six_datasets}
\end{figure*}

\paragraph{Main Results}
Figure~\ref{fig:accuracy_six_datasets} shows that accuracy consistently decreases as the number of options grows across tasks and model families. The same trend holds across additional datasets
(Appendix~\ref{fig:accuracy_appendix_datasets}), indicating that
large-option degradation is not dataset-specific.

\paragraph{Robustness and Extended Scaling}
The degradation remains robust under bootstrap confidence intervals and 1,000-example evaluations, and persists when scaling candidate sets to $N=640$ and evaluating a strong closed-source model. Its magnitude varies across models and tasks. Full statistical and extended-scaling results are provided in Appendix~\ref{appendix:accuracy_robustness}.

\begin{figure*}[t]
    \centering

    \begin{subfigure}[t]{0.40\linewidth}
        \centering
        \includegraphics[width=\linewidth]{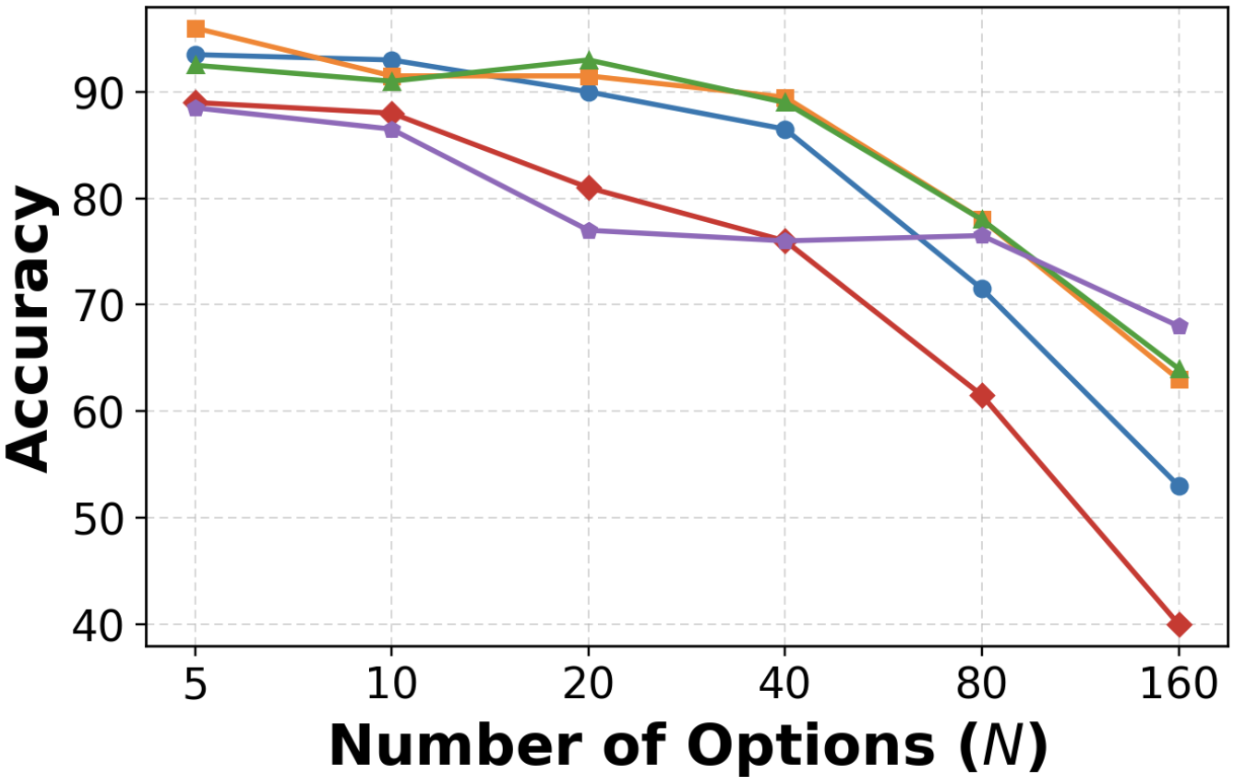}
        \caption{HotpotQA}
    \end{subfigure}
    \hspace{0.05\linewidth}
    \begin{subfigure}[t]{0.40\linewidth}
        \centering
        \includegraphics[width=\linewidth]{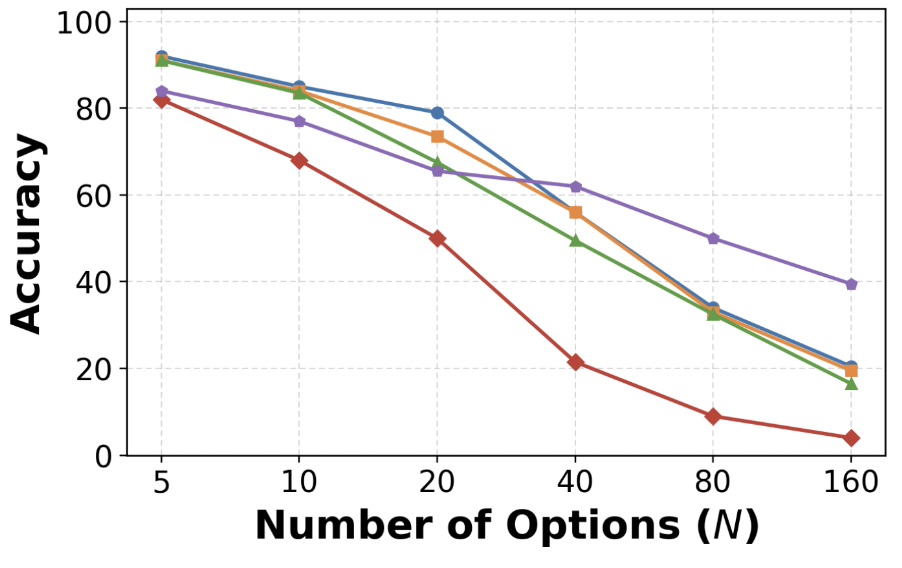}
        \caption{MIMIC}
    \end{subfigure}

    \vspace{0.15em}
    \includegraphics[width=0.5\linewidth]{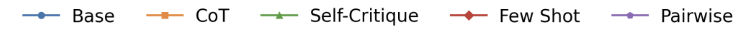}

    \caption{System-2 reasoning under large candidate sets (Llama-3.1-8B).
    All reasoning strategies degrade substantially as the number of options increases.}
    \label{fig:system2_combined}
\end{figure*}

\paragraph{System-2 Reasoning}
We evaluate chain-of-thought, self-critique, few-shot prompting, and
pairwise comparison. Although some strategies improve absolute accuracy, all continue to degrade as candidate sets grow (Figure~\ref{fig:system2_combined}), suggesting that stronger reasoning alone does not resolve large-option degradation. Additional results are provided in Appendix~\ref{appendix:system2_reasoning}.

\section{Analysis}
\label{sec:analysis}

To understand why performance degrades under large candidate sets, we analyze model confidence dynamics, candidate interaction behavior, and the potential effects of context length and attention-based positional biases during candidate comparison.

\subsection{Gold-Margin Collapse Under Large Candidate Sets}

We analyze the autoregressive sequence scores assigned to candidate
options as the number of candidates increases. Each candidate score is
the summed token-level log-probability of generating its answer index in the format \texttt{<Answer> idx </Answer>}. We define the \emph{gold margin} (GM) as the difference between the gold score and the highest-scoring distractor, and \emph{distractor mass} as the log-sum-exp of all distractor scores.

\begin{figure*}[t]
    \centering
    \begin{subfigure}[t]{0.31\linewidth}
        \centering
        \includegraphics[width=\linewidth]{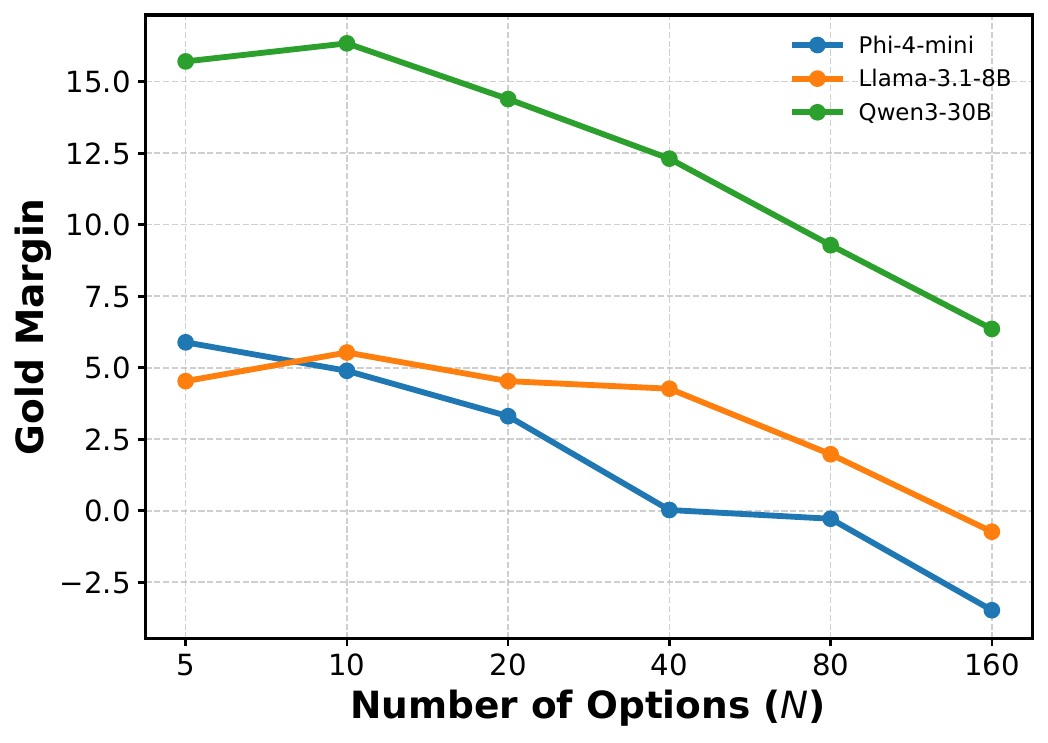}
        \caption{HotpotQA}
    \end{subfigure}
    \hfill
    \begin{subfigure}[t]{0.31\linewidth}
        \centering
        \includegraphics[width=\linewidth]{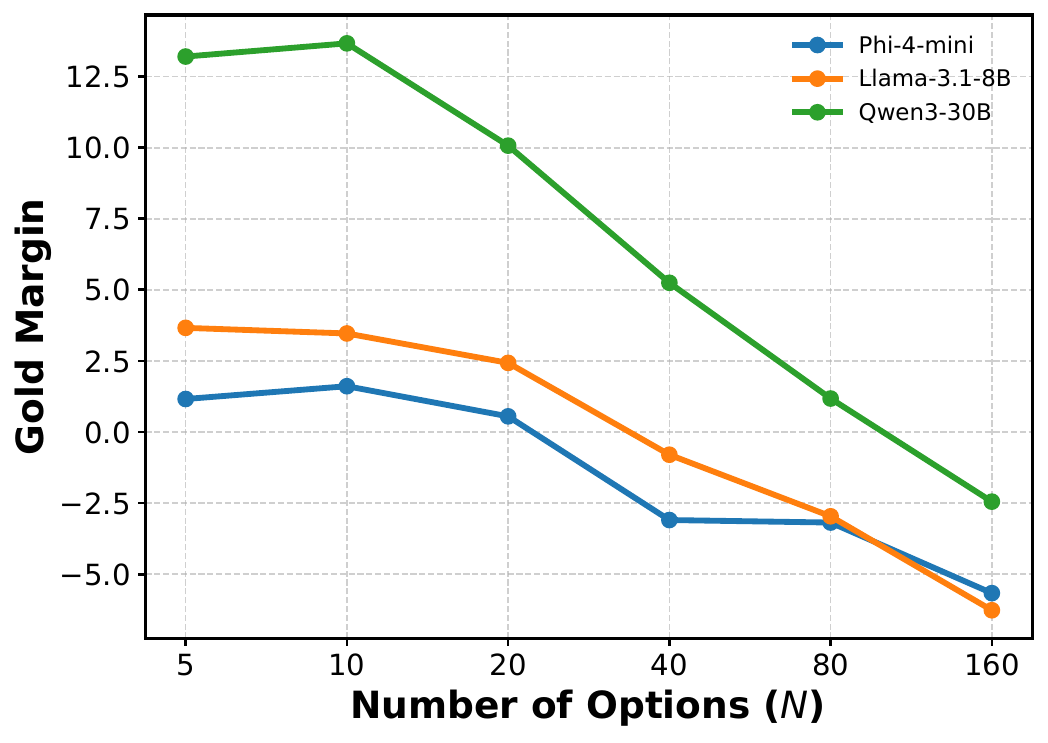}
        \caption{MIMIC}
    \end{subfigure}
    \hfill
    \begin{subfigure}[t]{0.31\linewidth}
        \centering
        \includegraphics[width=\linewidth]{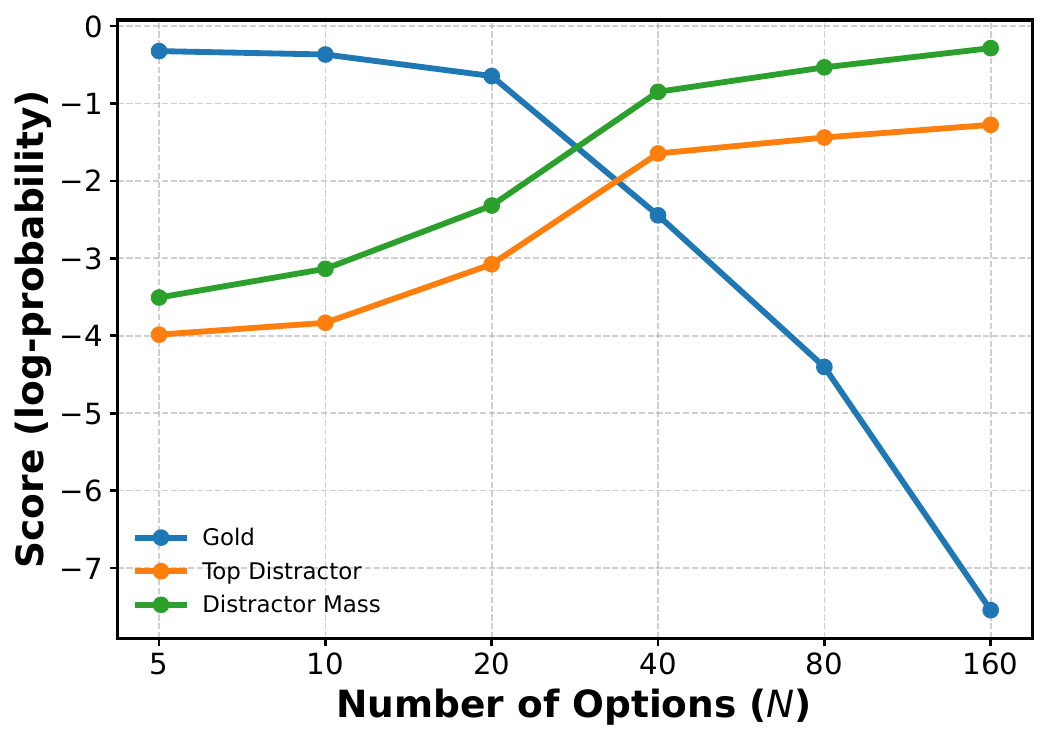}
        \caption{Score dynamics (MIMIC)}
    \end{subfigure}

    \caption{Gold margin across models and score dynamics for Llama-3.1-8B as the number of options increases.
    As the number of options grows, the gold margin decreases while the gold score collapses substantially faster than the distractor score.}
    \label{fig:main_analysis}
\end{figure*}

Figure~\ref{fig:main_analysis}~(a,b) shows that the gold margin decreases consistently as the number of options increases, eventually becoming negative in some large-option regimes. Figure~\ref{fig:main_analysis}(c) further shows that while distractor scores remain relatively stable, the gold score decreases sharply. This suggests that degradation is driven primarily by collapsing confidence in the correct answer rather than increasingly strong distractors.

\subsection{Context Length Alone Does Not Fully Explain Large-Option Degradation}

To isolate the effect of context length, we increase prompt length while holding candidate comparison structure fixed.

Table~\ref{tab:noise_delta} shows that increasing irrelevant context length within the prompt does not produce the degradation patterns observed under candidate-set scaling (Figure~\ref{fig:accuracy_six_datasets}).

Similar trends are observed under distractor-verbosity controls (Appendix~\ref{appendix:verbosity_control}), where substantially longer options also fail to consistently reproduce the degradation induced by candidate-set expansion.

Overall, these results suggest that context length alone cannot fully explain the stable degradation observed under candidate-set scaling.

\begin{table*}[t]
\caption{
Accuracy under increasing irrelevant context length.
Base denotes accuracy without added tokens; remaining columns report
changes relative to Base.
}
\label{tab:noise_delta}

\centering
\small
\setlength{\tabcolsep}{5pt}

\begin{tabular}{llcccccccc}
\toprule
\multirow{2}{*}{Task} & \multirow{2}{*}{Model}
& \multicolumn{7}{c}{Added Tokens ($\Delta$ Accuracy)}
& \multirow{2}{*}{Avg. $\Delta$} \\
& & Base & +235 & +470 & +940 & +1880 & +3760 & +7520 & \\
\midrule

\multirow{3}{*}{HotpotQA}
& Phi-4-mini   & 72.00 & +7.00 & +7.00 & +5.50 & +7.00 & +6.00 & +2.50 & +6.43 \\
& Llama-3.1-8B & 86.50 & +0.50 & +2.00 & +0.50 & +1.50 & +0.50 & +1.50 & +1.07 \\
& Qwen3-30B    & 99.00 & -0.50 & -0.50 & -0.50 & -0.50 & -0.50 & +0.00 & -0.43 \\

\midrule

\multirow{3}{*}{MIMIC}
& Phi-4-mini   & 35.00 & -5.50 & -3.50 & -3.50 & -3.00 & -1.50 & -5.00 & -3.57 \\
& Llama-3.1-8B & 56.00 & -3.50 & -6.00 & -5.50 & -8.50 & -7.50 & -8.50 & -6.50 \\
& Qwen3-30B    & 80.00 & -2.50 & -1.00 & -1.00 & -4.50 & -3.00 & -3.00 & -2.50 \\

\bottomrule
\end{tabular}
\end{table*}

\subsection{Attention Distribution Does Not Directly Reflect Prediction Decisions}

Prior work on \textit{lost-in-the-middle} and subsequent studies on long-context reasoning links long-context positional behavior to transformer attention patterns. We therefore compare attention allocation with empirical prediction-position distributions under large candidate sets.

\begin{figure*}[t]
    \centering

    \begin{subfigure}[t]{0.42\linewidth}
        \centering
        \includegraphics[width=\linewidth]{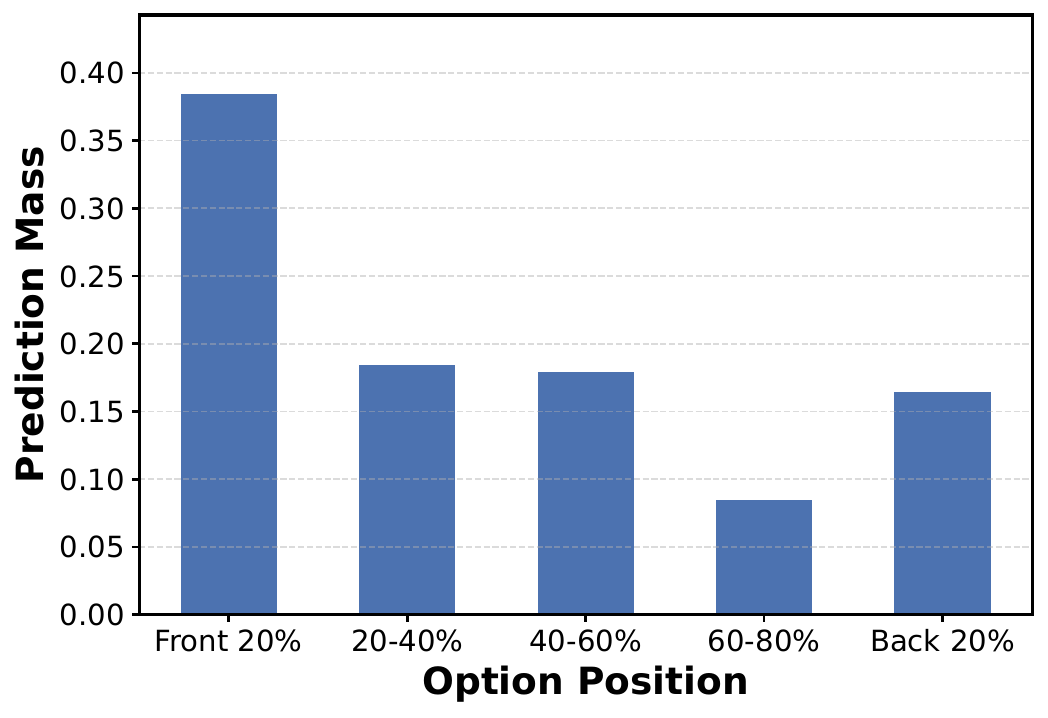}
        \caption{Llama-3.1-8B prediction}
    \end{subfigure}
    \hspace{0.05\linewidth}
    \begin{subfigure}[t]{0.42\linewidth}
        \centering
        \includegraphics[width=\linewidth]{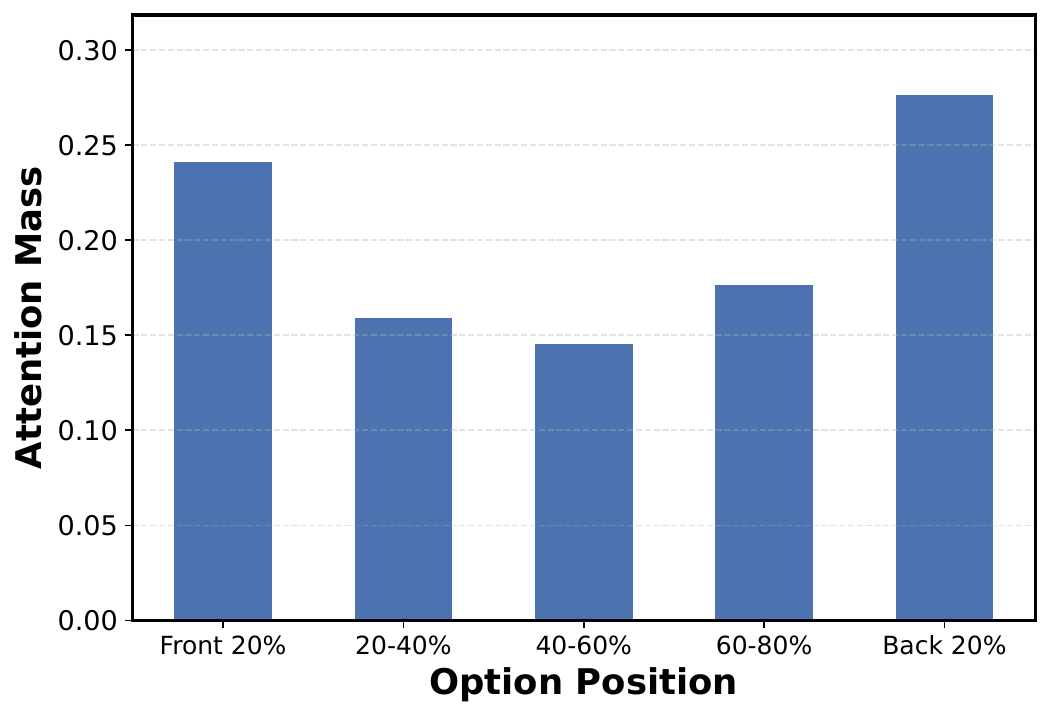}
        \caption{Llama-3.1-8B attention}
    \end{subfigure}

    \begin{subfigure}[t]{0.42\linewidth}
        \centering
        \includegraphics[width=\linewidth]{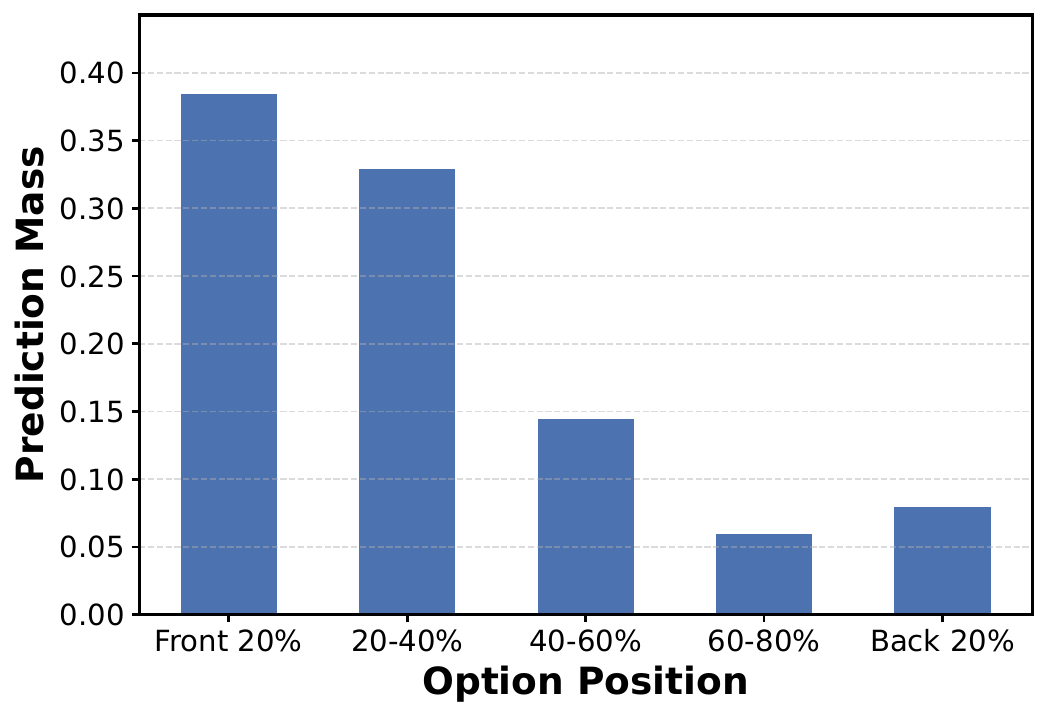}
        \caption{Phi-4-mini prediction}
    \end{subfigure}
    \hspace{0.05\linewidth}
    \begin{subfigure}[t]{0.42\linewidth}
        \centering
        \includegraphics[width=\linewidth]{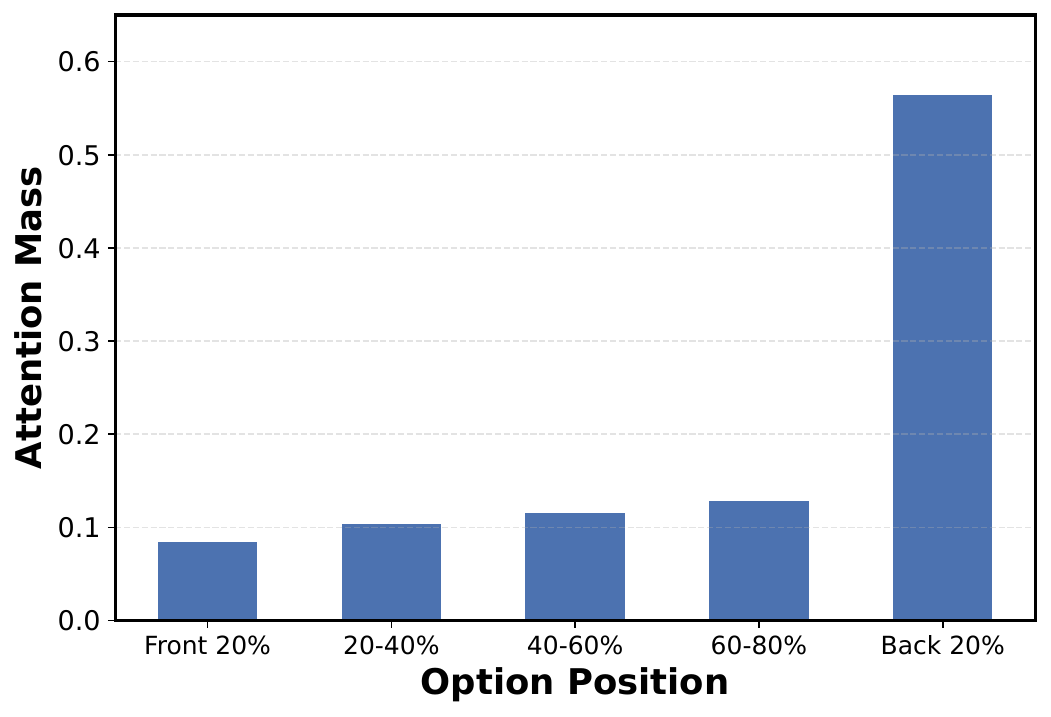}
        \caption{Phi-4-mini attention}
    \end{subfigure}

    \caption{
    Prediction-position and attention distributions on HotpotQA at $N=80$. Both Llama-3.1-8B and Phi-4-mini exhibit front-skewed prediction distributions, despite substantially different attention patterns. Llama-3.1-8B shows relatively balanced attention across candidate positions, whereas Phi-4-mini allocates substantially more attention to later positions.
    }
    \label{fig:attn_pred_compare}
\end{figure*}

Figure~\ref{fig:attn_pred_compare} compares attention distributions and empirical prediction-position distributions on HotpotQA with $N=80$.

While attention allocation patterns vary substantially across models, prediction distributions remain consistently front-skewed. This discrepancy persists across candidate-set sizes and datasets (Appendix~\ref{appendix:attention_generalization}), suggesting that
attention allocation does not directly explain prediction-position
behavior.

\subsection{Early-Option Dominance in Large-Option Comparison}

Previous analyses showed that model predictions become front-skewed under large candidate settings (Figure~\ref{fig:attn_pred_compare}(a,c)). The trend becomes progressively stronger as the number of candidate options increases and persists under reversed-index controls (Appendix~\ref{sec:appendix-pos_bias_generalization}).

\begin{figure*}[t]
    \centering

    \begin{subfigure}[t]{0.40\linewidth}
        \centering
        \includegraphics[width=\linewidth]{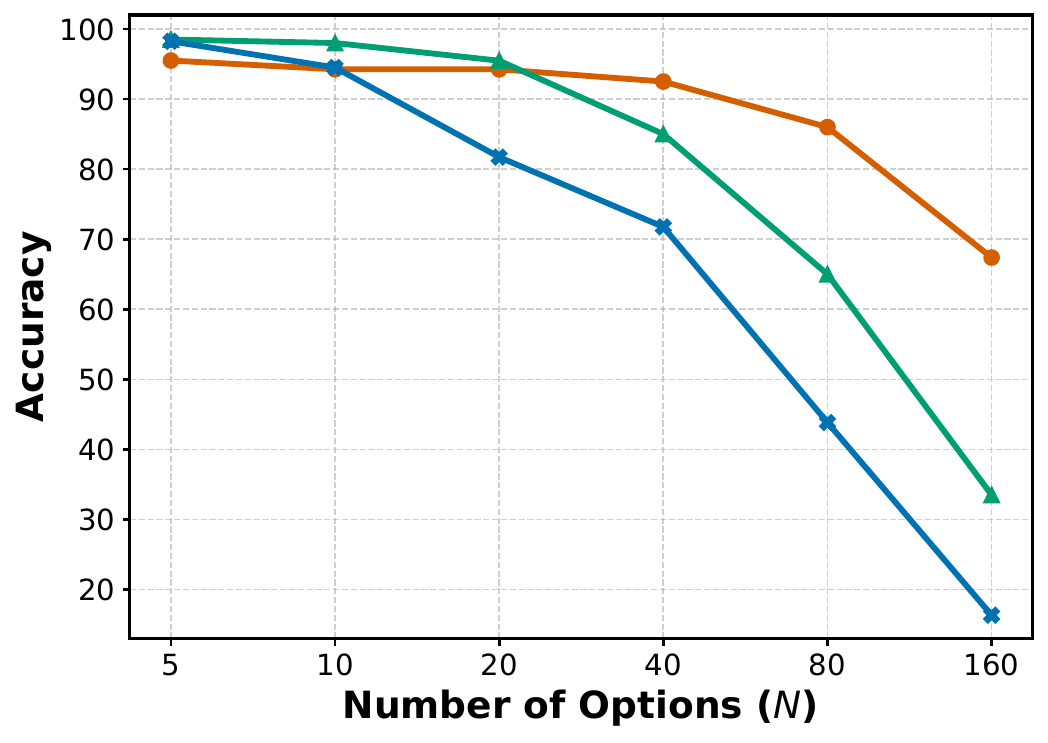}
        \caption{HotpotQA}
    \end{subfigure}
    \hspace{0.03\linewidth}
    \begin{subfigure}[t]{0.40\linewidth}
        \centering
        \includegraphics[width=\linewidth]{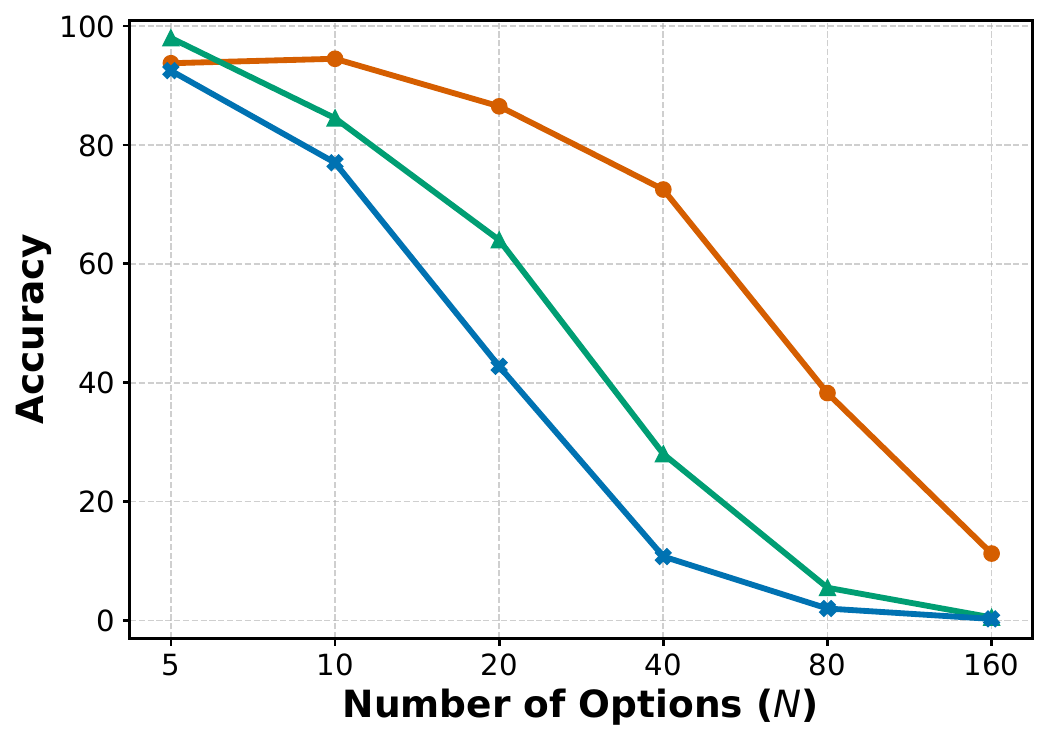}
        \caption{MIMIC}
    \end{subfigure}

    \vspace{0.1em}
    \includegraphics[width=0.35\linewidth]{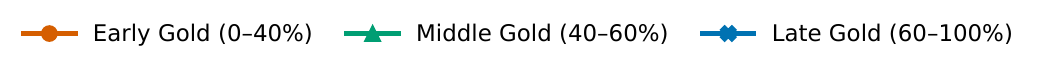}

    \caption{
    Controlled gold-position intervention on Llama-3.1-8B.
    Later-positioned gold answers become progressively harder to recover as the candidate set grows.
    }
    \label{fig:controlled_gold_position}
\end{figure*}

To further examine how early-option dominance affects prediction accuracy, we vary the gold insertion region while holding the distractor set fixed. Figure~\ref{fig:controlled_gold_position} shows that later-positioned gold answers become progressively harder to recover as $N$ increases. Probability-space analyses show the same pattern (Appendix~\ref{sec:appendix-additional-probability-dynamics}), where prediction mass becomes progressively less responsive to later gold insertions as the candidate set grows.

To further characterize the comparison dynamics underlying early-option dominance, we conduct a fine-grained gold-insertion experiment. For each example, we first identify the strongest distractor before gold insertion, which we refer to as the \emph{pre-anchor}, and then insert the gold answer at different positions. We report two persistence-based measures: \emph{anchor persistence}, which measures whether the final prediction remains the pre-anchor after gold insertion, and \emph{retained anchor rate}, which measures whether the pre-anchor remains the highest-scoring distractor even when the final prediction changes.

\begin{figure*}[t]
    \centering

    \begin{subfigure}[t]{0.42\linewidth}
        \centering
        \includegraphics[width=\linewidth]{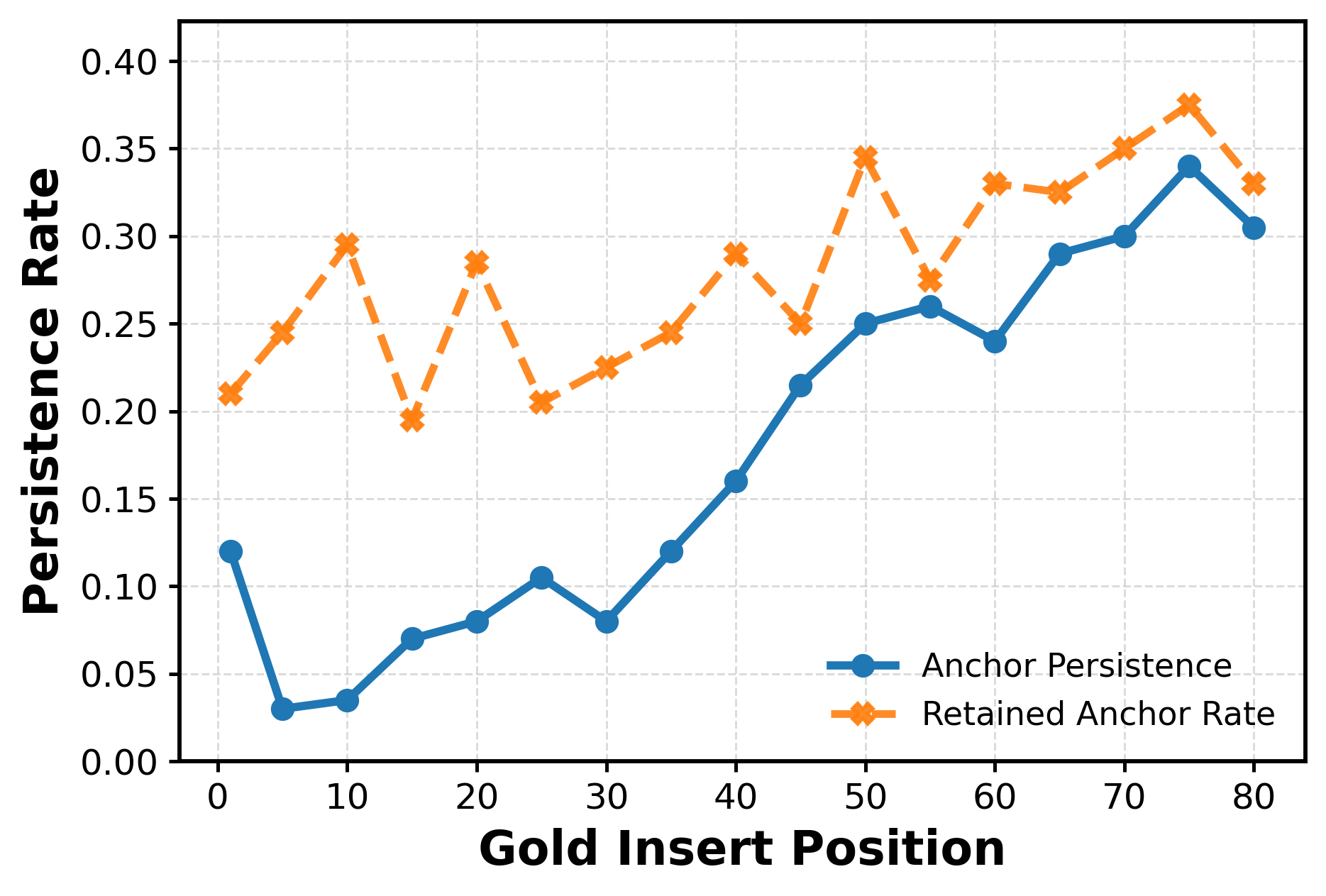}
        \caption{Persistence measures ($N=80$)}
    \end{subfigure}
    \hspace{0.025\linewidth}
    \begin{subfigure}[t]{0.42\linewidth}
        \centering
        \includegraphics[width=\linewidth]{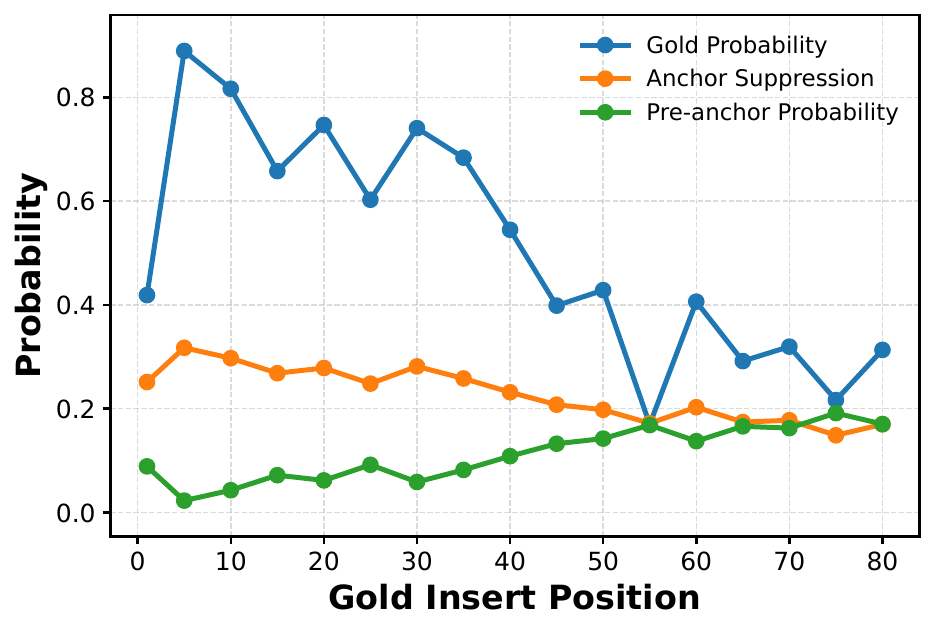}
        \caption{Probability dynamics ($N=80$)}
    \end{subfigure}

    \caption{Revision dynamics under large candidate sets on HotpotQA (Llama-3.1-8B, $N=80$).
    (a) Anchor persistence measures whether the prediction remains the pre-anchor after gold insertion, while retained anchor rate measures whether the pre-anchor remains the strongest distractor.
    (b) Gold probability decreases for later insertions, weakening suppression of the pre-anchor and making earlier preferences harder to overturn.}
    \label{fig:seq_anchor}
\end{figure*}

Figure~\ref{fig:seq_anchor} shows that later-positioned gold answers progressively lose their ability to overturn earlier distractor preferences. As the gold answer is inserted later in the sequence, gold probability steadily decreases, weakening its suppressive effect on the pre-anchor distractor. As a result, anchor persistence increases and earlier distractor preferences become more likely to survive as the final prediction. Despite minor edge-position effects, the overall trend remains consistent across datasets and models (Appendix~\ref{sec:appendix-additional-sequential-anchor}).

These results suggest that progressive early-option dominance reduces effective candidate exploration under large candidate sets, contributing to the observed accuracy degradation under large-option scaling.

\section{Analysis-Driven Intervention}

\subsection{Overview}

Motivated by the observed gold-margin collapse and early-option dominance, we evaluate two inference-time interventions that restructure the comparison process: (1) Hierarchical Partitioning, which performs staged candidate selection over small groups, and (2) Permutation-based Inference, which aggregates predictions across multiple candidate orderings. We additionally include Chain-of-Thought (CoT) as a strong reasoning-based baseline. Detailed implementation settings are provided in Appendix~\ref{sec:appendix-addn-impl-details}.

\subsection{Intervention Results}

\begin{figure*}[t]
    \centering
    \begin{subfigure}[t]{0.42\linewidth}
        \centering
        \includegraphics[width=\linewidth]{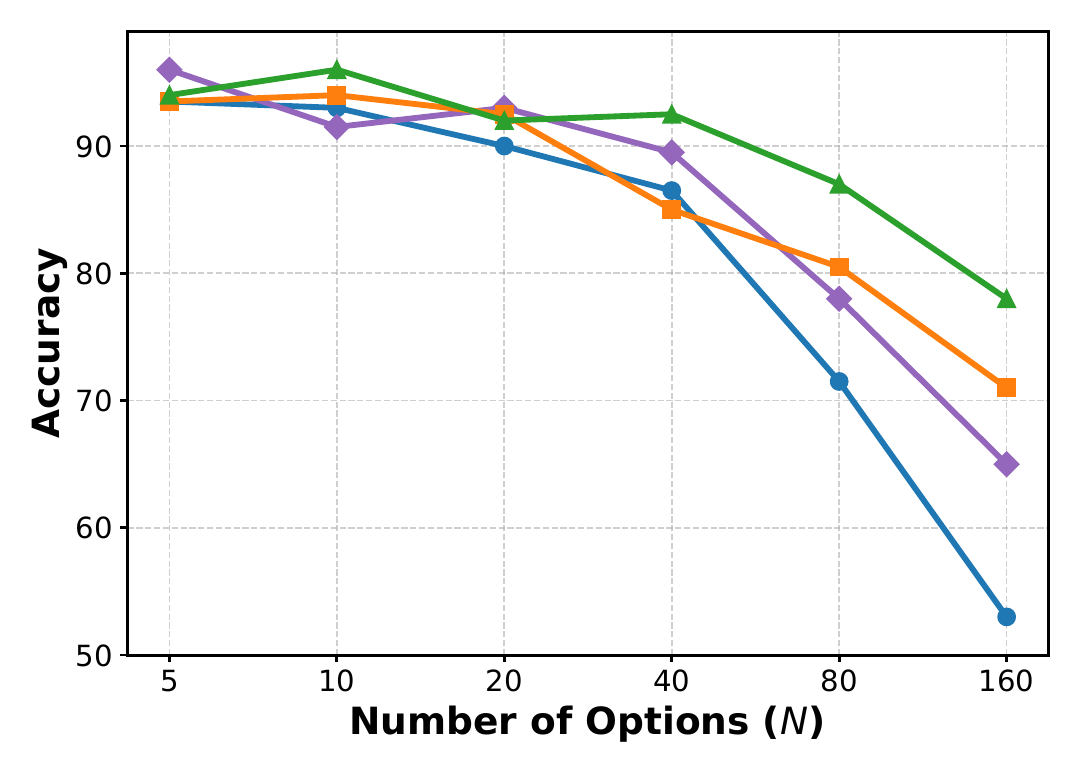}
        \caption{HotpotQA}
    \end{subfigure}
    \hspace{0.025\linewidth}
    \begin{subfigure}[t]{0.42\linewidth}
        \centering
        \includegraphics[width=\linewidth]{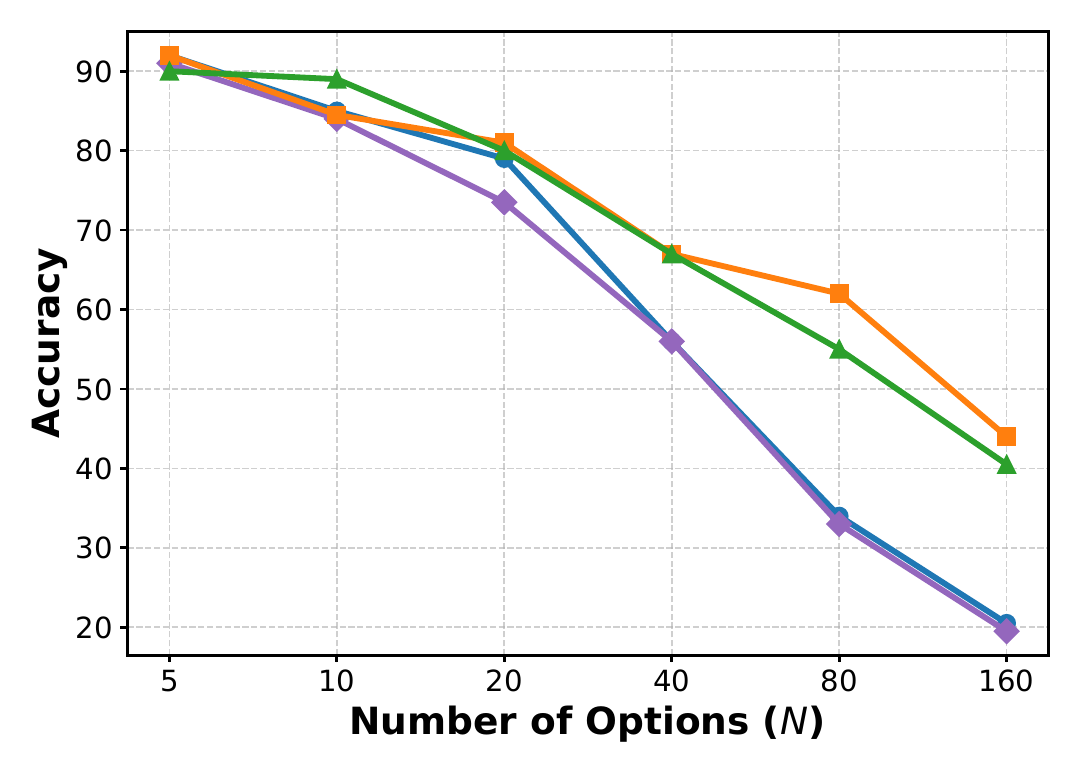}
        \caption{MIMIC}
    \end{subfigure}

    \vspace{0.3em}

    \begin{subfigure}[t]{0.42\linewidth}
        \centering
        \includegraphics[width=\linewidth]{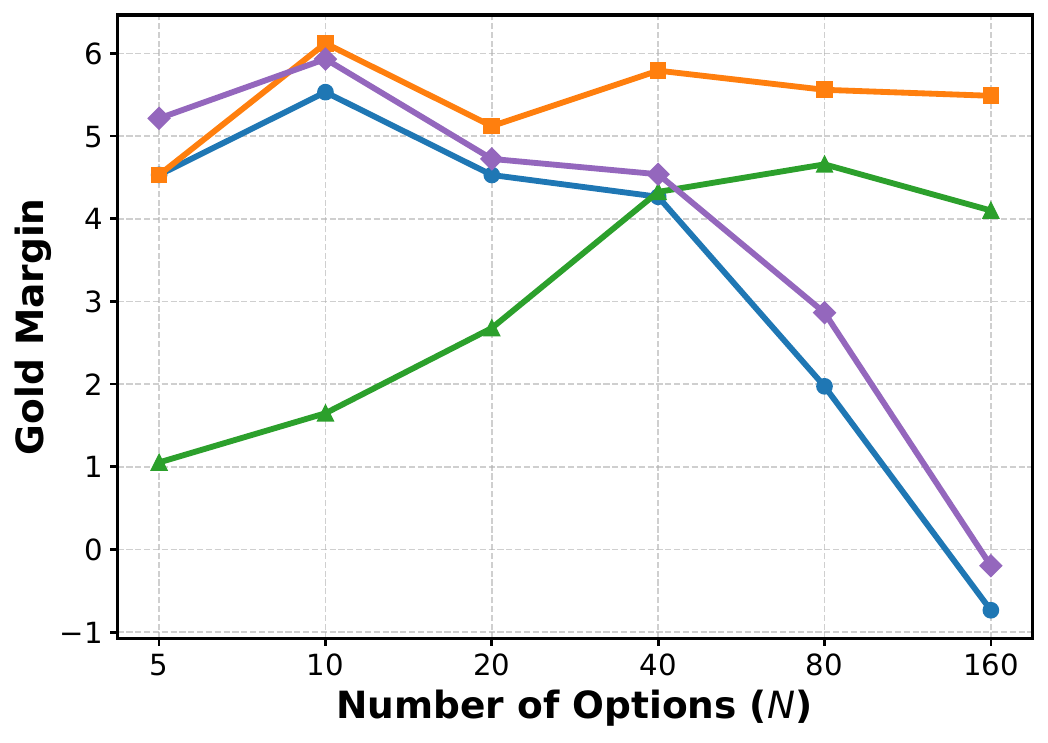}
        \caption{Gold margin (HotpotQA)}
    \end{subfigure}
    \hspace{0.025\linewidth}
    \begin{subfigure}[t]{0.42\linewidth}
        \centering
        \includegraphics[width=\linewidth]{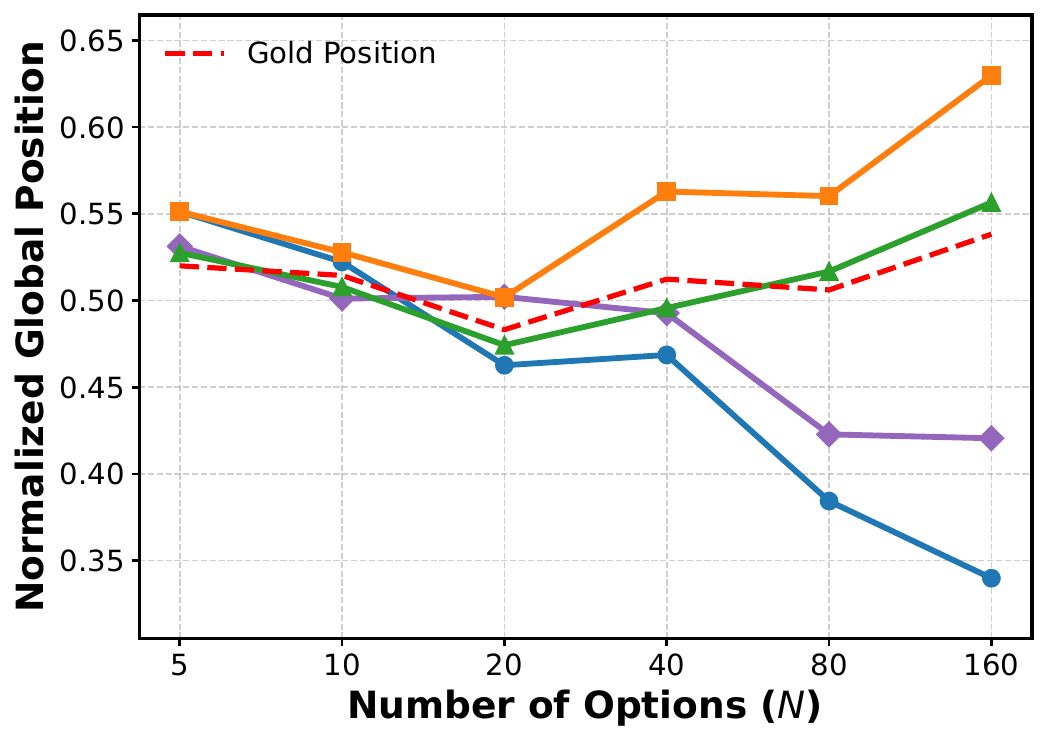}
        \caption{Position bias (HotpotQA)}
    \end{subfigure}

    \vspace{0.1em}
    \includegraphics[width=0.55\linewidth]{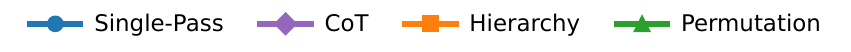}

    \caption{Effects of restructuring candidate comparison (Llama-3.1-8B-Instruct).
    (a,b) Hierarchical partitioning and permutation-based inference increasingly outperform single-pass inference as candidate-set size grows.
    (c,d) Both interventions maintain larger gold margins and reduce early-position bias under large candidate sets.}
    \label{fig:all_methods_main}
\end{figure*}

As shown in Figure~\ref{fig:all_methods_main}(a,b), both Hierarchy and Permutation consistently outperform Single-Pass inference, with gains increasing as the number of options grows. At \(N=160\), both methods improve accuracy by roughly 20 percentage points on both HotpotQA and MIMIC. In contrast, although CoT improves absolute accuracy, it still degrades substantially at larger option sizes.

\subsection{Mechanism Analysis}

To analyze the source of improvement, we evaluate model behavior under the final decision stage of each intervention.

Figure~\ref{fig:all_methods_main}(c,d) shows that hierarchical partitioning and permutation-based inference maintain larger gold margins and weaker early-position bias than single-pass inference as candidate sets grow. In contrast, CoT improves overall accuracy without substantially mitigating either failure pattern. Permutation exhibits lower final-round gold margins at small $N$ because examples resolved through early agreement do not enter the final comparison stage, leaving a more difficult subset (Figure~\ref{fig:inference_mechanisms}(b)).

These results suggest that restructuring candidate comparison improves performance by directly mitigating the failure patterns identified in our analysis. Additional intervention-specific behaviors and computational trade-offs are provided in Appendix~\ref{sec:appendix_intervention_mechanism}.

\section{Related Work}

\paragraph{MCQ-based Benchmarking}
Traditional MCQ benchmarks typically evaluate LLMs with few answer choices
\citep{mmlu,gpqa,gsm-mc}. Recent work shows that performance declines with
larger option sets, stronger distractors, or increased reasoning demands
\citep{mmlu-pro,zhang2025rethinking,elhady-etal-2025-wicked}.
We focus on the failure mechanisms that emerge as candidate comparison scales.

\paragraph{Long-Context Reasoning and Positional Bias}
LLMs exhibit context interference, retrieval degradation, and positional biases
under long contexts
\citep{hong2025contextrot,shi2023distracted,du-etal-2025-context,
lee-etal-2025-critical,liu-etal-2024-lost,hsieh2024found,wu2025positionbias}.
While this literature primarily studies retrieval-oriented behavior, we examine
comparison among many competing candidates.

\paragraph{Decision-Making Degradation and Mitigation}
Prior work studies distractor competition and selection bias
\citep{elhady-etal-2025-wicked,amanlou2026knight,
DBLP:journals/corr/abs-2309-03882}, as well as inference-time mitigation
\citep{parekh-etal-2025-dicore,gan2025rag,zheng2026mitigating}.
Our work connects large-option degradation to comparison-level failure patterns
and evaluates interventions that restructure the comparison process.

\section{Conclusion}

We study large-option decision making in LLMs and find consistent performance degradation as candidate sets scale. Controlled analyses suggest that this degradation cannot be fully explained by standard long-context retrieval failures, revealing systematic weaknesses in scalable candidate comparison, including gold-margin collapse and early-option dominance. Restructuring the comparison process through hierarchical partitioning and permutation-based inference substantially improves robustness under large candidate sets. Overall, our results highlight candidate-set scale as an important evaluation variable and suggest that strong performance in small-option settings may not reliably reflect scalable candidate-comparison ability.

\FloatBarrier
\setlength{\bibsep}{0pt}

\section{Limitations}

Our study focuses on inference-time behavior in primarily English,
text-only, single-answer candidate-selection tasks. While our controlled
experiments reveal systematic gold-margin collapse and early-option
dominance, they characterize behavioral patterns rather than establishing
a complete causal account of the underlying model mechanisms.
Furthermore, the proposed restructuring strategies require additional
inference computation. Extending these findings to multilingual,
multimodal, and interactive settings, as well as developing more
compute-efficient interventions, remains important future work.

\bibliographystyle{plainnat}
\bibliography{custom}

\section{Appendix}

\subsection{Dataset Statistics}
For each dataset, we randomly sample 200 evaluation instances to construct the large-option benchmark setting.

Table~\ref{tab:dataset_stats} summarizes the datasets used in our experiments,
including task types, average option lengths, and total prompt lengths at $N{=}160$.

\begin{table}[t]
\caption{Dataset statistics.
We report task type, average option length, and total prompt length at $N{=}160$.}
\label{tab:dataset_stats}

\centering
\scriptsize
\renewcommand{\arraystretch}{1.08}
\setlength{\tabcolsep}{2.5pt}

\resizebox{\linewidth}{!}{
\begin{tabular}{llcc}
\toprule
\textbf{Dataset} &
\textbf{Task} &
\shortstack[c]{\textbf{Avg. Option}\\\textbf{Tokens}} &
\shortstack[c]{\textbf{Prompt Tokens}\\\textbf{at $N{=}160$}} \\
\midrule
MSMARCO   & Reranking                    & 79.8  & 12,763.4 \\
HotpotQA  & Multi-hop Reranking          & 90.2  & 14,434.8 \\
MIMIC   & Medical Diagnosis            & 116.4 & 18,641.5 \\
arXiv     & Long-context Classification  & 8.8   & 1,413.9 \\
Amazon-3M & Taxonomy Prediction          & 13.9  & 2,229.9 \\
MMLU-Pro  & Knowledge Reasoning          & 11.2  & 2,202.6 \\
\bottomrule
\end{tabular}
}
\end{table}

\vspace{0.8em}

\subsection{Additional Datasets for Accuracy Degradation}

\begin{figure*}[t]
    \centering

    \begin{subfigure}[t]{0.31\linewidth}
        \centering
        \includegraphics[width=\linewidth]{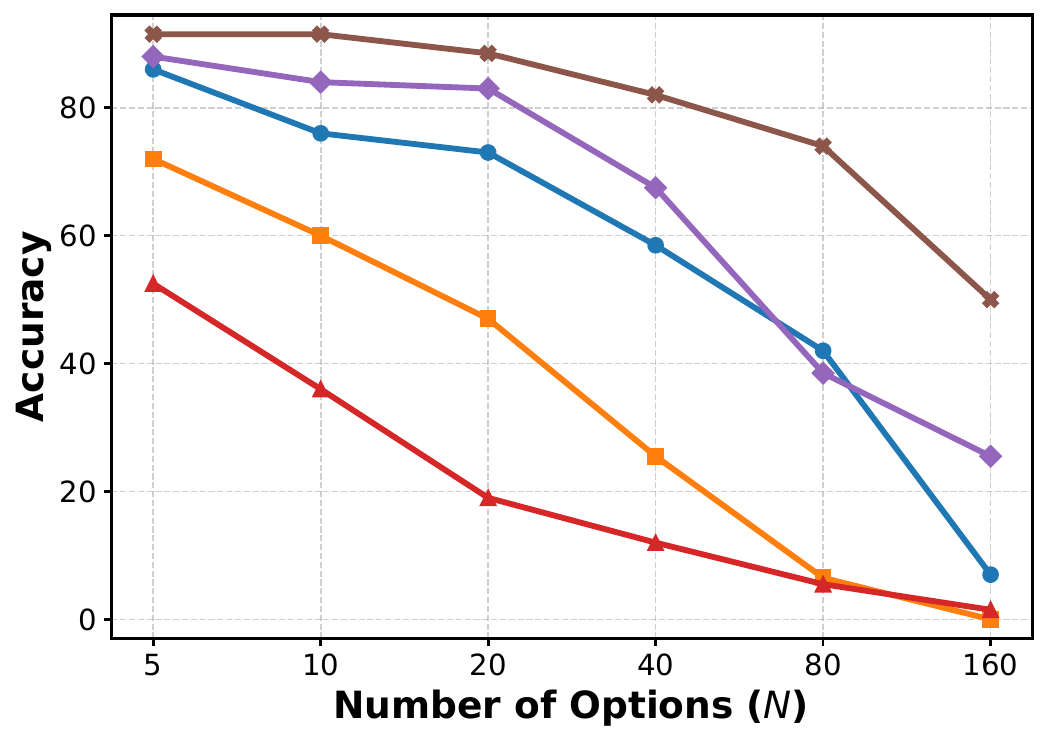}
        \caption{Arguana}
    \end{subfigure}
    \hfill
    \begin{subfigure}[t]{0.31\linewidth}
        \centering
        \includegraphics[width=\linewidth]{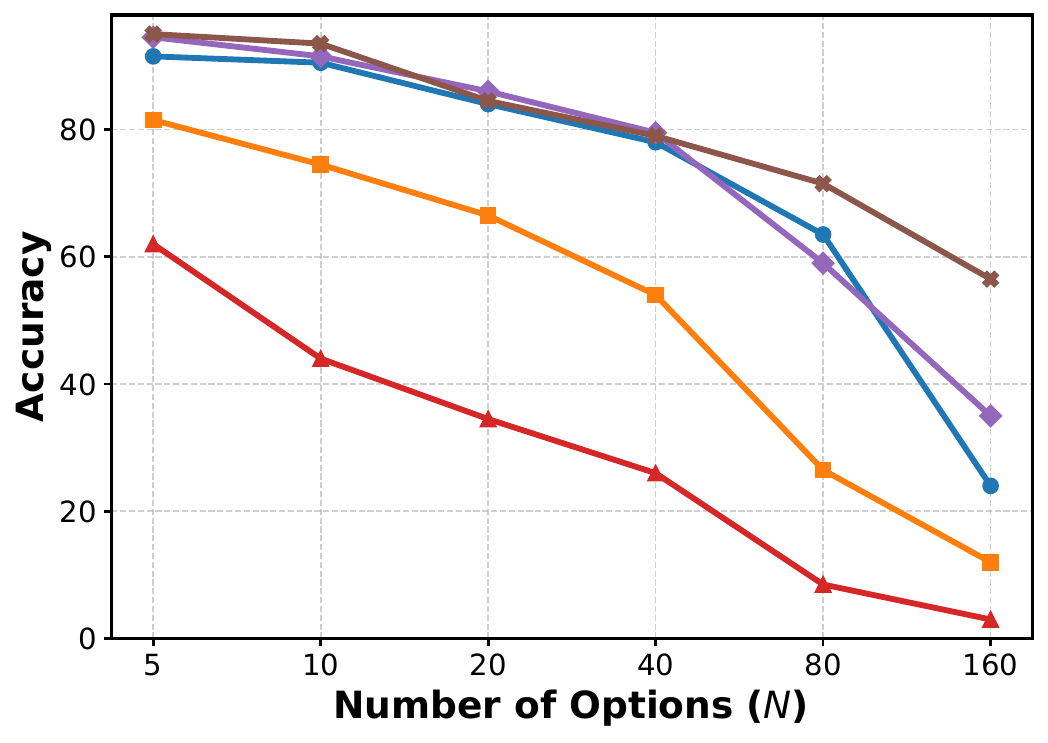}
        \caption{Climate-Fever}
    \end{subfigure}
    \hfill
    \begin{subfigure}[t]{0.31\linewidth}
        \centering
        \includegraphics[width=\linewidth]{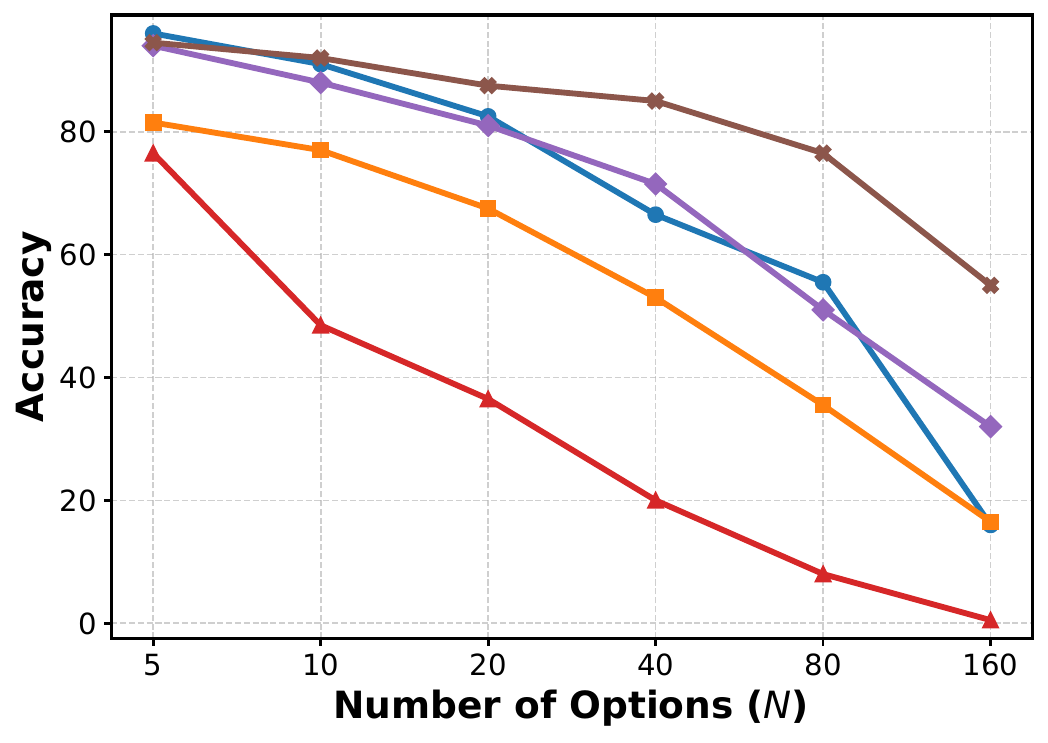}
        \caption{FiQA}
    \end{subfigure}

    \vspace{0.35em}

    \begin{subfigure}[t]{0.31\linewidth}
        \centering
        \includegraphics[width=\linewidth]{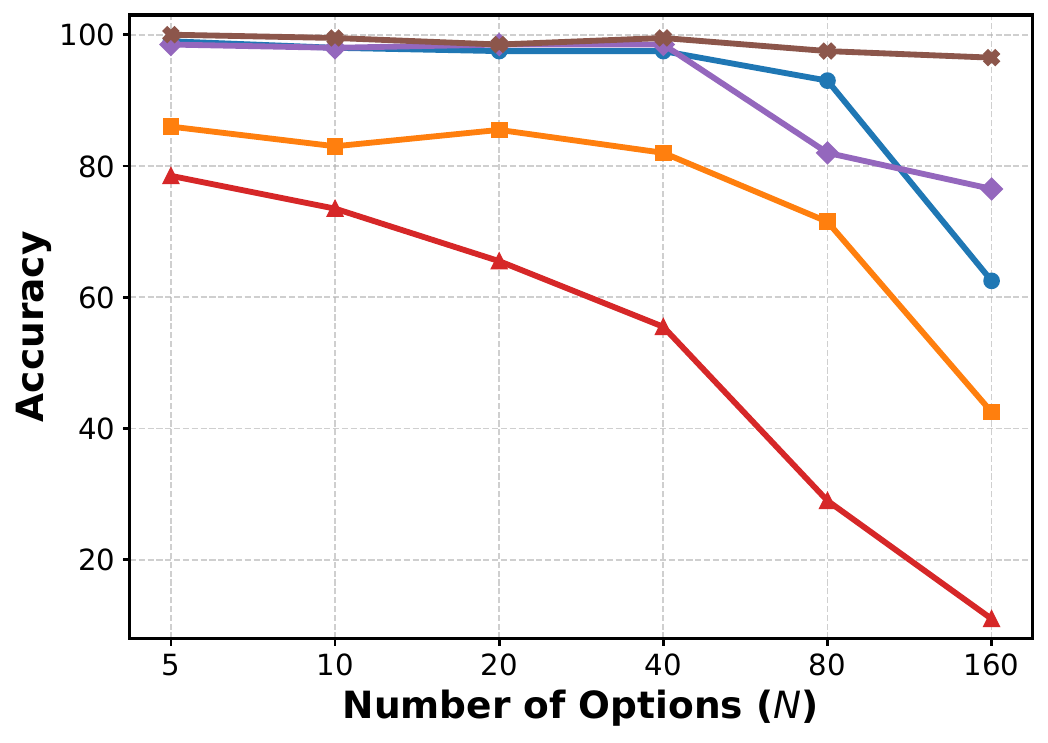}
        \caption{FEVER}
    \end{subfigure}
    \hfill
    \begin{subfigure}[t]{0.31\linewidth}
        \centering
        \includegraphics[width=\linewidth]{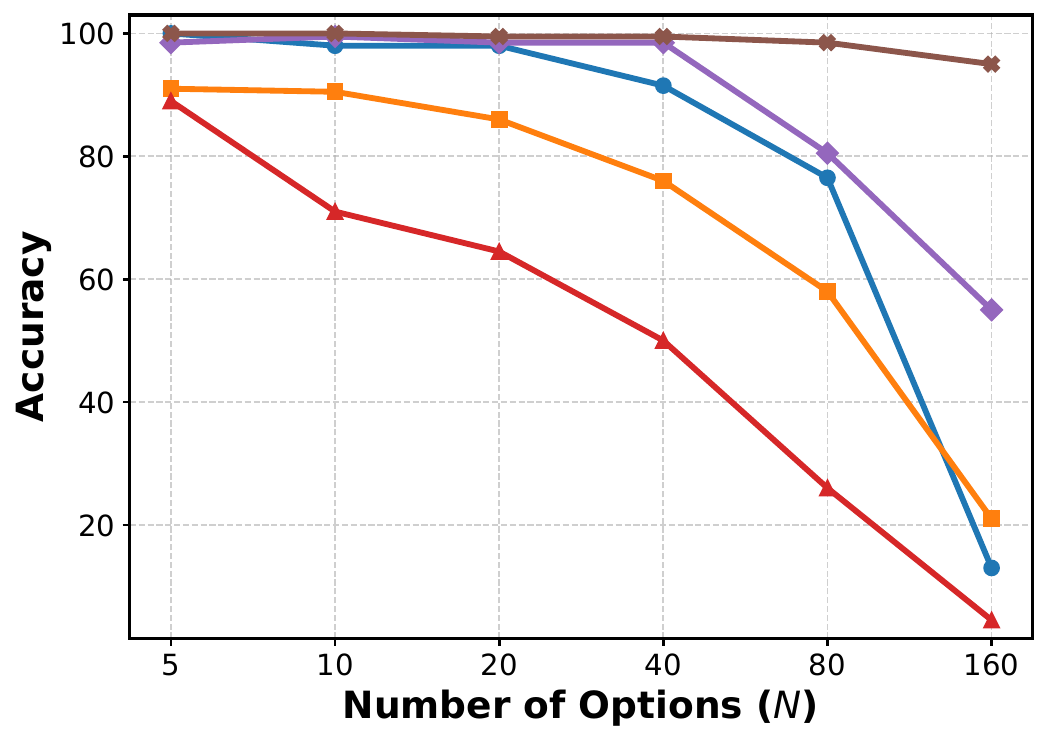}
        \caption{GermanQuAD}
    \end{subfigure}
    \hfill
    \begin{subfigure}[t]{0.31\linewidth}
        \centering
        \includegraphics[width=\linewidth]{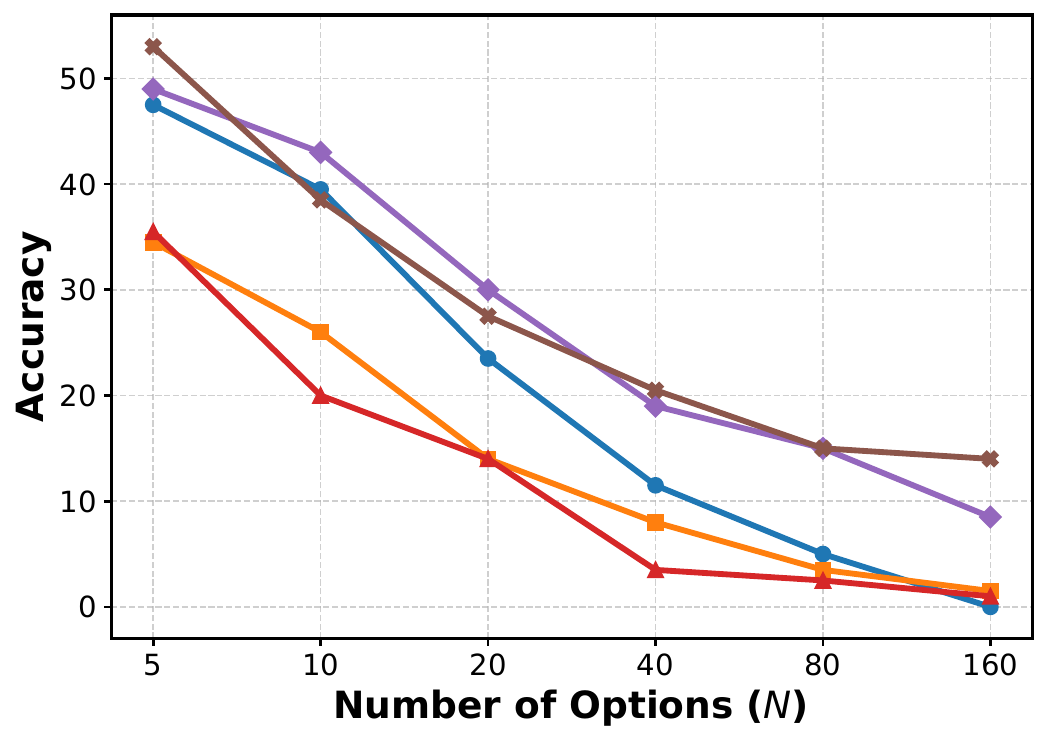}
        \caption{NFCorpus}
    \end{subfigure}

    \vspace{0.35em}

    \begin{subfigure}[t]{0.31\linewidth}
        \centering
        \includegraphics[width=\linewidth]{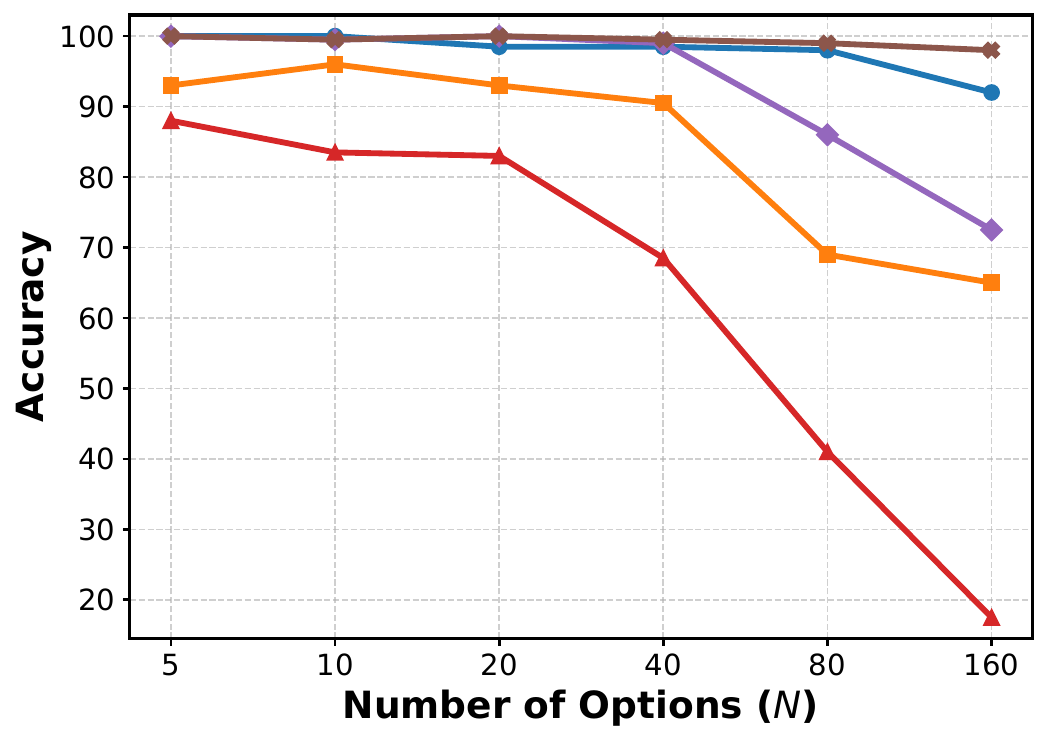}
        \caption{NQ}
    \end{subfigure}
    \hfill
    \begin{subfigure}[t]{0.31\linewidth}
        \centering
        \includegraphics[width=\linewidth]{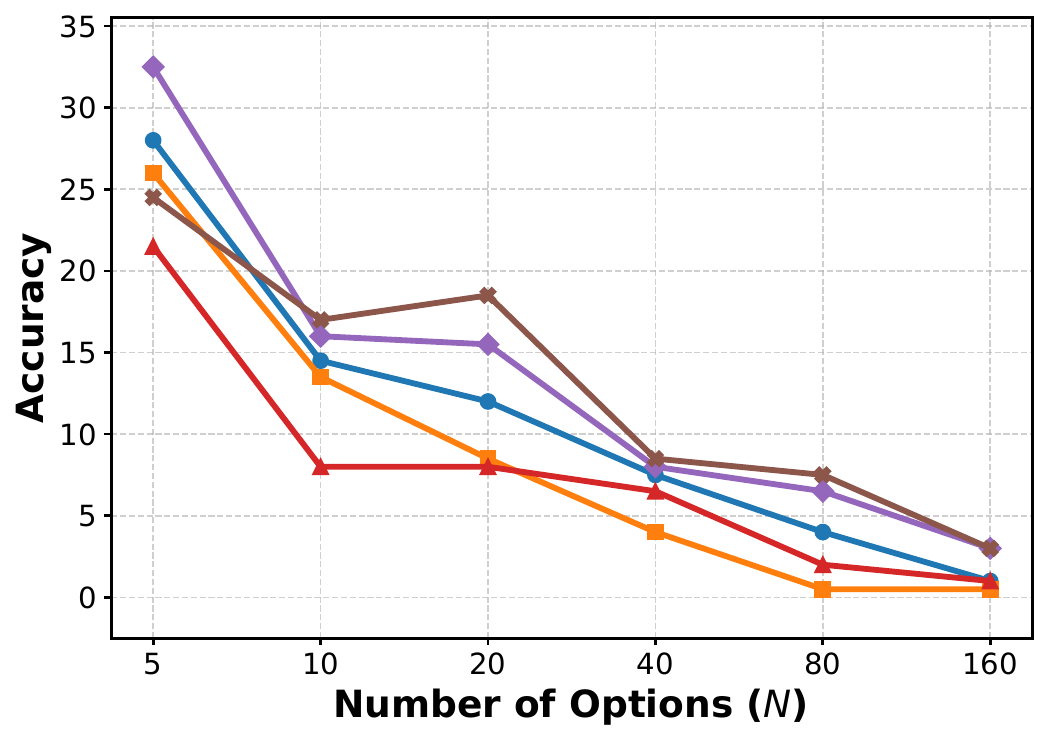}
        \caption{SciDocs}
    \end{subfigure}
    \hfill
    \begin{subfigure}[t]{0.31\linewidth}
        \centering
        \includegraphics[width=\linewidth]{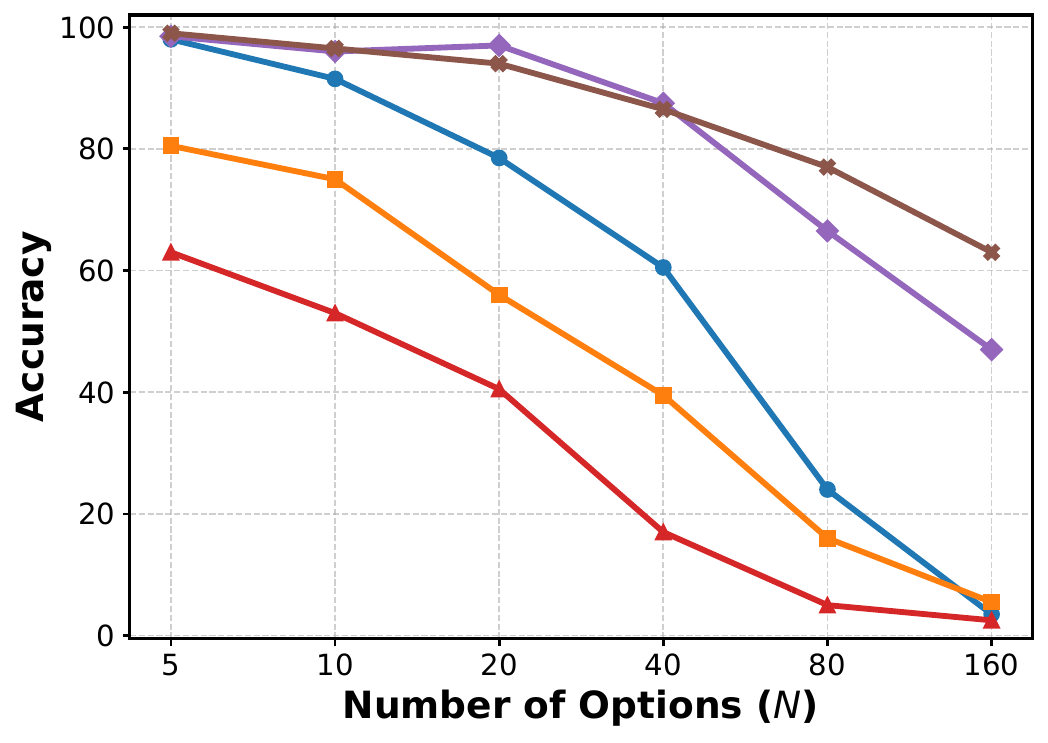}
        \caption{SciFact}
    \end{subfigure}

    \vspace{0.35em}
    \includegraphics[width=0.5\linewidth]{Figs/option_acc/legend_only.pdf}

    \caption{Accuracy degradation across additional datasets.
    The degradation trend remains consistent across a wide range of tasks,
    supporting the generality of our findings beyond the main benchmarks.}
    \label{fig:accuracy_appendix_datasets}
\end{figure*}

To further validate the generality of the observed degradation, we evaluate a broader set of datasets beyond the main text.
As shown in Figure~\ref{fig:accuracy_appendix_datasets}, accuracy consistently declines as the number of options increases across diverse tasks.
This trend holds across different domains, indicating that the degradation is not dataset-specific.

\vspace{0.5em}

\vspace{0.8em}

\subsection{Additional Implementation Details and Results for System-2 Reasoning}
\label{appendix:system2_reasoning}

\textbf{Additional Implementation Details.} We utilized the following common System-2 inference-time strategies to test our hypothesis on accuracy degradation at scale: \textbf{(1) Chain-of-thought (CoT) Reasoning}, where the model develops a reasoning chain before selecting a response, \textbf{(2) Self-Critique}, where the model makes a choice and critiques it before selecting a response, \textbf{(3) Few-shot Prompting}, where the model is provided sample questions and answers in its context, and \textbf{(4) Pairwise Comparison}, where the model is repeatedly shown two options at a time and performs an aggregation to select a final answer. 

\begin{figure*}[t]
    \centering

    \begin{subfigure}[t]{0.44\linewidth}
        \centering
        \includegraphics[width=\linewidth]{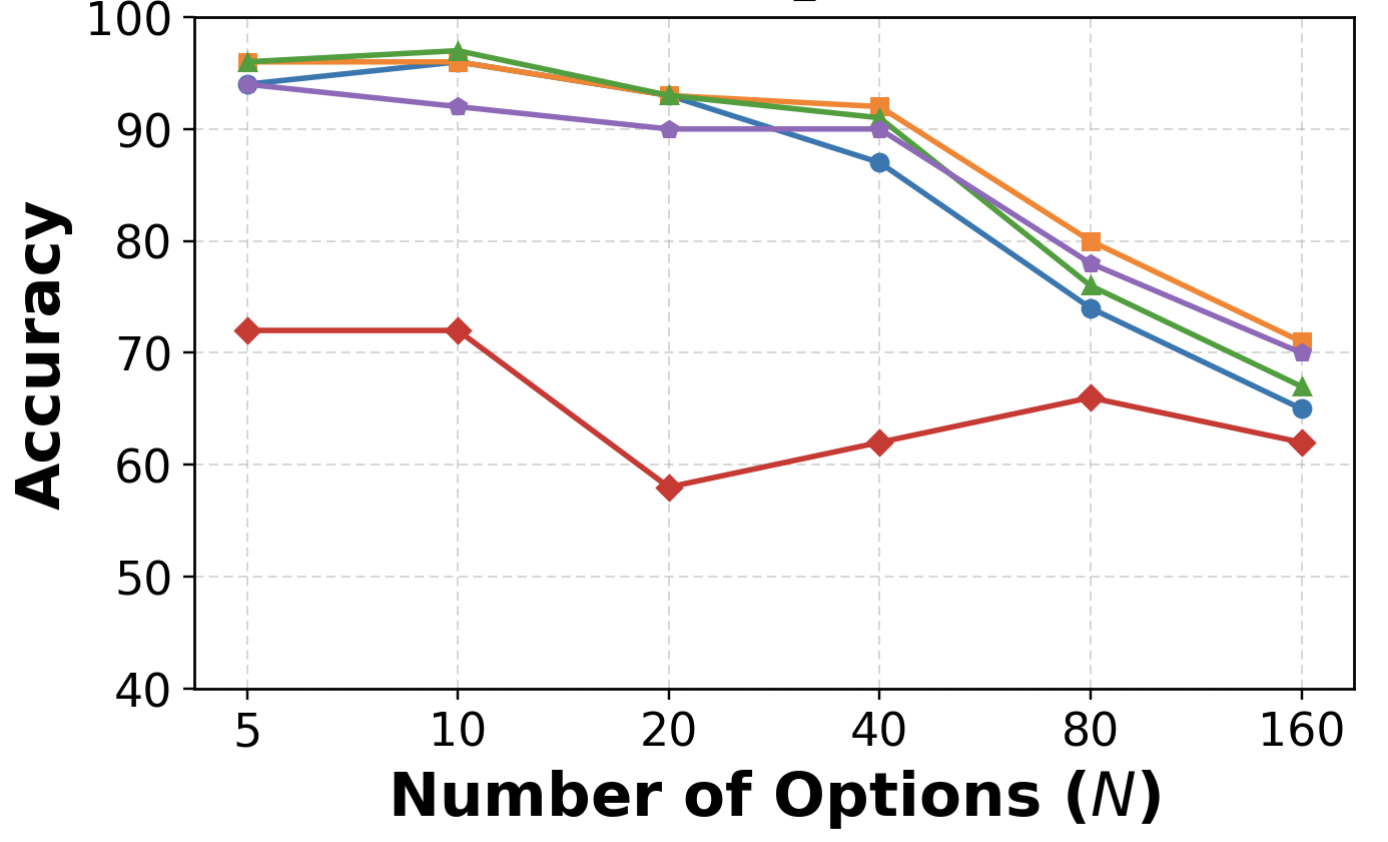}
        \caption{HotpotQA (Qwen3-4B)}
    \end{subfigure}
    \hfill
    \begin{subfigure}[t]{0.44\linewidth}
        \centering
        \includegraphics[width=\linewidth]{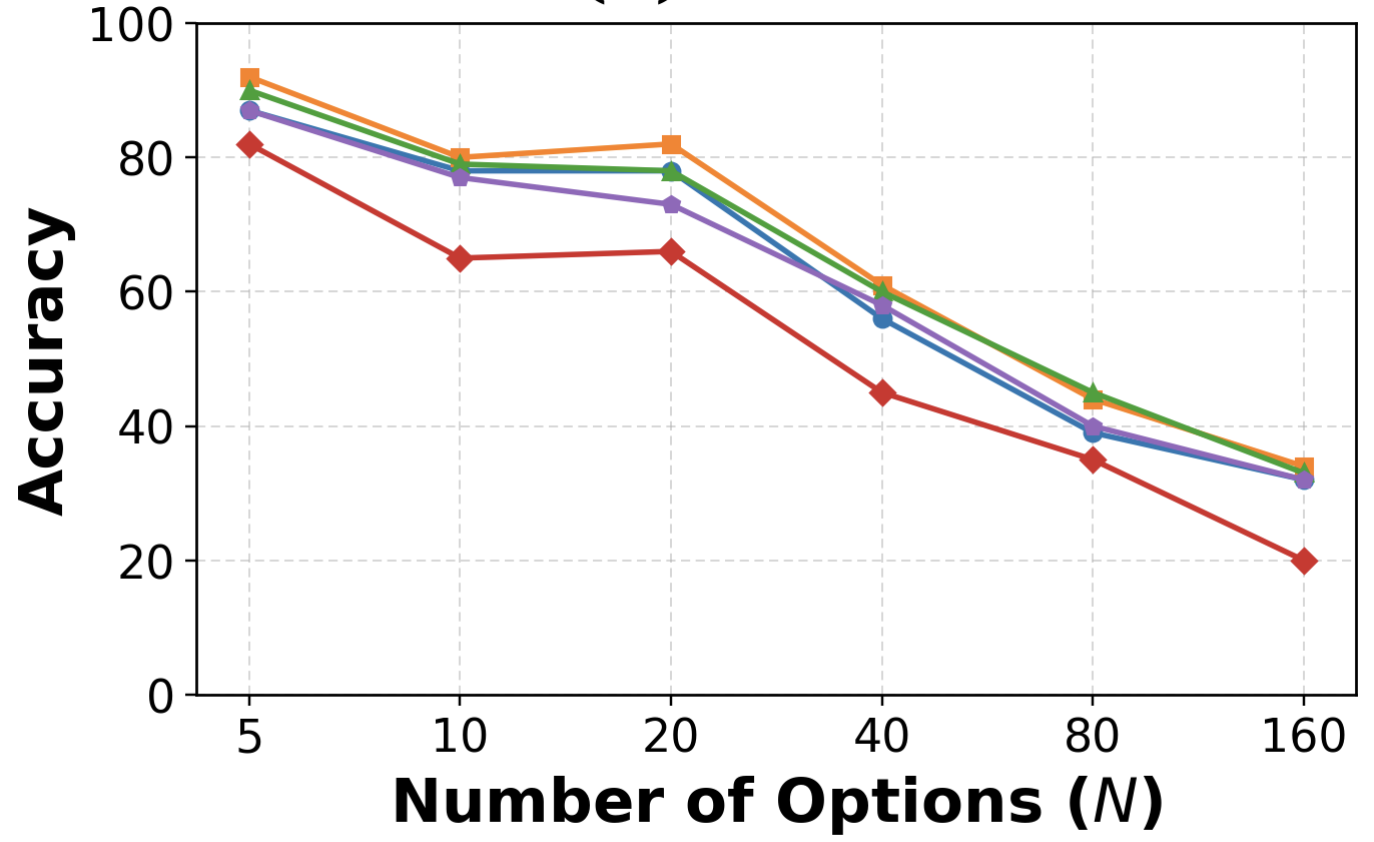}
        \caption{MIMIC (Qwen3-4B)}
    \end{subfigure}

    \vspace{0.15em}
    \includegraphics[width=0.5\linewidth]{Figs/acc_system2_compare/legend_all.png}

    \caption{Additional System-2 reasoning results on Qwen3-4B.
    All reasoning strategies continue to degrade substantially as the number of options increases.}
    \label{fig:system2_qwen_appendix}
\end{figure*}

\textbf{Additional Results.} We provide additional System-2 reasoning results on Qwen3-4B in Figure~\ref{fig:system2_qwen_appendix}. Similar to the main-text observations on Llama-8B, all reasoning strategies continue to exhibit substantial degradation as the number of options increases, despite improving absolute accuracy in smaller candidate settings. 

These results further support our conclusion that increasing reasoning depth alone does not fundamentally resolve failures in scalable candidate comparison.

\subsection{Additional Robustness of Large-Option Degradation}
\label{appendix:accuracy_robustness}

\subsubsection{Statistical Robustness and Larger Evaluation Samples}
\label{appendix:statistical_robustness}

To assess whether the observed large-option degradation could be
explained by finite-sample variability, we conduct additional
bootstrap uncertainty analysis and larger-sample evaluation.

\paragraph{Bootstrap uncertainty.}
We compute 95\% bootstrap confidence intervals over evaluation
instances for the primary accuracy experiments on HotpotQA and
MIMIC. Across all model--dataset--candidate-size configurations,
the mean confidence-interval half-width is 4.88 percentage points
(4.17 on HotpotQA and 5.60 on MIMIC), with a maximum half-width
of 7.00 points. These uncertainty estimates are substantially smaller
than the observed degradation under candidate-set scaling, which
frequently exceeds 20 percentage points and reaches substantially
larger values in several settings. For example, Phi-4-mini decreases
from 83.0\% to 19.5\% on HotpotQA and from 85.0\% to 9.0\% on
MIMIC between $N=5$ and $N=160$.

\paragraph{Larger evaluation sample.}
We further repeat the HotpotQA evaluation using 1,000 instances to
test whether the degradation persists with a substantially larger
evaluation sample. As shown in Table~\ref{tab:large_sample_robustness},
all evaluated models continue to exhibit degradation as the candidate
set grows from $N=5$ to $N=160$. Llama-3.1-8B decreases from 95.3\%
to 62.2\%, Phi-4-mini from 88.0\% to 28.7\%, and Qwen3-30B
from 99.3\% to 91.2\%. Across the 18 model--candidate-size
configurations in this larger-sample evaluation, the mean 95\%
bootstrap confidence-interval half-width decreases to 1.85 percentage
points, with a maximum of 3.10 points.

These results indicate that the observed large-option degradation is
substantially larger than the uncertainty attributable to finite
evaluation samples and persists when the evaluation set is increased
fivefold.

\begin{table}[t]
\caption{Accuracy degradation with 1,000 HotpotQA evaluation instances.
All values are accuracy (\%).}
\label{tab:large_sample_robustness}
\centering
\small
\setlength{\tabcolsep}{4pt}
\begin{tabular}{lccc}
\toprule
Model & $N=5$ & $N=160$ & $\Delta$ \\
\midrule
Phi-4-mini       & 88.0 & 28.7 & $-59.3$ \\
Llama-3.1-8B     & 95.3 & 62.2 & $-33.1$ \\
Qwen3-30B    & 99.3 & 91.2 & $-8.1$ \\
\bottomrule
\end{tabular}
\end{table}

\subsubsection{Extended Candidate-Set Scaling and Closed-Source Models}
\label{appendix:extended_scaling}

To examine whether large-option degradation persists beyond the candidate-set
sizes considered in our primary experiments, we extend the MIMIC
evaluation from $N=160$ to $N=320$ and $N=640$. We additionally evaluate
GPT-5.4 to test whether the observed degradation also occurs in a strong
closed-source model.

\begin{figure*}[t]
    \centering

    \includegraphics[width=0.72\linewidth]
    {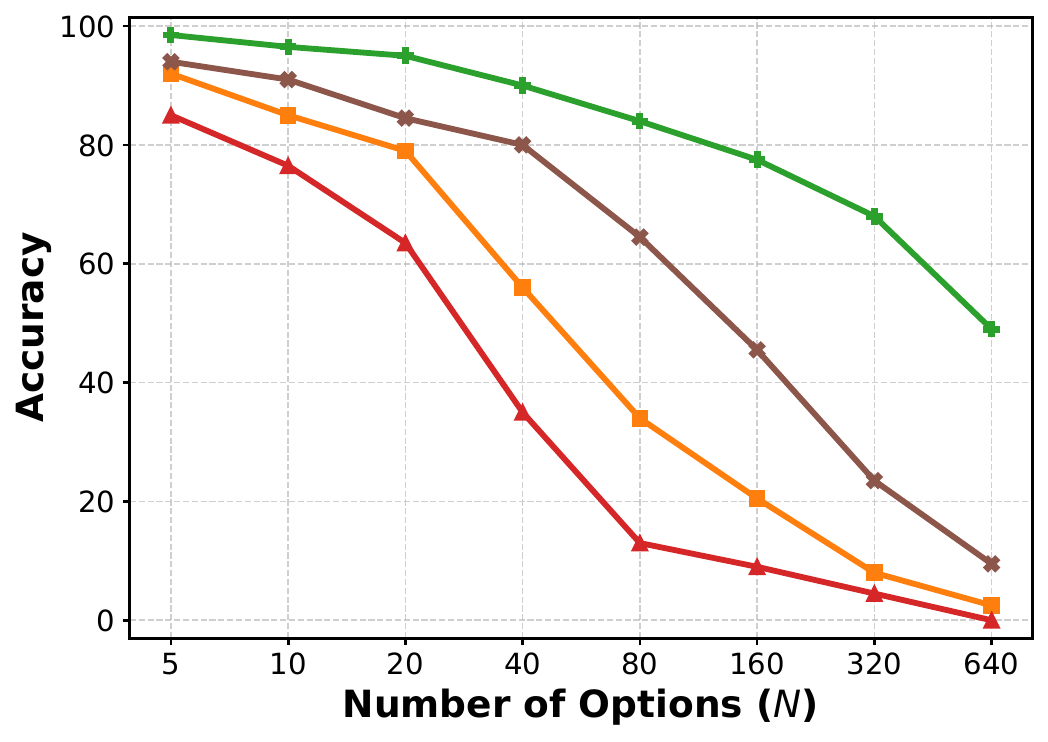}

    \vspace{0.2em}

    \includegraphics[width=0.72\linewidth]
    {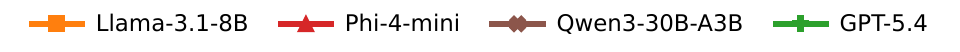}

    \caption{Extended candidate-set scaling on MIMIC.
    We extend the number of candidate options to $N=320$ and $N=640$
    and additionally evaluate GPT-5.4. Accuracy continues to decrease
    beyond the $N=160$ regime across all evaluated models. Although
    GPT-5.4 is substantially more robust than the open-source models,
    its accuracy decreases from 98.5\% at $N=5$ to 49.0\% at $N=640$.}
    \label{fig:extended_scaling_mimic}
\end{figure*}

As shown in Figure~\ref{fig:extended_scaling_mimic}, accuracy continues to
decrease as the candidate set expands beyond $N=160$ across all evaluated
models. The degradation is particularly pronounced for the open-source
models: at $N=640$, Llama-3.1-8B, Phi-4-mini, and Qwen3-30B reach
approximately 2\%, 0\%, and 9\% accuracy, respectively.

Importantly, GPT-5.4 exhibits substantially greater robustness at smaller
candidate-set sizes, maintaining 98.5\% accuracy at $N=5$ and 77.5\% at
$N=160$. Nevertheless, its performance further decreases to 68.0\% at
$N=320$ and 49.0\% at $N=640$. These results indicate that large-option
degradation is not restricted to the original $N\leq160$ regime or to the
open-source models considered in our main experiments. At the same time,
the substantially slower degradation of GPT-5.4 suggests that robustness
to candidate-set scaling varies considerably across models.

\subsection{Robustness of Gold-Margin Analysis}
\label{appendix:gold_margin_robustness}

Our main analysis computes candidate scores using summed token-level
log-probabilities. We examine whether the observed gold-margin collapse
depends on this score formulation using two alternative analyses on
HotpotQA.

\paragraph{Length-Normalized Scores.}
All generated answers follow the template
\texttt{<Answer> index </Answer>}, with output lengths varying only
slightly across answer indices (6--8 tokens). We nevertheless recompute
gold margins using length-normalized sequence scores. As shown in
Table~\ref{tab:gold_margin_robustness}, the normalized gold margin
continues to shrink substantially from $N=5$ to $N=160$ across all three
models, becoming negative for Llama-3.1-8B and Phi-4-mini.

\paragraph{Probability-Space Analysis.}
We further compare the softmax probabilities assigned to the gold
candidate and the highest-probability distractor. As the candidate set
grows, gold probability decreases substantially across models, while the
strongest distractor probability increases. Importantly, the reduction
in gold confidence is substantial even under this alternative
representation. Together, these results show that gold-margin collapse
is robust to score normalization and is not an artifact of using summed
token-level log-probabilities.

\begin{table}[t]
\caption{
Robustness of gold-margin collapse under alternative score formulations on HotpotQA. Values report changes from $N=5$ to $N=160$.
}
\label{tab:gold_margin_robustness}
\centering
\small
\setlength{\tabcolsep}{3.5pt}
\begin{tabular}{lccc}
\toprule
Model &
Norm. GM &
$P(\mathrm{gold})$ &
$P(\mathrm{top\ dist.})$ \\
\midrule
Qwen3-30B
& $1.97 \rightarrow 0.80$
& $0.97 \rightarrow 0.89$
& $0.02 \rightarrow 0.08$ \\

Llama-3.1-8B
& $0.57 \rightarrow -0.10$
& $0.88 \rightarrow 0.44$
& $0.11 \rightarrow 0.22$ \\

Phi-4-mini
& $0.75 \rightarrow -0.45$
& $0.92 \rightarrow 0.16$
& $0.05 \rightarrow 0.35$ \\
\bottomrule
\end{tabular}
\end{table}

\subsection{Irrelevant Contents}
\label{sec:irrelevant_content}
For the context length experiments, we use a fixed template of neutral text (47 tokens) that is semantically unrelated to the task.
This text contains no information relevant to the question or answer choices, and is repeated to control the total input length.
The template is shown below.

\begin{quote}
This is additional background text used to increase the length of the input.
It does not contain any information relevant to the question or answer choices.
The purpose of this text is solely to vary the context length without affecting the decision process.
\end{quote}

\subsection{Distractor Verbosity Control}
\label{appendix:verbosity_control}

To further examine whether prompt length alone explains large-option degradation, we compare candidate-set expansion using distractor options with systematically different verbosity levels.

\begin{figure*}[t]
    \centering

    \begin{subfigure}[c]{0.40\linewidth}
        \centering
        \includegraphics[width=0.95\linewidth]{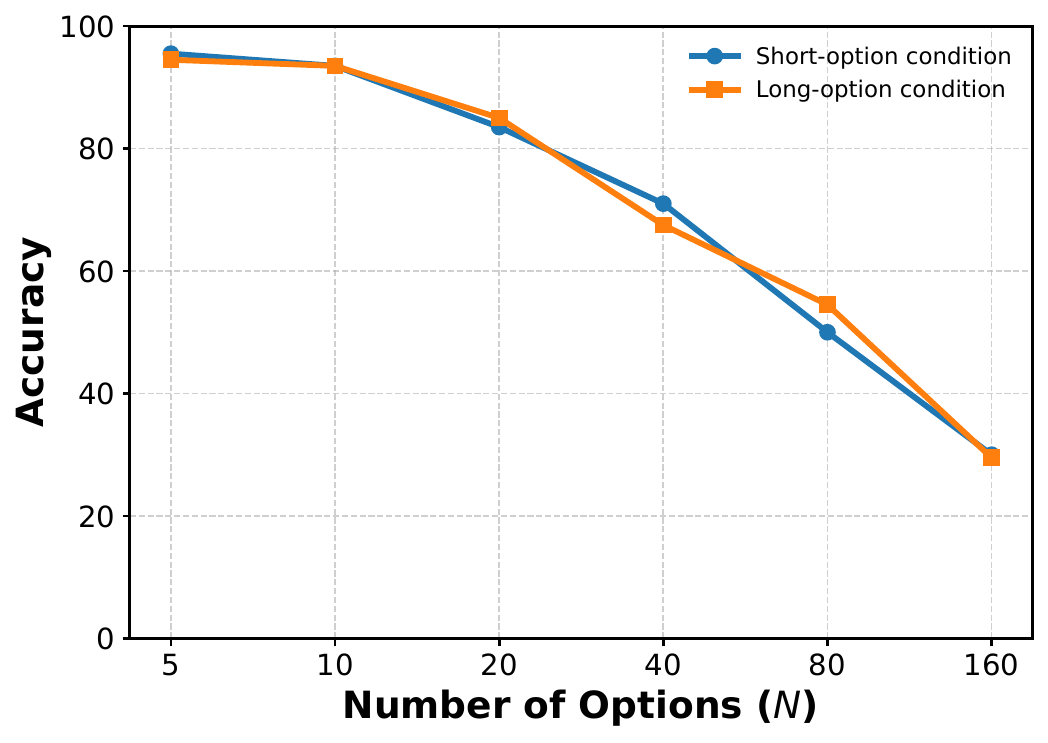}
        \caption{Verbosity control}
    \end{subfigure}
    \hspace{0.02\linewidth}
    \begin{subfigure}[c]{0.48\linewidth}
        \centering
        \vspace{1.2em}

        \small
        \begin{tabular}{lcc}
        \toprule
        $N$ & Short-option & Long-option \\
        \midrule
        5   & 3688  & 3866  \\
        10  & 3990  & 4382  \\
        20  & 4597  & 5426  \\
        40  & 5810  & 7518  \\
        80  & 8240  & 11693 \\
        160 & 13120 & 20075 \\
        \bottomrule
        \end{tabular}

        \vspace{0.5em}
        \caption{Average prompt token length}
    \end{subfigure}

    \caption{Distractor verbosity control under candidate-set expansion
    (MIMIC, Llama-3.1-8B).
    Although longer distractor options substantially increase total context length,
    they do not consistently reproduce the degradation induced by increasing the
    number of candidate options.}
    \label{fig:length_control_appendix}
\end{figure*}

Figure~\ref{fig:length_control_appendix} shows that although long-option distractors substantially increase average prompt length, the resulting degradation trends remain substantially less consistent than those induced by increasing the number of candidate options. These results further support that context length alone cannot fully explain large-option degradation.

\subsection{Attention-Prediction Discrepancies Across Candidate Sizes and Datasets}
\label{appendix:attention_generalization}

We further examine attention allocation and prediction-position distributions across different candidate-set sizes and datasets.

\begin{figure*}[t]
    \centering

    \begin{subfigure}[t]{0.48\linewidth}
        \centering
        \includegraphics[width=\linewidth]{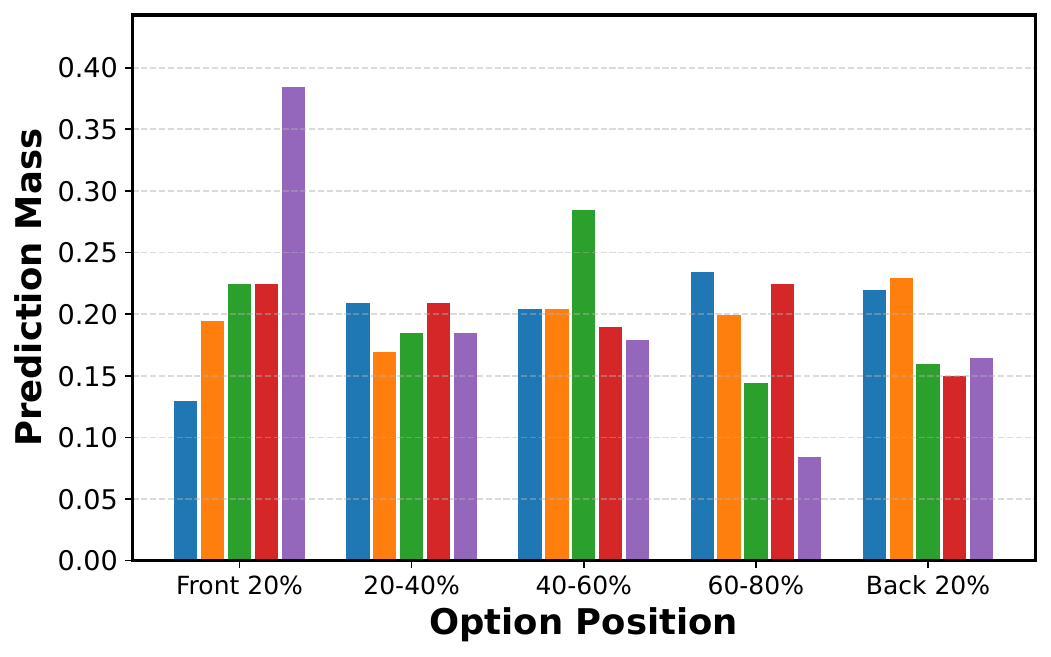}
        \caption{Llama-3.1-8B prediction}
    \end{subfigure}
    \hfill
    \begin{subfigure}[t]{0.48\linewidth}
        \centering
        \includegraphics[width=\linewidth]{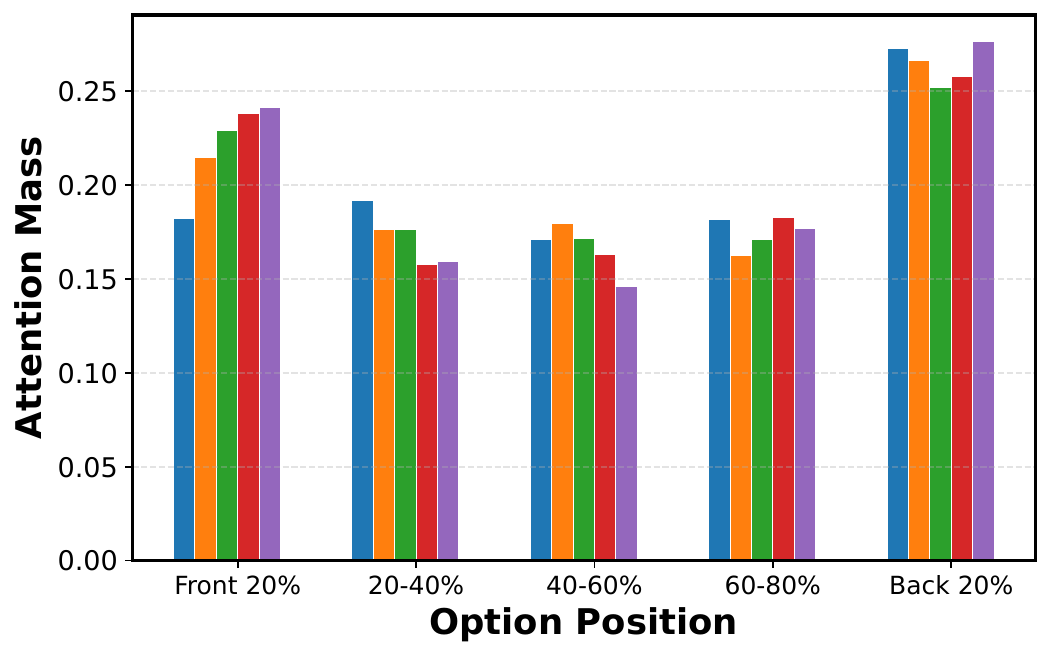}
        \caption{Llama-3.1-8B attention}
    \end{subfigure}

    \vspace{0.4em}

    \begin{subfigure}[t]{0.48\linewidth}
        \centering
        \includegraphics[width=\linewidth]{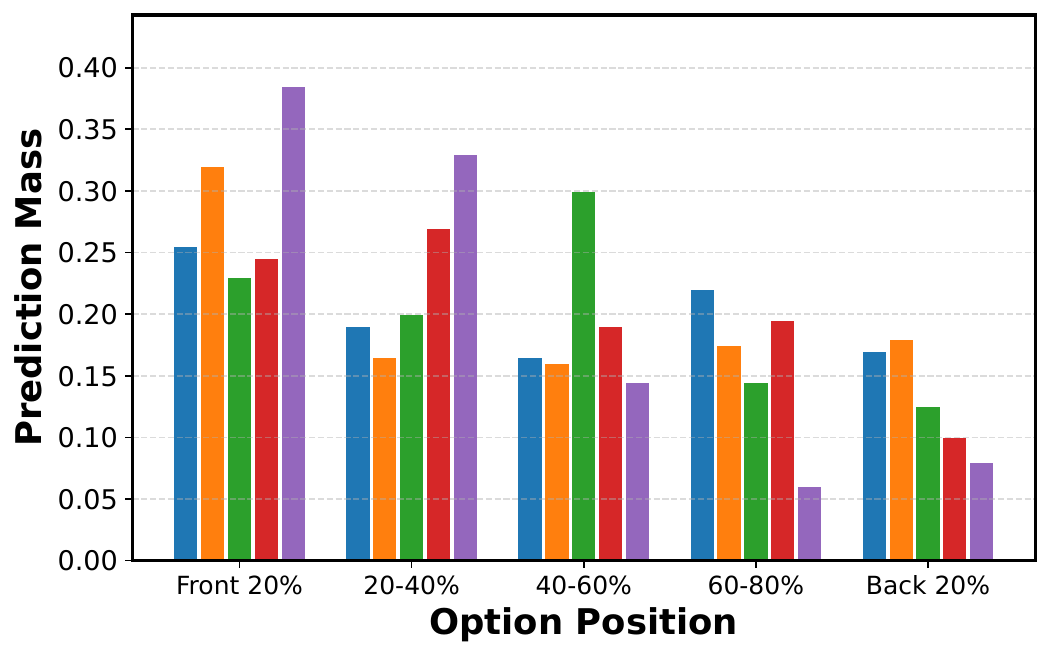}
        \caption{Phi-4-mini prediction}
    \end{subfigure}
    \hfill
    \begin{subfigure}[t]{0.48\linewidth}
        \centering
        \includegraphics[width=\linewidth]{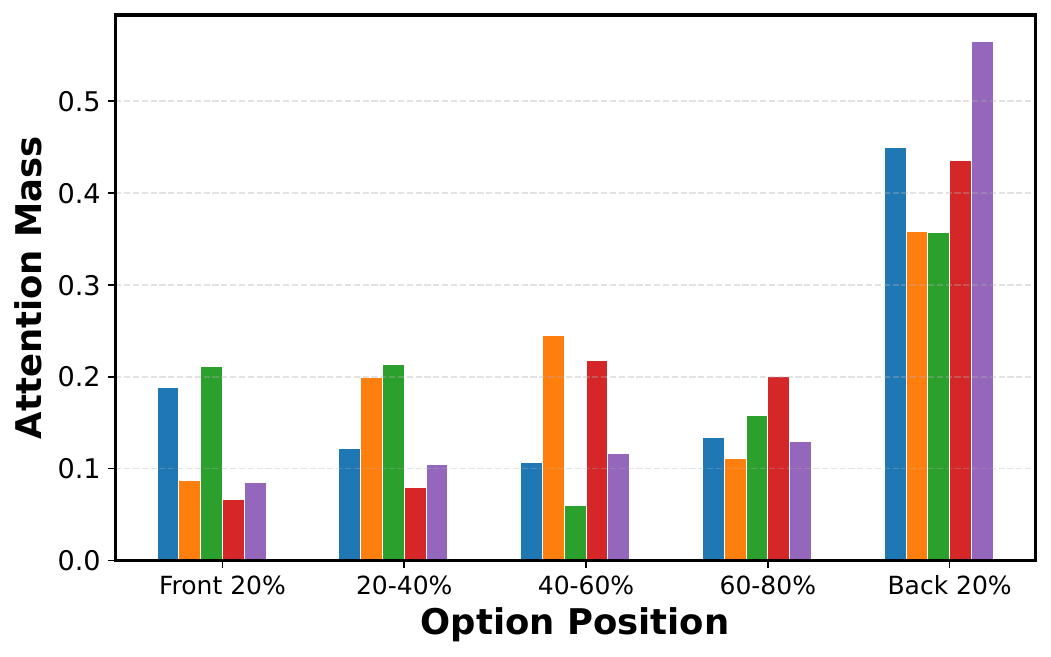}
        \caption{Phi-4-mini attention}
    \end{subfigure}

    \vspace{0.5em}

    \includegraphics[width=0.55\linewidth]{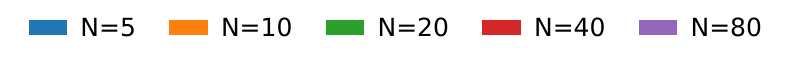}

    \caption{Prediction-position and attention distributions across candidate-set sizes on HotpotQA.
    Prediction distributions become increasingly concentrated toward earlier candidate positions as the number of options increases. However, attention allocation patterns vary substantially across models and candidate-set sizes, with several settings exhibiting increased attention concentration toward later option positions.}
    \label{fig:attn_pred_hotpot_multiN}
\end{figure*}

\begin{figure*}[t]
    \centering

    \begin{subfigure}[t]{0.48\linewidth}
        \centering
        \includegraphics[width=\linewidth]{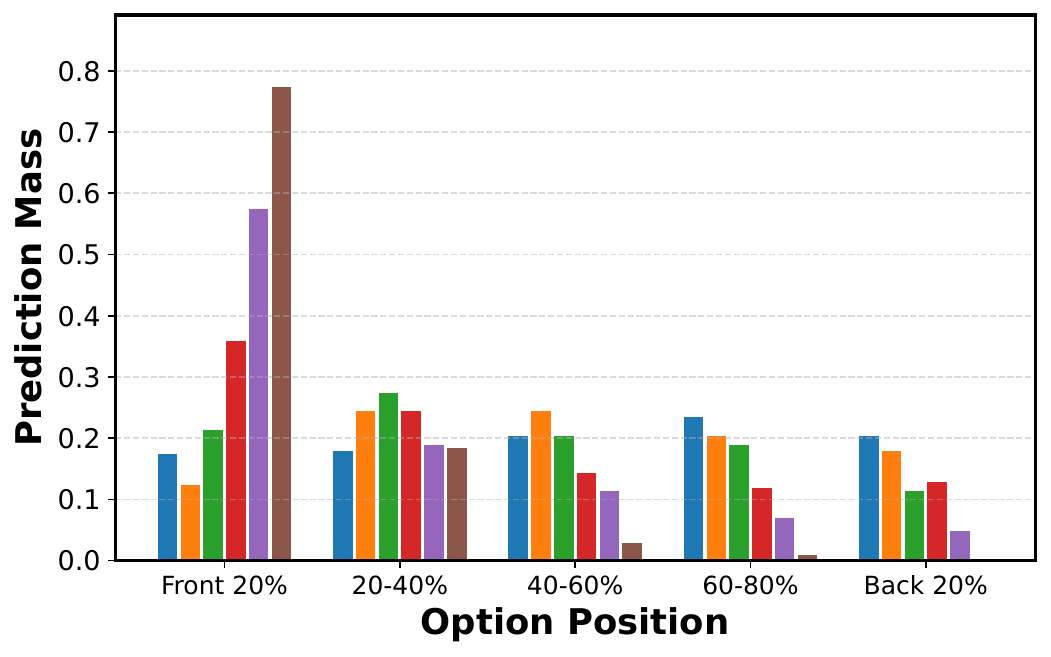}
        \caption{Llama-3.1-8B prediction}
    \end{subfigure}
    \hfill
    \begin{subfigure}[t]{0.48\linewidth}
        \centering
        \includegraphics[width=\linewidth]{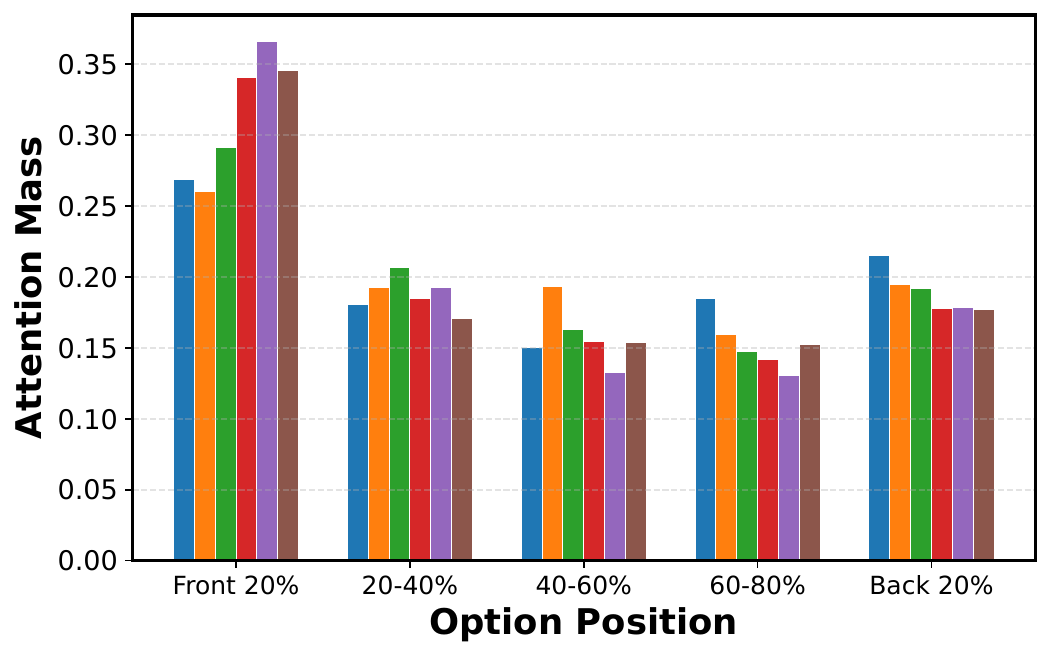}
        \caption{Llama-3.1-8B attention}
    \end{subfigure}

    \vspace{0.45em}

    \begin{subfigure}[t]{0.48\linewidth}
        \centering
        \includegraphics[width=\linewidth]{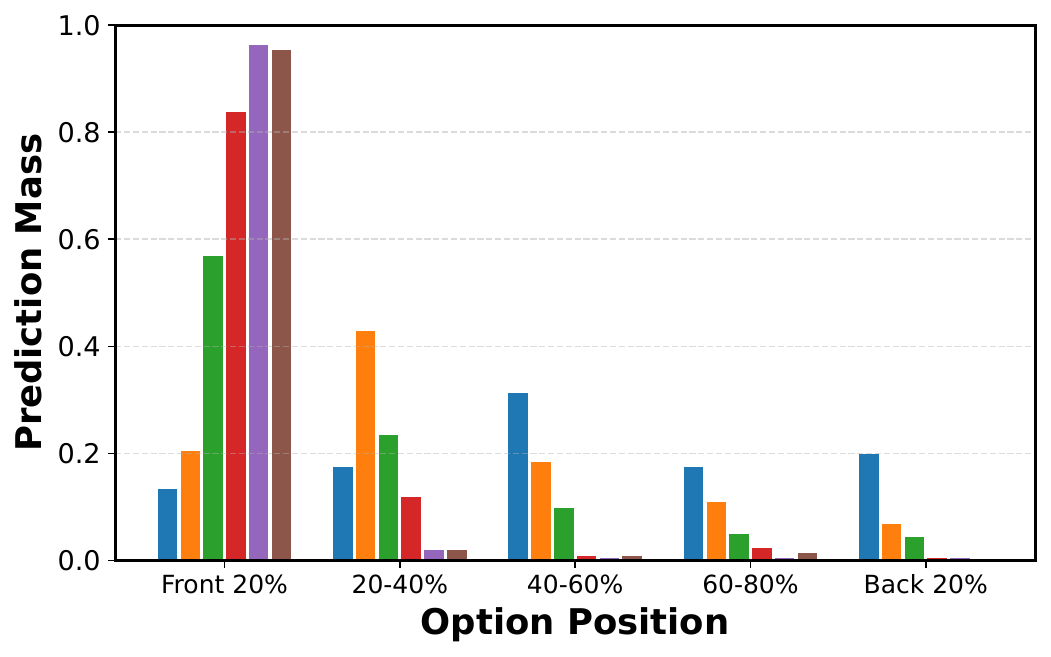}
        \caption{Phi-4-mini prediction}
    \end{subfigure}
    \hfill
    \begin{subfigure}[t]{0.48\linewidth}
        \centering
        \includegraphics[width=\linewidth]{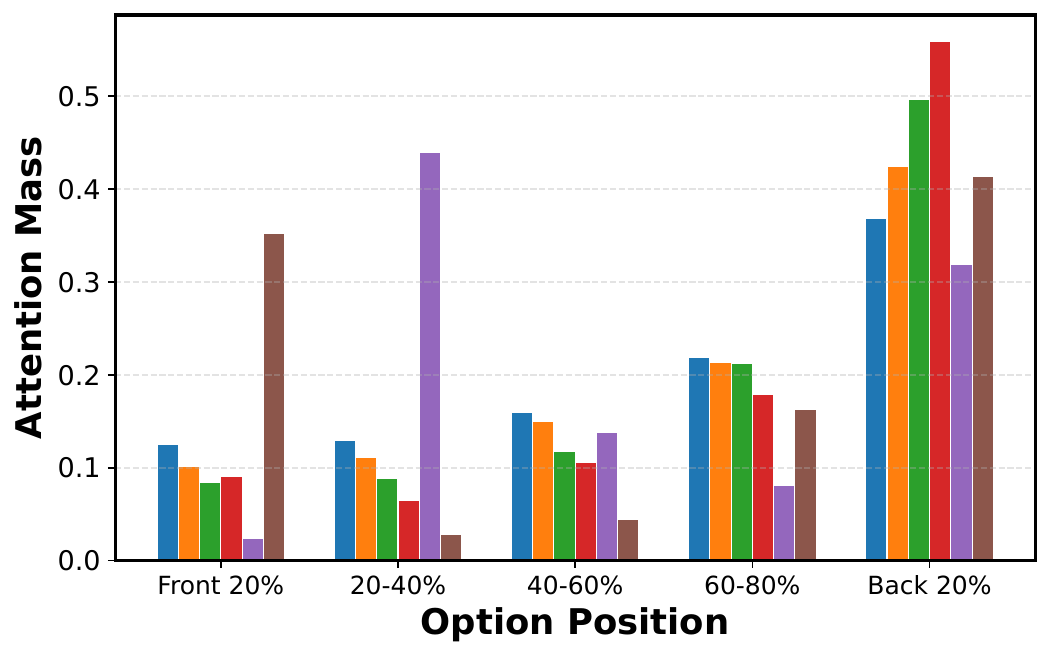}
        \caption{Phi-4-mini attention}
    \end{subfigure}

    \vspace{0.45em}

    \begin{subfigure}[t]{0.48\linewidth}
        \centering
        \includegraphics[width=\linewidth]{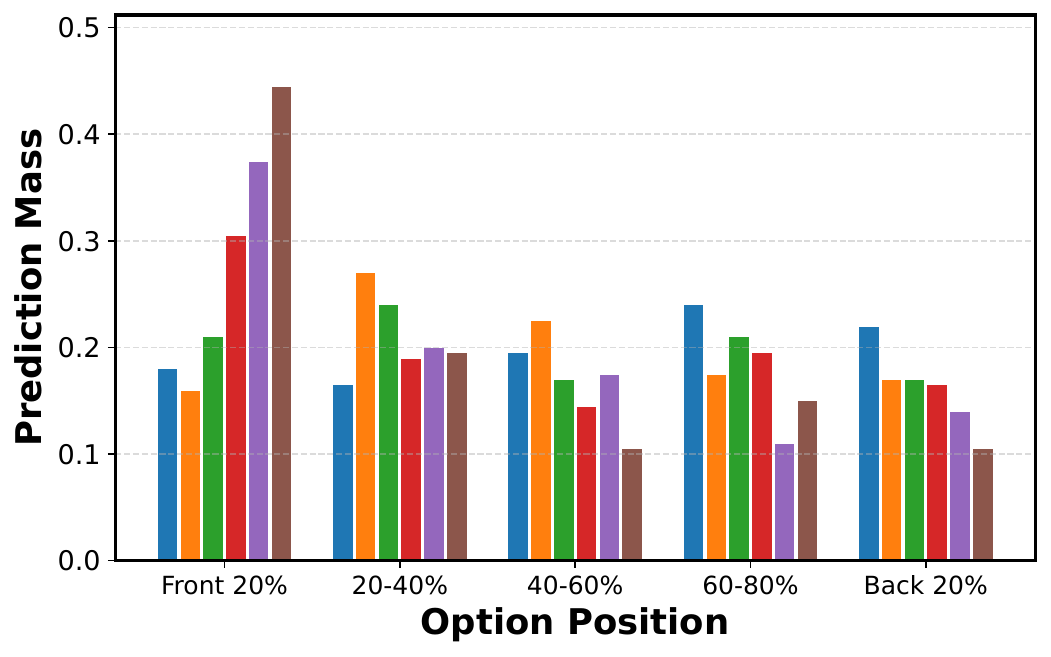}
        \caption{Qwen3-30B prediction}
    \end{subfigure}
    \hfill
    \begin{subfigure}[t]{0.48\linewidth}
        \centering
        \includegraphics[width=\linewidth]{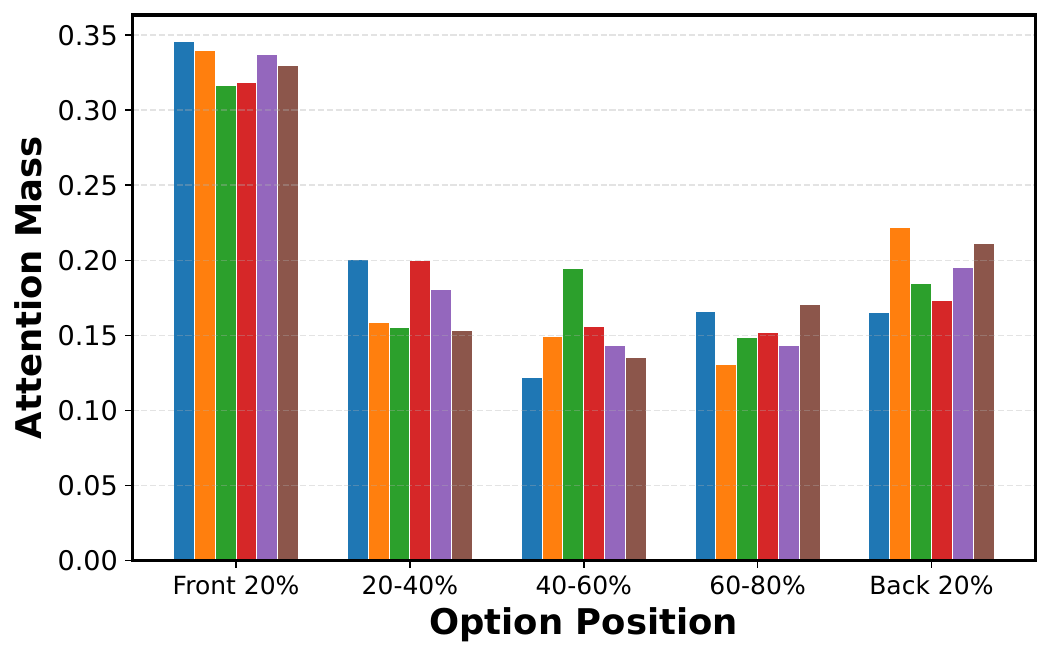}
        \caption{Qwen3-30B attention}
    \end{subfigure}

    \vspace{0.6em}

    \includegraphics[width=0.55\linewidth]{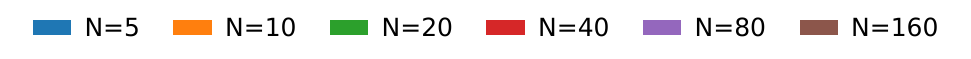}

    \caption{Prediction-position and attention distributions across candidate-set sizes on ArXiv.
    Prediction distributions remain consistently concentrated toward earlier candidate positions across models and candidate-set sizes, despite substantial variation in attention allocation patterns.}
    \label{fig:attn_pred_arxiv_multiN}
\end{figure*}

Figures~\ref{fig:attn_pred_hotpot_multiN} and~\ref{fig:attn_pred_arxiv_multiN} show that attention distributions vary substantially across models, datasets, and candidate-set sizes. In several settings, models allocate increasing attention mass toward later option positions as the number of candidates increases. However, prediction distributions remain consistently front-skewed even under these changing attention patterns.

These observations further support that prediction behavior under large-option decision making cannot be directly explained by attention allocation patterns alone.

\subsection{Prediction Positional Bias Analysis}
\label{sec:appendix-pos_bias_generalization}

We further provide additional analyses on prediction-position bias under large candidate sets, including average prediction-position trends, fine-grained position distributions, and reversed-index controls.

\begin{figure*}[t]
    \centering

    \begin{subfigure}[t]{0.48\linewidth}
        \centering
        \includegraphics[width=\linewidth]{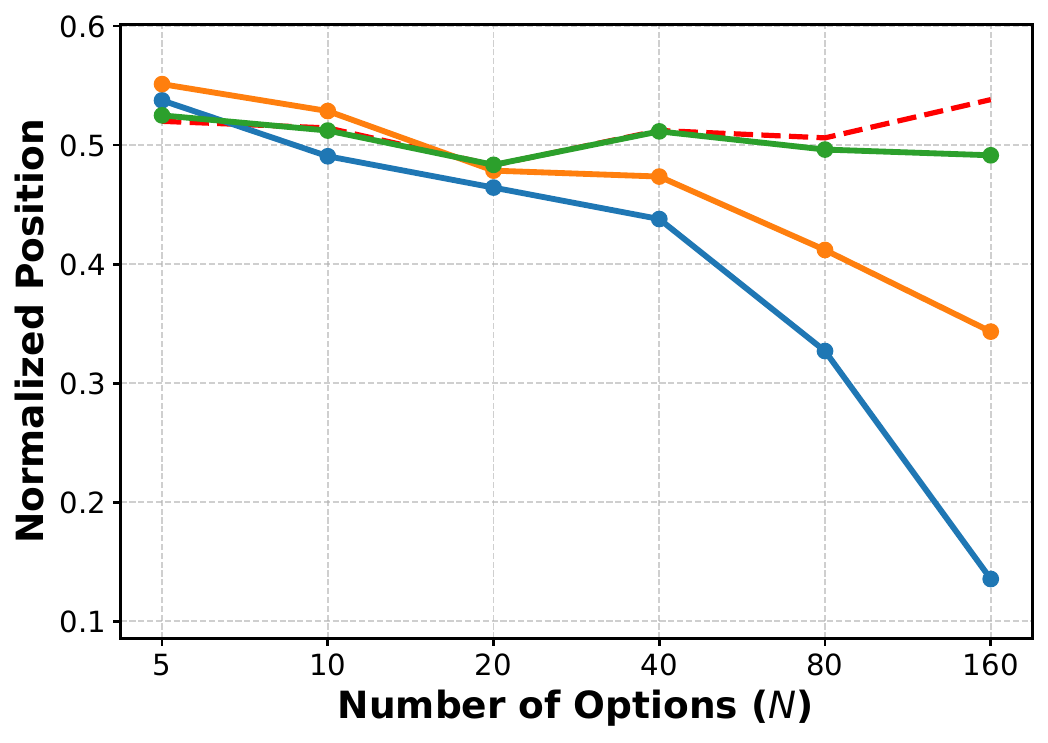}
        \caption{HotpotQA}
    \end{subfigure}
    \hfill
    \begin{subfigure}[t]{0.48\linewidth}
        \centering
        \includegraphics[width=\linewidth]{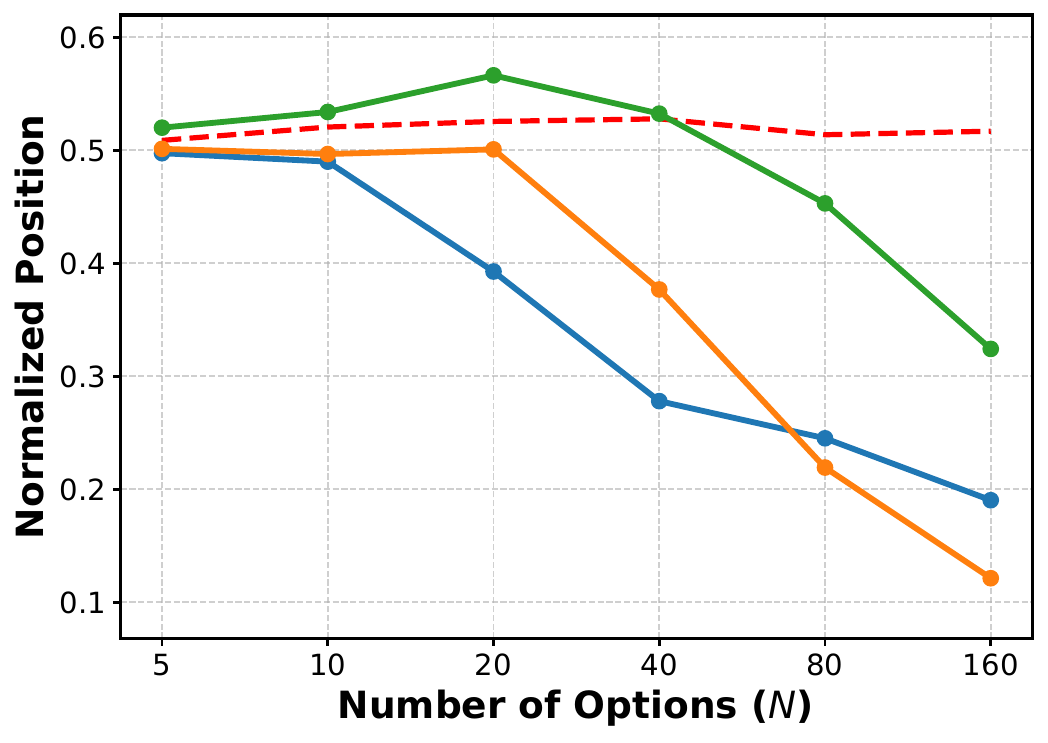}
        \caption{MIMIC}
    \end{subfigure}

    \vspace{0.15em}

    \includegraphics[width=0.42\linewidth]{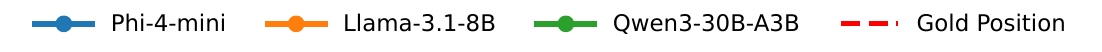}

    \caption{Average prediction position under candidate-set scaling.
    As the number of candidate options increases, model predictions progressively shift toward earlier candidate positions across datasets and model families, while the gold-answer distribution remains stable.}
    \label{fig:position_bias_combined_appendix}
\end{figure*}

Figure~\ref{fig:position_bias_combined_appendix} shows that model predictions progressively shift toward earlier candidate positions as the number of candidate options increases, despite uniformly distributed gold-answer positions. This trend remains consistent across datasets and model families, indicating increasingly strong early-position preference under large-option settings.

To further quantify this effect, Table~\ref{tab:five_bin_160_combined} reports five-bin prediction distributions at $N=160$. Predictions remain heavily concentrated toward earlier candidate regions for most models, with smaller models exhibiting particularly strong front-position concentration.

\vspace{0.3em}

\begin{table*}[t]
\caption{Five-bin prediction-position distributions at $N=160$.
Values indicate the proportion of predictions falling into each position region (front to back).}
\label{tab:five_bin_160_combined}

\centering
\small
\renewcommand{\arraystretch}{1.08}
\setlength{\tabcolsep}{5pt}

\resizebox{\linewidth}{!}{
\begin{tabular}{lcc}
\toprule
\multirow{2}{*}{Model} & \multicolumn{2}{c}{$N=160$} \\
\cmidrule(lr){2-3}
 & HotpotQA & MIMIC \\
\midrule
Phi-4-mini 
& 0.81 / 0.10 / 0.02 / 0.01 / 0.01 
& 0.69 / 0.14 / 0.06 / 0.06 / 0.03 \\

Llama-3.1-8B 
& 0.48 / 0.14 / 0.10 / 0.13 / 0.13 
& 0.81 / 0.13 / 0.04 / 0.01 / 0.01 \\

Qwen3-30B
& 0.26 / 0.15 / 0.21 / 0.20 / 0.20 
& 0.46 / 0.19 / 0.14 / 0.14 / 0.09 \\
\bottomrule
\end{tabular}
}
\end{table*}

We further perform a reversed-index control experiment to distinguish positional ordering effects from potential token-level index priors.

\begin{figure*}[t]
    \centering

    \begin{subfigure}[t]{0.44\linewidth}
        \centering
        \includegraphics[width=\linewidth]{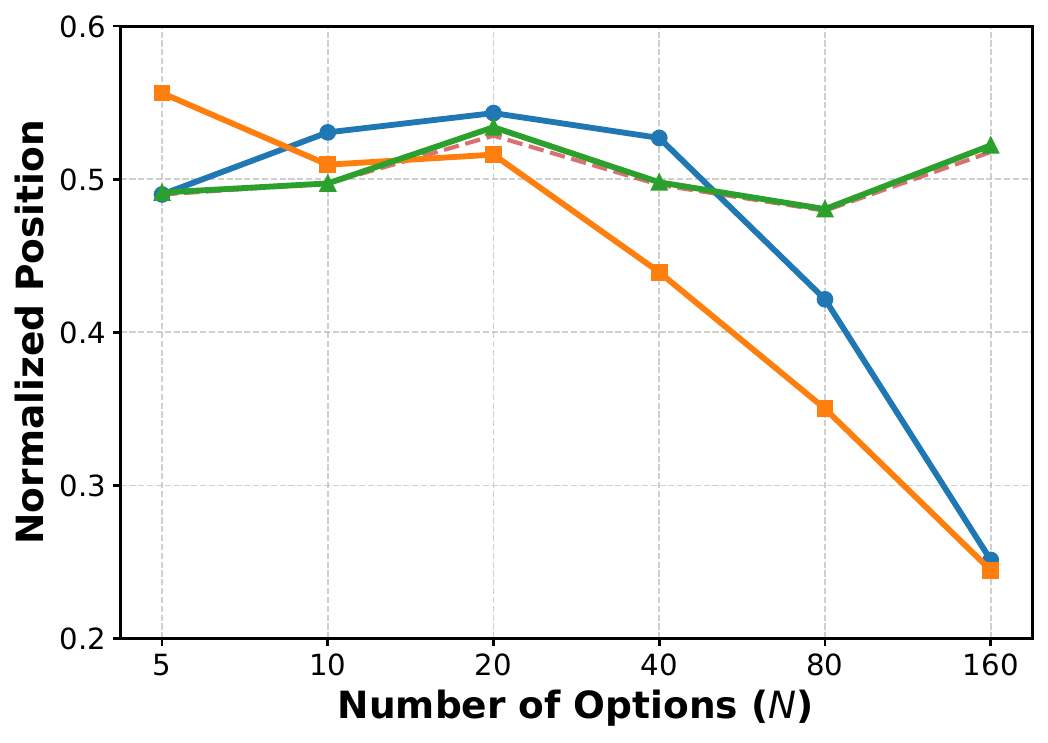}
        \caption{HotpotQA (reverse index)}
    \end{subfigure}
    \hfill
    \begin{subfigure}[t]{0.44\linewidth}
        \centering
        \includegraphics[width=\linewidth]{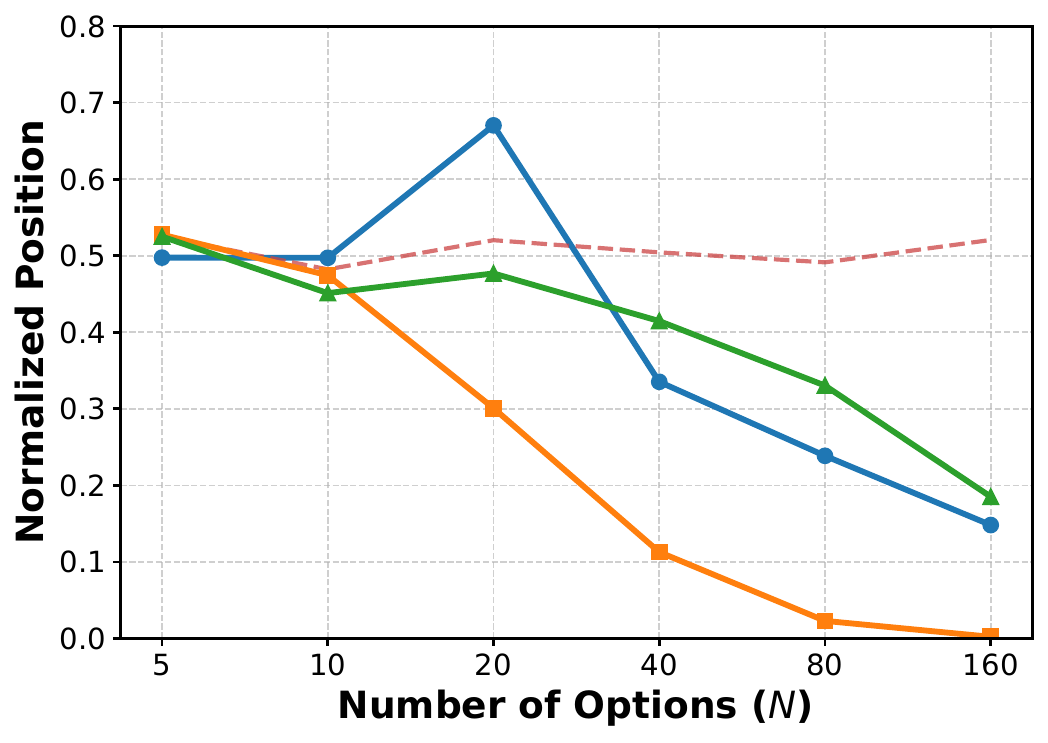}
        \caption{MIMIC (reverse index)}
    \end{subfigure}

    \vspace{0.3em}

    \includegraphics[width=0.5\linewidth]{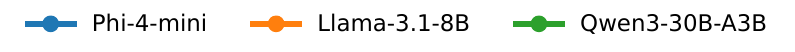}

    \caption{Early-position bias persists under reversed index assignment.
    Even when smaller answer indices are assigned to later candidates, model predictions continue to increasingly concentrate toward earlier candidate positions as the number of options grows.}
    \label{fig:reverse_index_bias_appendix}
\end{figure*}

Under this setting, earlier candidates are assigned larger index tokens while later candidates receive smaller indices. Figure~\ref{fig:reverse_index_bias_appendix} shows that similar early-position trends persist even under reversed indexing across both HotpotQA and MIMIC.

These findings suggest that the observed positional bias is driven primarily by candidate ordering rather than superficial preferences for specific answer indices.

\subsection{Additional Probability Dynamics under Controlled Gold Insertion}
\label{sec:appendix-additional-probability-dynamics}

\begin{figure*}[t]
    \centering

    \begin{subfigure}[t]{0.28\linewidth}
        \centering
        \includegraphics[width=\linewidth]{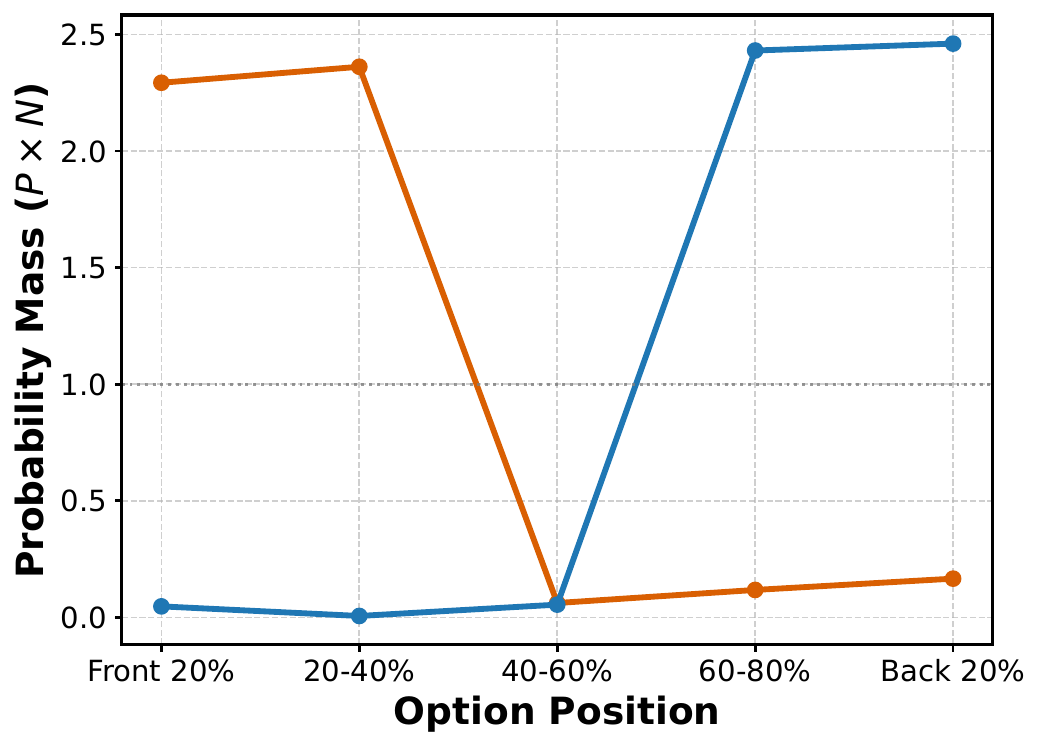}
        \caption{$N=5$}
    \end{subfigure}
    \hfill
    \begin{subfigure}[t]{0.28\linewidth}
        \centering
        \includegraphics[width=\linewidth]{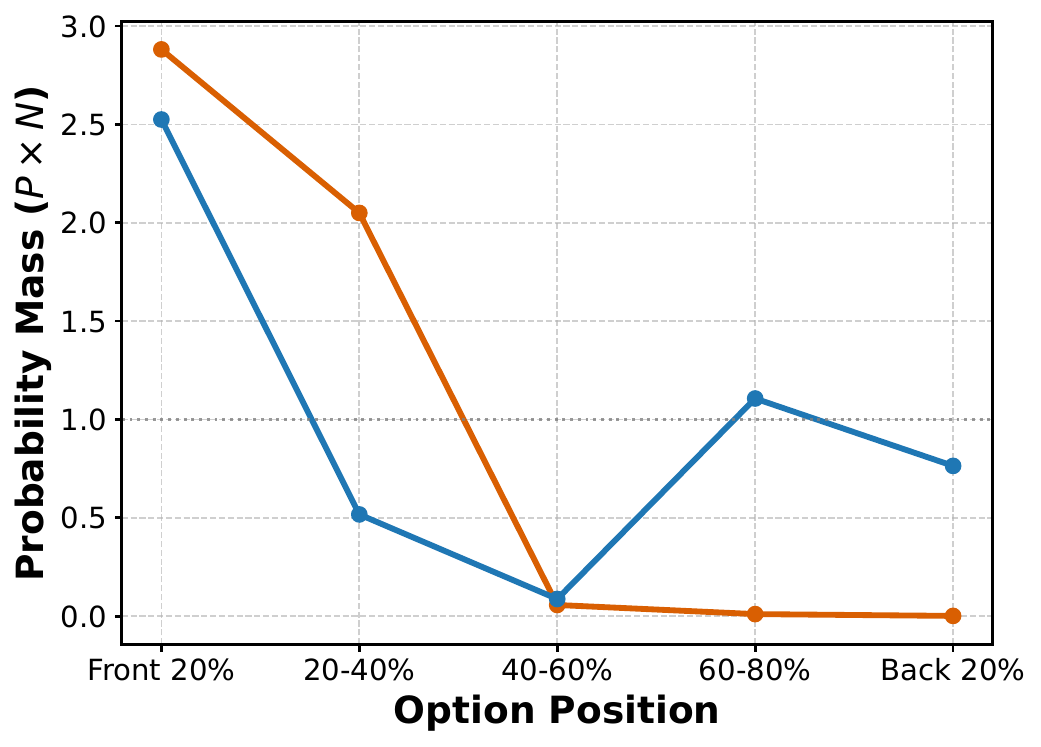}
        \caption{$N=80$}
    \end{subfigure}
    \hfill
    \begin{subfigure}[t]{0.28\linewidth}
        \centering
        \includegraphics[width=\linewidth]{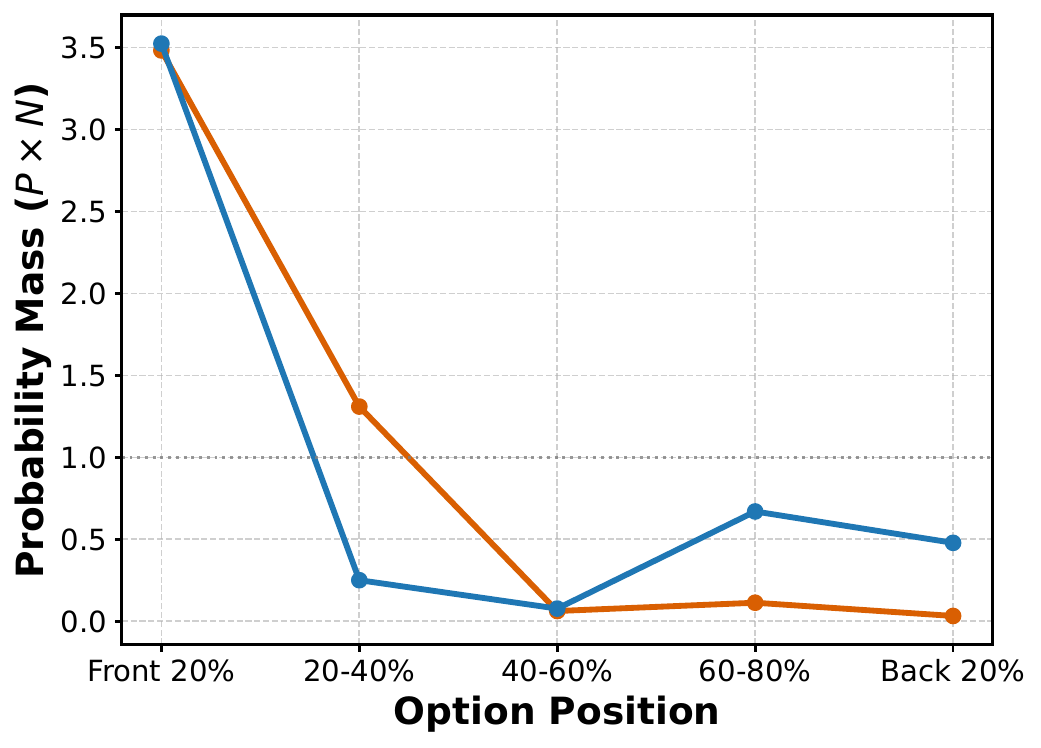}
        \caption{$N=160$}
    \end{subfigure}

    \vspace{0.1em}
    \includegraphics[width=0.35\linewidth]{Figs/gold_group_acc_main/legend_only.pdf}

    \caption{Probability distributions under controlled gold-position intervention.
    Probability mass progressively concentrates toward earlier candidates as the option set grows.}
    \label{fig:appendix_gold_position_prob}
\end{figure*}

Figure~\ref{fig:appendix_gold_position_prob} further visualizes the corresponding probability distributions under controlled gold-position intervention. As the number of options increases, prediction mass becomes progressively less responsive to the gold insertion region and increasingly concentrates toward earlier candidates.

\subsection{Early-Option Dominance Generalization}
\label{sec:appendix-early_option_generalization}

We further evaluate whether the early-option dominance generalizes across tasks and models using the arXiv dataset.

\begin{figure*}[t]
    \centering

    \begin{subfigure}[t]{0.44\linewidth}
        \centering
        \includegraphics[width=\linewidth]
        {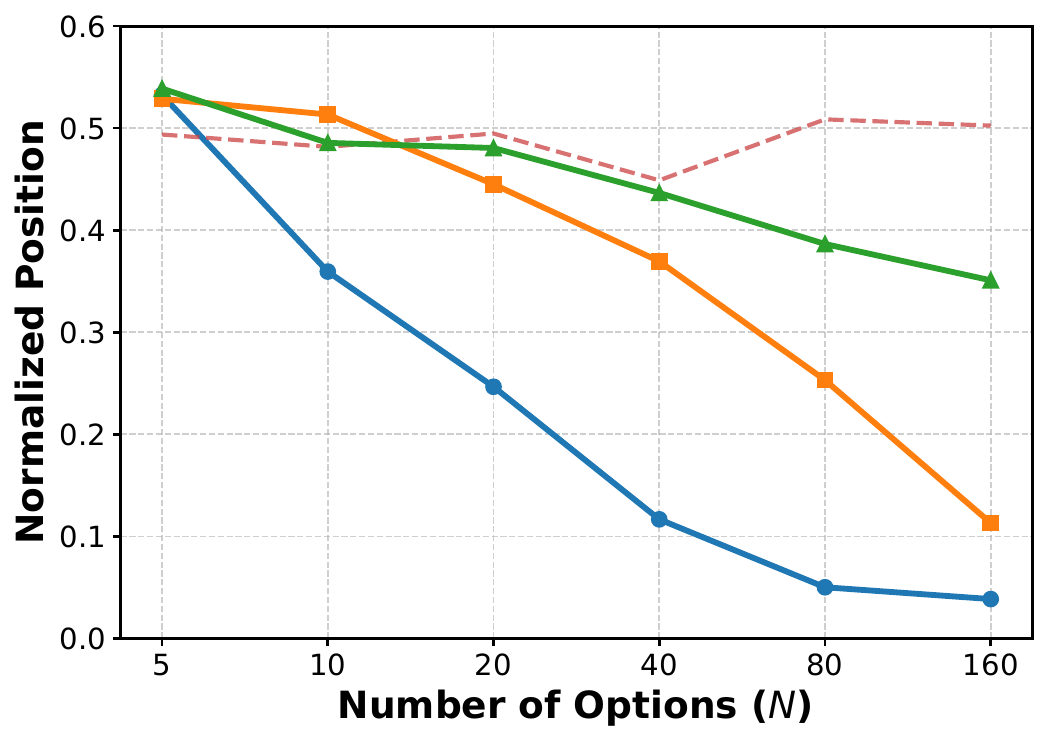}
        \caption{Original indexing}
    \end{subfigure}
    \hfill
    \begin{subfigure}[t]{0.44\linewidth}
        \centering
        \includegraphics[width=\linewidth]
        {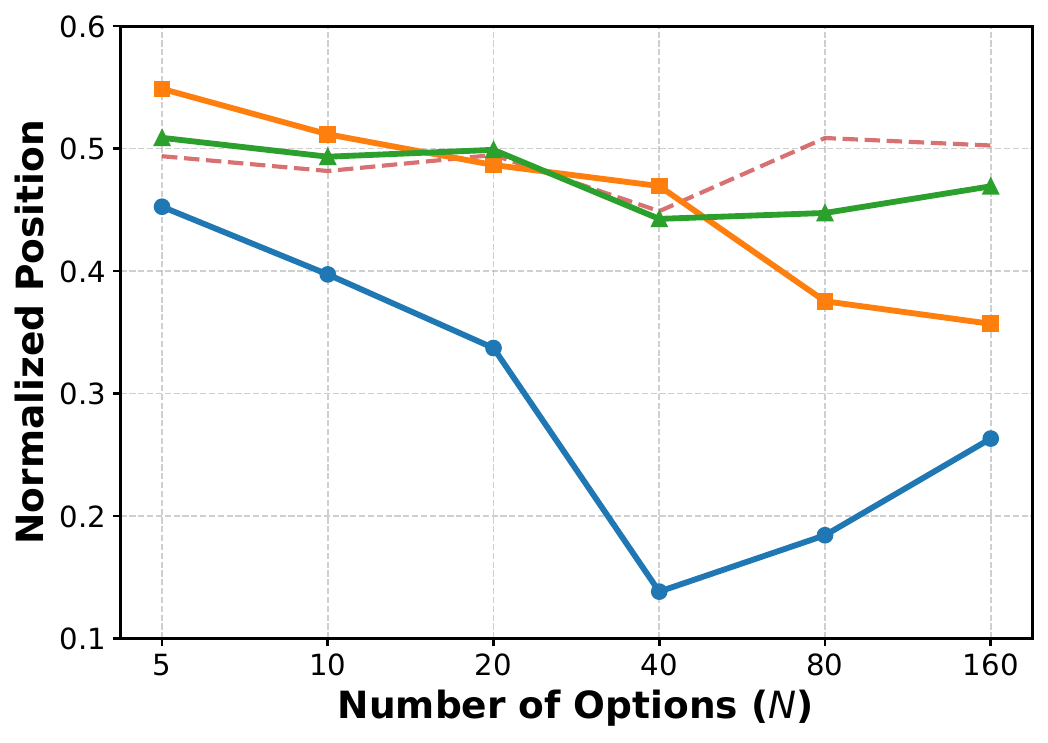}
        \caption{Reversed indexing}
    \end{subfigure}

    \vspace{0.3em}

    \includegraphics[width=0.5\linewidth]
    {Figs/pos_bias/hotpotqa/figures/legend_models_only.pdf}

    \caption{Early-position bias on arXiv under original and reversed indexing.
    Predictions progressively concentrate toward earlier candidate positions as the number of options increases across models. The same trend persists under reversed index assignment, indicating that the observed bias is driven primarily by the option position rather than token-level index priors.}
    \label{fig:appendix_arxiv_norm_position}
\end{figure*}

Figure~\ref{fig:appendix_arxiv_norm_position} shows that predictions progressively shift toward earlier candidate positions as the number of options increases across models. We additionally evaluate a reversed-index control on arXiv, where answer indices are assigned in reverse order while preserving candidate positions. The same early-position preference remains under reversed indexing, further confirming that the observed behavior is driven by candidate-order effects rather than superficial preferences for smaller answer indices.

\begin{figure*}[t]
    \centering

    \begin{subfigure}[t]{0.32\linewidth}
        \centering
        \includegraphics[width=\linewidth]
        {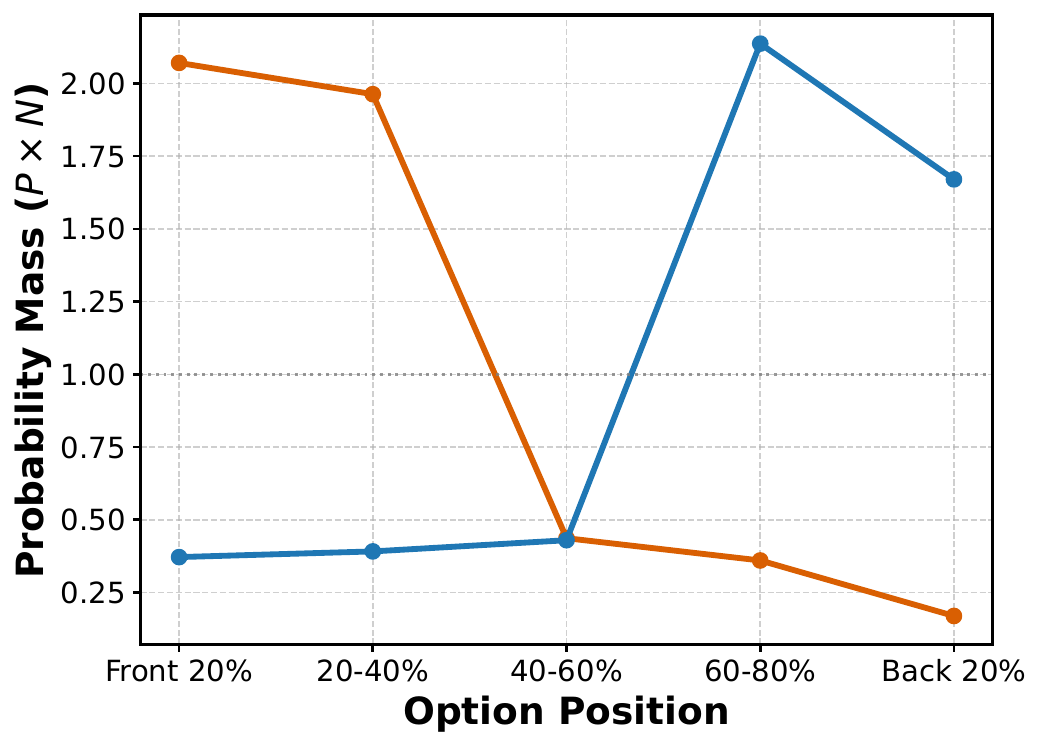}
        \caption{$N=5$}
    \end{subfigure}
    \hfill
    \begin{subfigure}[t]{0.32\linewidth}
        \centering
        \includegraphics[width=\linewidth]
        {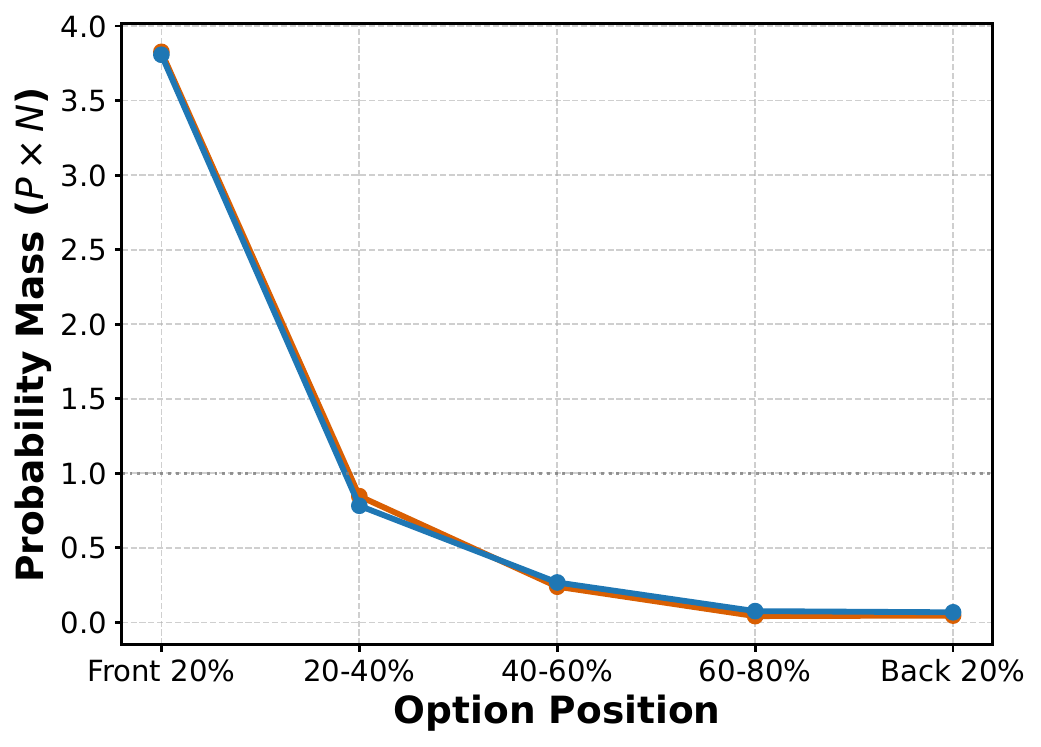}
        \caption{$N=80$}
    \end{subfigure}
    \hfill
    \begin{subfigure}[t]{0.32\linewidth}
        \centering
        \includegraphics[width=\linewidth]
        {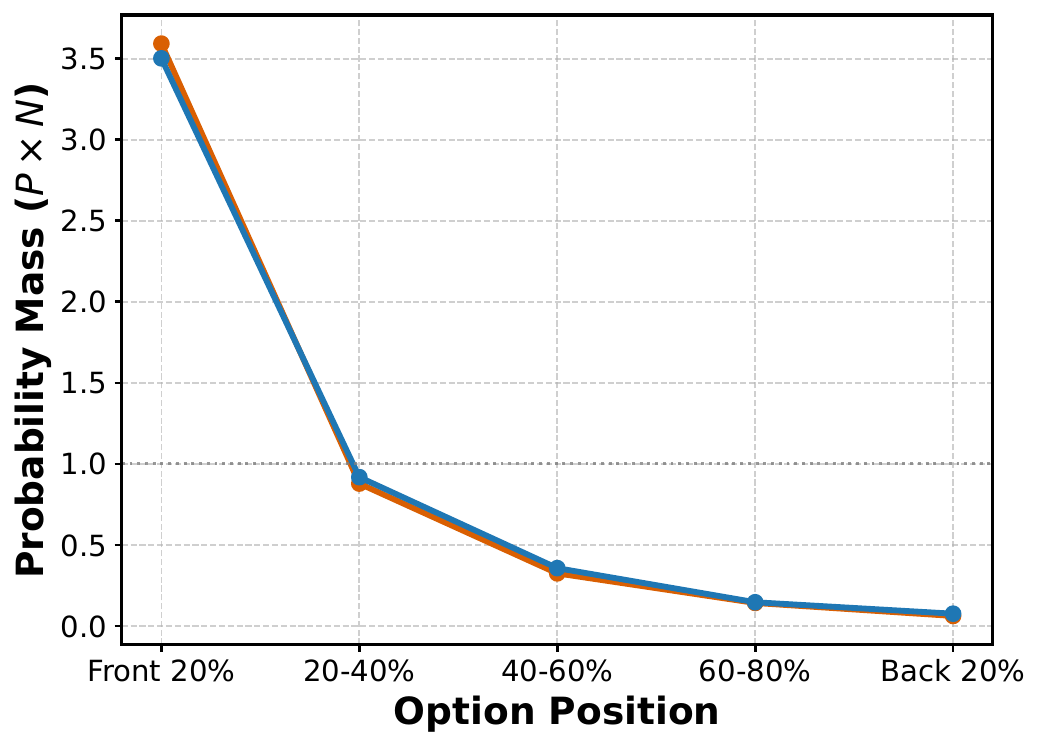}
        \caption{$N=160$}
    \end{subfigure}

    \vspace{0.2em}

    {\small \textbf{Phi-4-mini}}

    \vspace{0.6em}

    \begin{subfigure}[t]{0.32\linewidth}
        \centering
        \includegraphics[width=\linewidth]
        {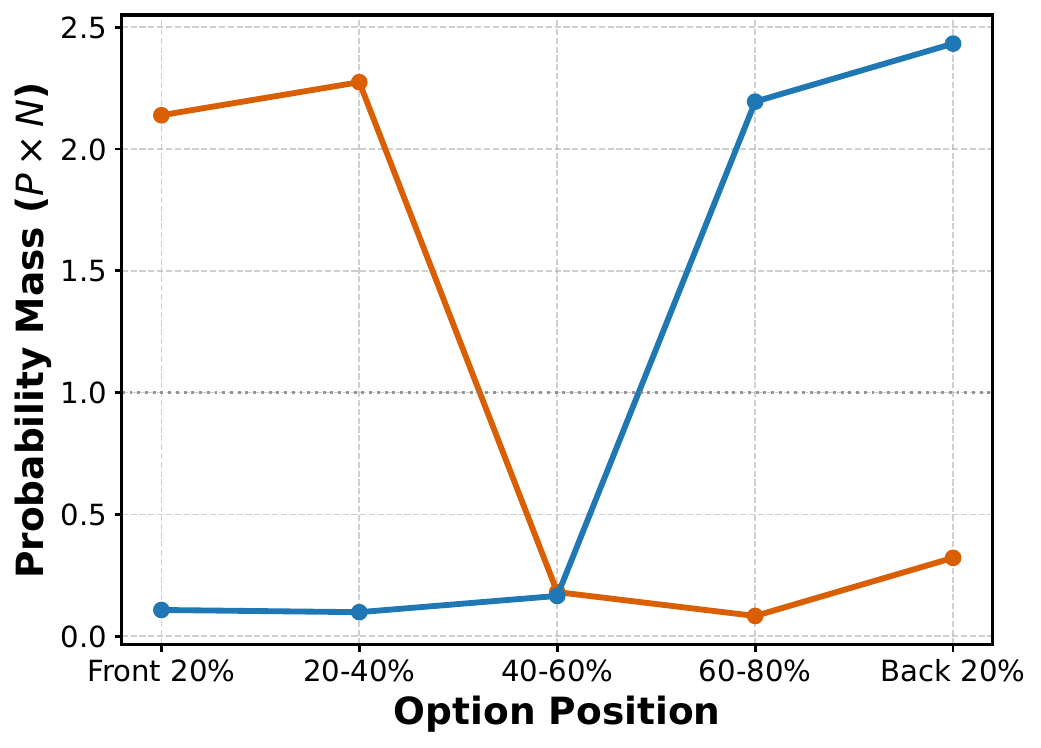}
        \caption{$N=5$}
    \end{subfigure}
    \hfill
    \begin{subfigure}[t]{0.32\linewidth}
        \centering
        \includegraphics[width=\linewidth]
        {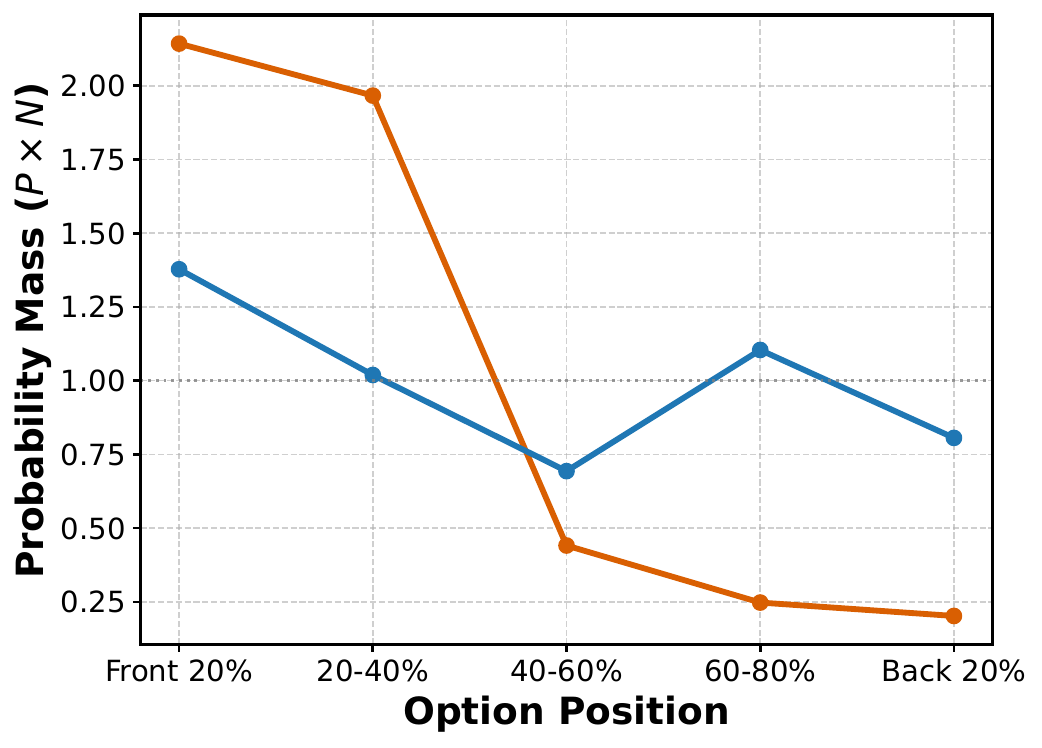}
        \caption{$N=80$}
    \end{subfigure}
    \hfill
    \begin{subfigure}[t]{0.32\linewidth}
        \centering
        \includegraphics[width=\linewidth]
        {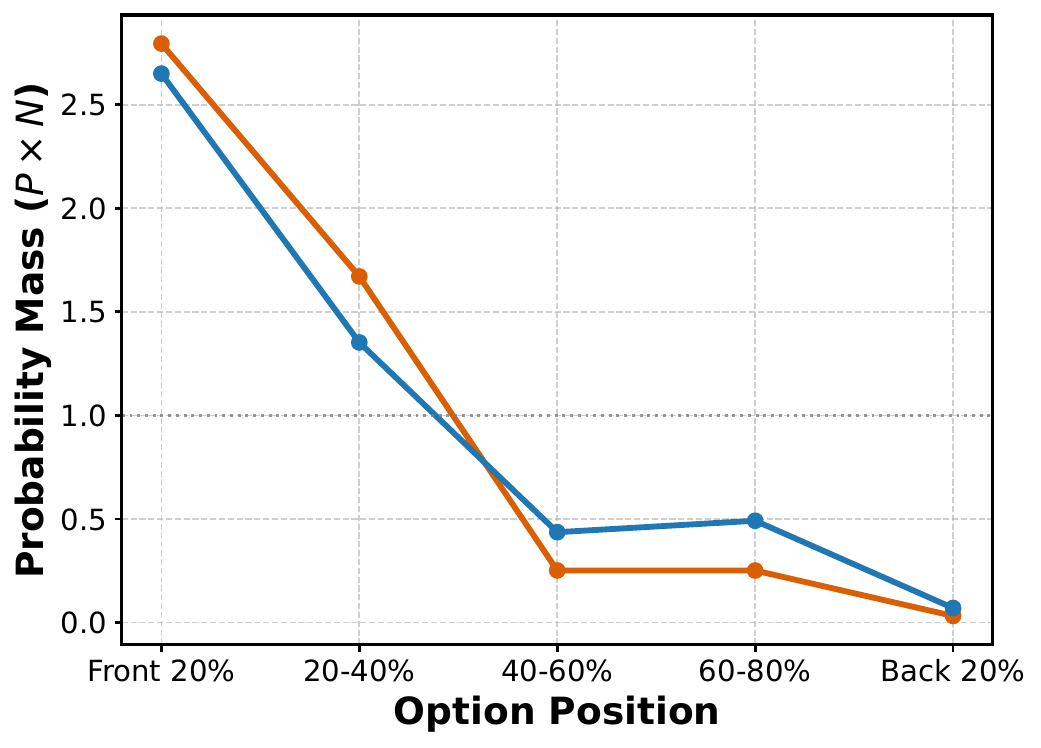}
        \caption{$N=160$}
    \end{subfigure}

    \vspace{0.2em}

    {\small \textbf{Llama-3.1-8B}}

    \vspace{0.4em}

    \includegraphics[width=0.35\linewidth]
    {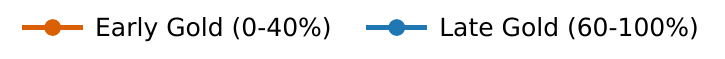}

    \caption{Controlled gold-position intervention on arXiv across models.
    Each row corresponds to a model, while columns vary the candidate set size.
    As the number of options increases, predictions become progressively concentrated toward earlier candidate positions regardless of the true gold location.}
    \label{fig:appendix_arxiv_gold_insert}
\end{figure*}

Figure~\ref{fig:appendix_arxiv_gold_insert} shows controlled gold-insertion results across models. For each model, we keep the distractor set fixed and vary only the insertion position of the gold answer. Across models, predictions become progressively less sensitive to the gold insertion position as the number of options increases.

\begin{figure*}[t]
    \centering

    \begin{subfigure}[t]{0.32\linewidth}
        \centering
        \includegraphics[width=\linewidth]{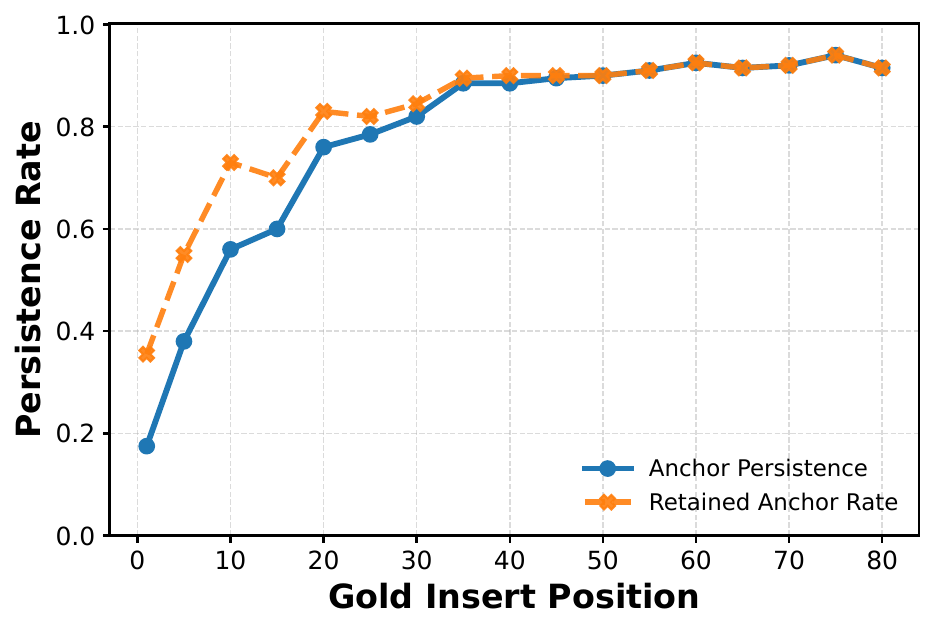}
        \caption{Phi-4-mini}
    \end{subfigure}
    \hfill
    \begin{subfigure}[t]{0.32\linewidth}
        \centering
        \includegraphics[width=\linewidth]{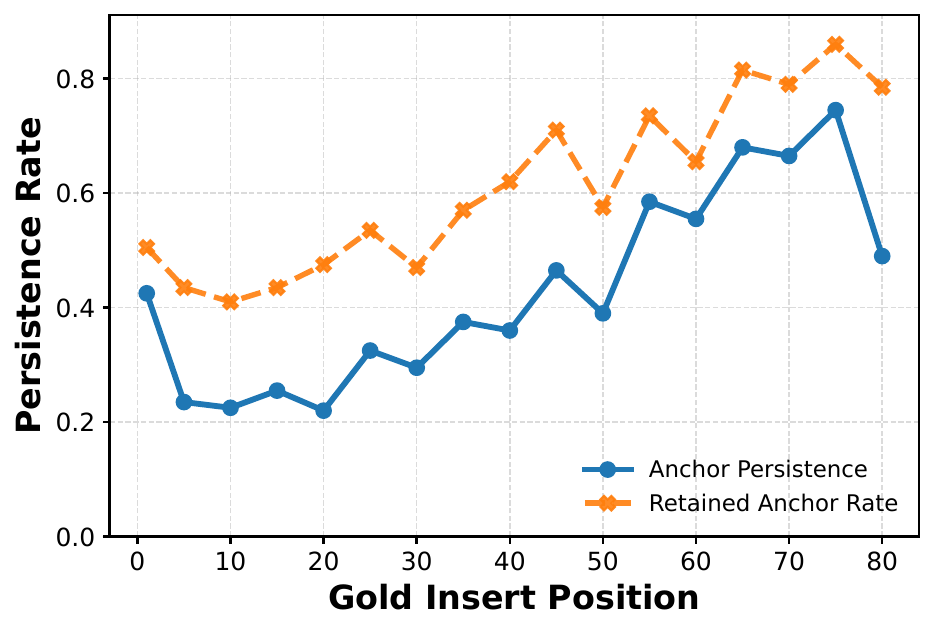}
        \caption{Llama-3.1-8B}
    \end{subfigure}
    \hfill
    \begin{subfigure}[t]{0.32\linewidth}
        \centering
        \includegraphics[width=\linewidth]{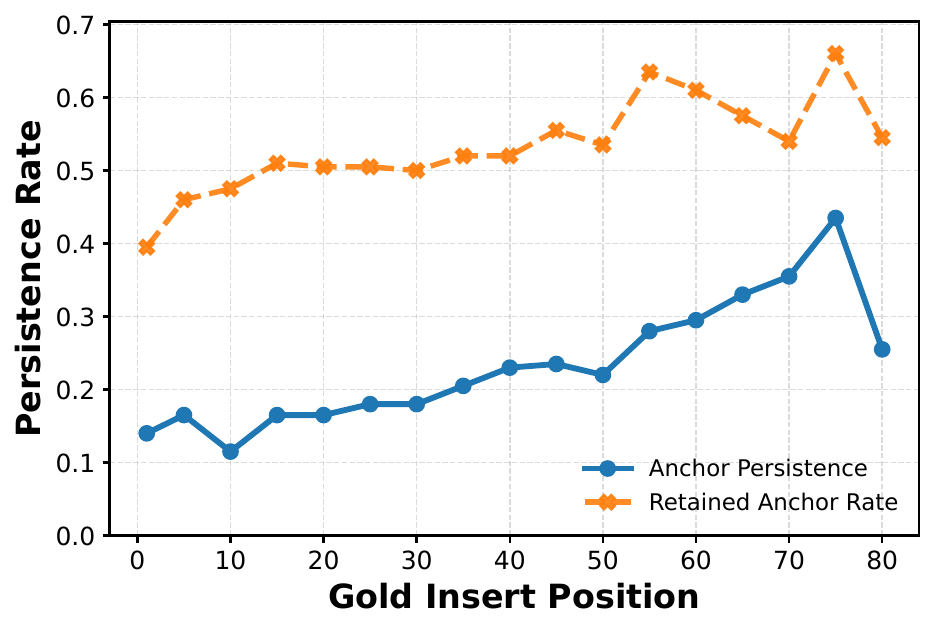}
        \caption{Qwen3-30B}
    \end{subfigure}

    \caption{Anchor persistence on arXiv across models.
    Later-positioned gold answers become progressively less effective at overturning earlier candidate preferences.}
    \label{fig:appendix_arxiv_persistence}
\end{figure*}

Figure~\ref{fig:appendix_arxiv_persistence} shows anchor persistence across models under large candidate sets. Across all models, later-positioned gold answers become progressively less effective at overturning earlier candidate preferences.

\begin{figure*}[t]
    \centering

    \begin{subfigure}[t]{0.32\linewidth}
        \centering
        \includegraphics[width=\linewidth]{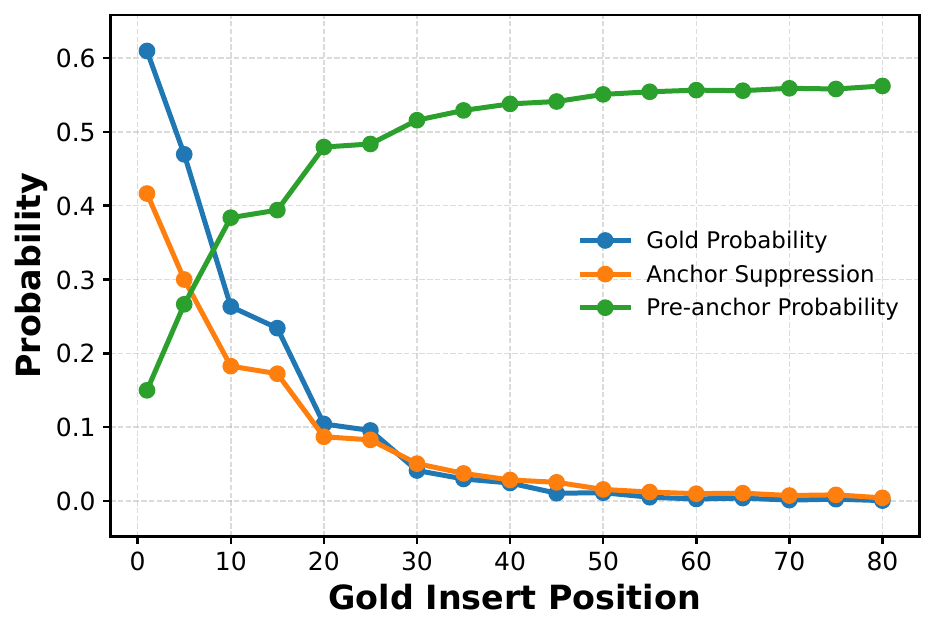}
        \caption{Phi-4-mini}
    \end{subfigure}
    \hfill
    \begin{subfigure}[t]{0.32\linewidth}
        \centering
        \includegraphics[width=\linewidth]{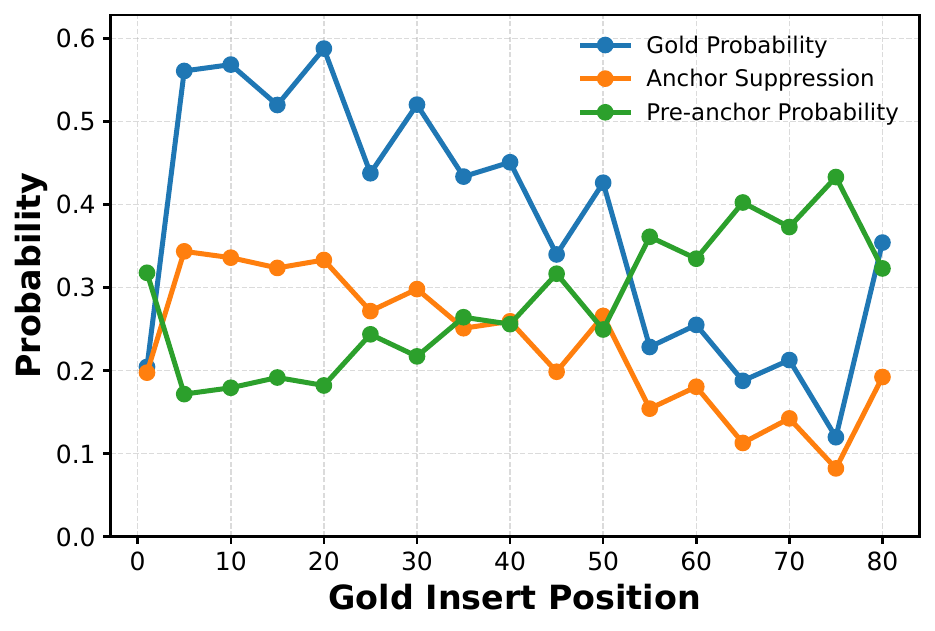}
        \caption{Llama-3.1-8B}
    \end{subfigure}
    \hfill
    \begin{subfigure}[t]{0.32\linewidth}
        \centering
        \includegraphics[width=\linewidth]{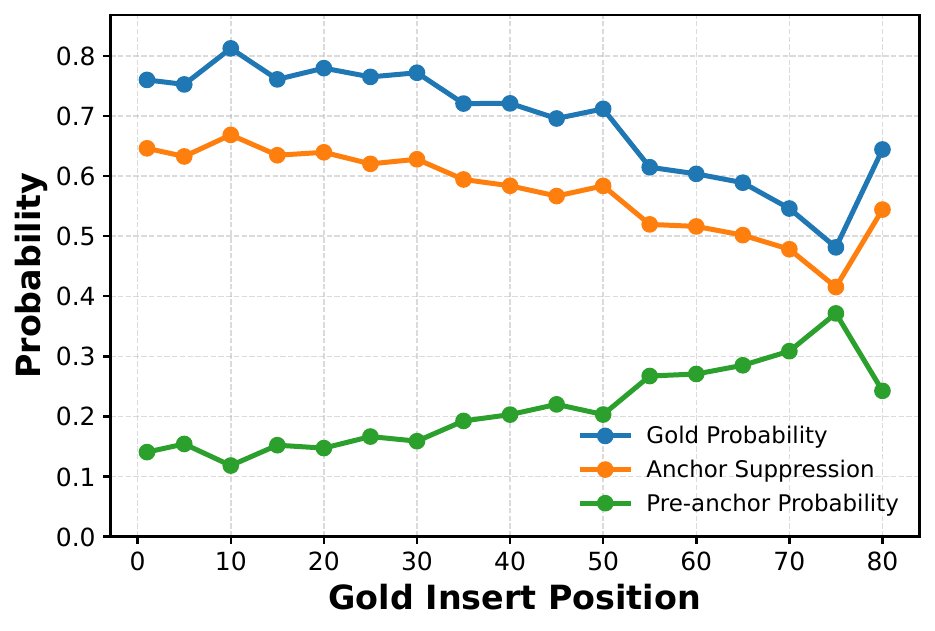}
        \caption{Qwen3-30B}
    \end{subfigure}

    \caption{Probability revision dynamics on arXiv across models.
    Later-arriving gold answers progressively lose the ability to overturn earlier candidate preferences.}
    \label{fig:appendix_arxiv_prob_dynamics}
\end{figure*}

Figure~\ref{fig:appendix_arxiv_prob_dynamics} shows probability revision dynamics across models. The probability assigned to the inserted gold generally decreases as the insertion position moves later, while the original anchor retains larger probability mass. Stronger models may exhibit a mild rebound at the final insertion positions, but the overall trend still indicates decreasing revision ability for later-arriving evidence.

\subsection{Additional Revision Dynamics Examples}
\label{sec:appendix-additional-sequential-anchor}

\begin{figure*}[t]
    \centering

    \begin{subfigure}[t]{0.45\linewidth}
        \centering
        \includegraphics[width=\linewidth]
        {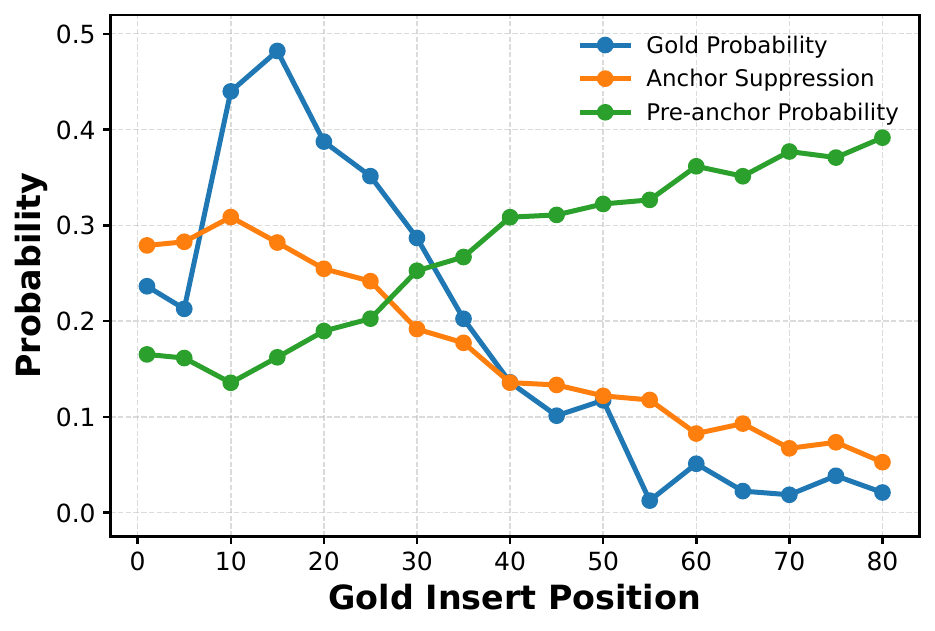}
        \caption{Llama-3.2-3B}
    \end{subfigure}
    \hfill
    \begin{subfigure}[t]{0.45\linewidth}
        \centering
        \includegraphics[width=\linewidth]
        {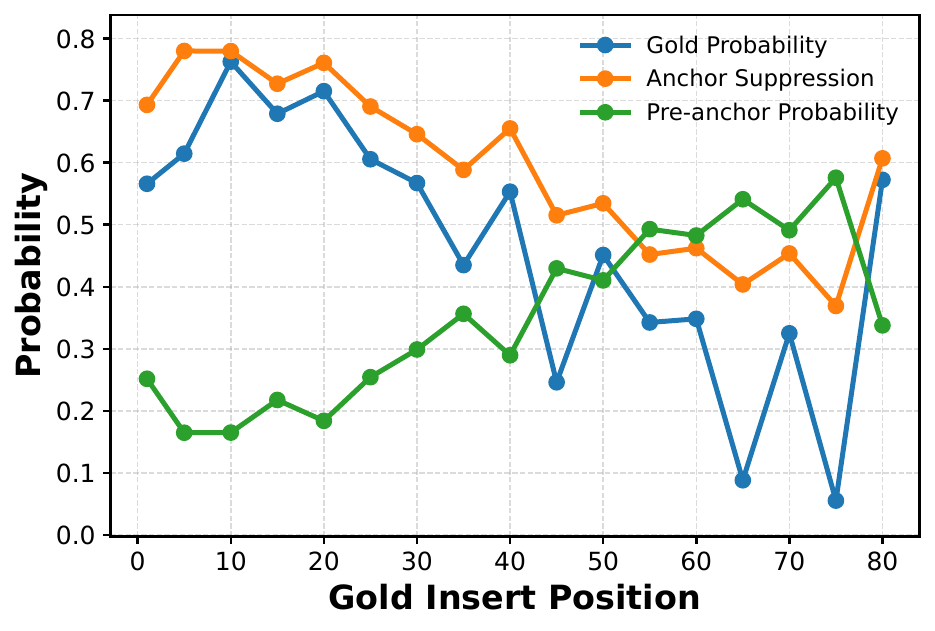}
        \caption{Qwen3-4B}
    \end{subfigure}

    \caption{Additional sequential revision dynamics across models.
    Across additional model families, later-arriving gold answers progressively lose the ability to revise earlier candidates under large-option comparison.}
    \label{fig:appendix_additional_prob_dynamics}
\end{figure*}

We observe similar revision dynamics on ArXiv dataset across additional model families and scales. 
Figure~\ref{fig:appendix_additional_prob_dynamics} provides additional examples using Llama-3.2-3B and Qwen3-4B. 
Despite differences in model scale and architecture, both models exhibit the same overall trend: later-position gold answers progressively lose the ability to overturn earlier candidate preferences. 
Interestingly, stronger models occasionally exhibit mild probability recovery near the final insertion positions, suggesting limited late-stage corrective ability despite strong early-anchor dominance.

\subsection{Additional Implementation Details}
\label{sec:appendix-addn-impl-details}

\paragraph{Prompting Strategies.}
We evaluate three inference strategies that differ in how the model is asked to evaluate the candidate set. These strategies allow us to  
(1) benchmark the LLM's ability in the MCQ reasoning tasks and 
(2) attempt to mitigate the accuracy decline at large $N$. 

\textit{Single-Pass Choice Inference.}
Our first strategy is a direct inference baseline in which the model receives the full task context together with all \(N\) candidate options in a single prompt, and is asked to select exactly one answer. This setting evaluates the model's ability to discriminate the correct option from all distractors under a flat large-option comparison structure, and serves as the standard reference baseline throughout our experiments.

\textit{Hierarchical Candidate Partitioning.}
Our second strategy decomposes the large candidate set into a sequence of smaller group-based comparisons. At each round, we partition the current candidate set into groups of size \(K\), ask the model to select one candidate from each group, and carry the selected candidates into the next round. This process is repeated until only one candidate remains. Formally, at round \(t\), we partition the active candidate set \(\mathcal{O}^{(t)}\) into groups \(\{\mathcal{O}^{(t)}_1,\dots,\mathcal{O}^{(t)}_{G_t}\}\), select
\[
\hat{o}^{(t)}_g = \arg\max_{o \in \mathcal{O}^{(t)}_g} p(o),
\]
and construct the next candidate set as
\[
\mathcal{O}^{(t+1)} = \{\hat{o}^{(t)}_1,\hat{o}^{(t)}_2,\dots,\hat{o}^{(t)}_{G_t}\}.
\]
The final prediction is the remaining candidate after repeated filtering. This hierarchical construction reduces the effective comparison set at each decision stage and tests whether large-option decision reliability improves when flat large-scale comparison is decomposed into smaller local comparisons.

\textit{Permutation-Based Candidate Aggregation.}
Our third strategy reduces sensitivity to candidate ordering by evaluating multiple shuffled views of the same candidate set. For each shuffled view \(v\), we randomly permute the candidate order and ask the model to select one candidate:
\[
\hat{o}_v = \arg\max_{o \in \pi_v(\mathcal{O})} p(o),
\]
where \(\pi_v(\mathcal{O})\) denotes the \(v\)-th shuffled candidate ordering. We then collect the selected candidates from all shuffled views into a reduced candidate set
\[
\mathcal{C} = \{\hat{o}_1,\hat{o}_2,\dots,\hat{o}_V\},
\]
and perform one final inference step over \(\mathcal{C}\):
\[
\hat{o} = \arg\max_{o \in \mathcal{C}} p(o).
\]
This strategy preserves candidates selected under different candidate orderings and reduces the influence of any single ordering on the final decision.

\subsection{Mechanism Analysis of Intervention}
\label{sec:appendix_intervention_mechanism}

We further analyze how hierarchical partitioning and permutation-based inference 
improve decision quality under large candidate sets.

While hierarchical partitioning addresses the limitations in scalable candidate comparison by reducing the effective decision complexity, permutation-based inference mitigates positional bias 
by reducing sensitivity to candidate ordering.

Both methods aim to alleviate the comparison bottleneck identified in our analysis, but operate through distinct mechanisms.

\paragraph{Hierarchical Partitioning.}

Hierarchical partitioning decomposes a large candidate set into smaller subsets and performs local selection within each group before a final comparison stage. This reduces comparison complexity at each step, enabling more reliable local decisions under large candidate spaces.

\begin{figure*}[t]
    \centering

    \begin{subfigure}[t]{0.46\linewidth}
        \centering
        \includegraphics[width=\linewidth]
        {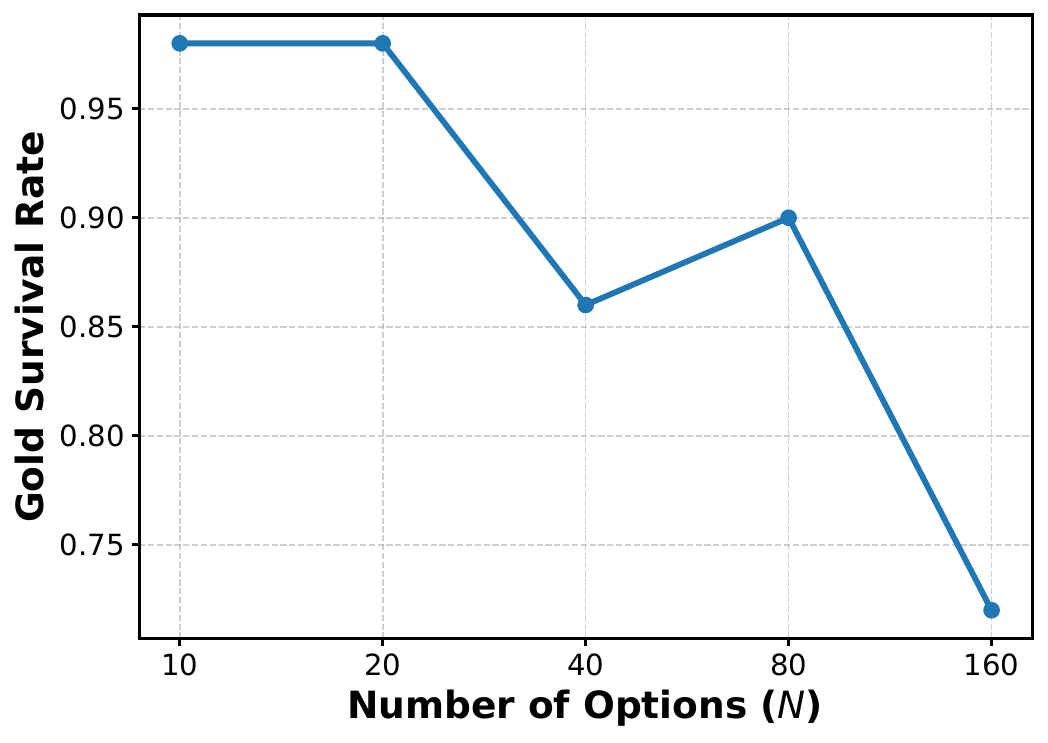}
        \caption{Hierarchical partitioning}
    \end{subfigure}
    \hfill
    \begin{subfigure}[t]{0.46\linewidth}
        \centering
        \includegraphics[width=\linewidth]
        {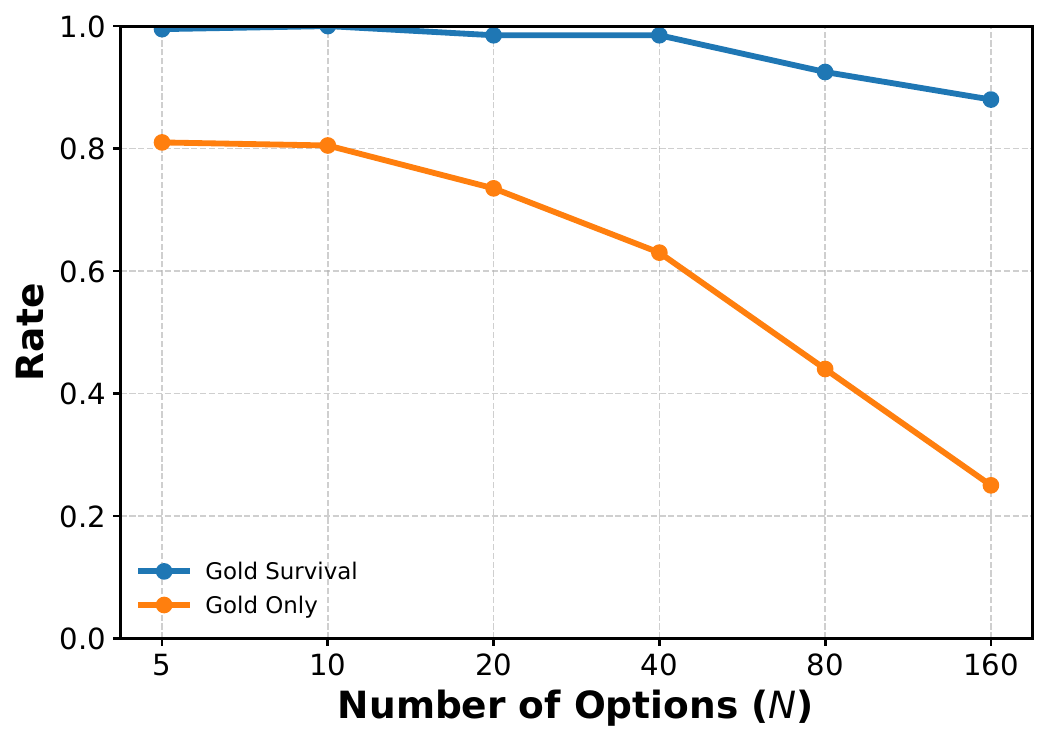}
        \caption{Permutation-based inference}
    \end{subfigure}

    \caption{Mechanisms of inference-time restructuring.
    (a) Hierarchical partitioning reduces comparison difficulty through local candidate selection, but introduces irreversible early elimination errors as the number of options increases.
    (b) Permutation-based inference mitigates positional bias by aggregating predictions across reordered candidate views, improving robustness during final-stage comparison among surviving candidates.}
    \label{fig:inference_mechanisms}
\end{figure*}

As shown in Figure~\ref{fig:inference_mechanisms}(a), this design introduces a key trade-off: the probability that the gold candidate survives to the final round decreases as the number of options increases. This suggests that hierarchical partitioning shifts the dominant failure mode from \textbf{unreliable large-scale comparison} to \textbf{irreversible early elimination}, where mistakes made during local selection cannot be corrected in later stages.

\paragraph{Permutation-Based Inference.}

While hierarchical partitioning reduces comparison complexity, permutation-based inference instead focuses on mitigating positional bias by reducing sensitivity to candidate ordering.

Permutation-based inference aggregates predictions across multiple reordered views of the same candidate set. To better understand its behavior, we decompose the final accuracy as:
\[
\mathrm{Acc} = P_{\text{only}} + P_{\text{multi}} \cdot \mathrm{Acc}_{\text{final}},
\]
where $P_{\text{only}}$ denotes cases where the gold candidate survives aggregation and is the \textbf{only remaining candidate}, while $P_{\text{multi}}$ corresponds to cases where the gold candidate survives but multiple candidates remain, requiring a non-trivial final comparison.

As shown in Figure~\ref{fig:inference_mechanisms}(b), the gap between gold survival and gold-only selection widens as the number of options increases. Although the gold candidate is increasingly preserved after aggregation, it more frequently coexists with competing candidates, shifting the task from \textbf{unambiguous selection} to \textbf{comparison-driven decision making}. In this regime, final-stage comparison quality becomes the dominant factor determining overall accuracy. Permutation-based inference improves robustness in this setting by reducing sensitivity to sequential candidate ordering.

To assess whether these improvements are sensitive to the sampled
permutations, we additionally repeat permutation-based inference with
five random seeds (0--4) on HotpotQA using Llama-3.1-8B.
Across candidate-set sizes, the standard deviation in accuracy ranges
from only 0.67 to 1.33 percentage points. At $N=160$,
permutation-based inference achieves $54.38 \pm 1.00$\% accuracy
across seeds. This low run-to-run variance suggests that the observed
improvement is not driven by a particular choice of candidate permutations.

\paragraph{Accuracy--Efficiency Trade-off.}

The improved robustness of inference-time restructuring comes with
additional computational cost. We therefore compare the accuracy and inference efficiency of
Single-Pass, Hierarchical Partitioning, and Permutation-Based Inference
under the same HotpotQA, Llama-3.1-8B, $N=160$ evaluation setting.

As shown in Table~\ref{tab:intervention_efficiency}, Single-Pass inference
requires only one model call per example and achieves 57.5\% accuracy.
Permutation-Based Inference improves accuracy to 76.5\%, a gain of
19.0 percentage points, while requiring an average of 5.76 model calls
and 8.74 seconds per example. Hierarchical Partitioning achieves 71.0\%
accuracy, a gain of 13.5 points, while requiring substantially more model
calls (42.0 per example) but only 6.04 seconds of wall-clock latency under
our implementation.

These results highlight an accuracy--efficiency trade-off: both
interventions substantially improve large-option decision quality, but
require additional inference computation. Permutation-Based Inference
provides the larger accuracy improvement with substantially fewer model
calls than Hierarchical Partitioning. Notably, model-call count does not
translate directly into wall-clock latency in our implementation. We use
vLLM with batched inference, which allows multiple independent local
comparisons to be processed efficiently within the same inference
workflow. Consequently, despite requiring more model calls, Hierarchical
Partitioning achieves lower measured wall-clock latency than
Permutation-Based Inference in our experimental setup.

\begin{table}[t]
\caption{
\textbf{Accuracy--efficiency trade-off of inference-time interventions.}
Latency reports measured wall-clock time per example under our
vLLM-based batched inference implementation; model-call counts therefore
do not translate directly into wall-clock latency.
}
\label{tab:intervention_efficiency}
\centering
\small
\setlength{\tabcolsep}{4pt}
\begin{tabular}{lccc}
\toprule
Method & Acc. (\%) & Calls / Sample & Latency (s) \\
\midrule
Single-Pass   & 57.5 & 1.00 & 2.46 \\
Permutation   & 76.5 & 5.76 & 8.74 \\
Hierarchy     & 71.0 & 42.0 & 6.04 \\
\bottomrule
\end{tabular}
\end{table}

\paragraph{Unified Interpretation.}

Taken together, these results reveal that the two methods address distinct but complementary challenges.

Hierarchical partitioning improves decision reliability by reducing comparison complexity,
while permutation-based inference mitigates positional bias by aggregating across different orderings.

Together, they improve decision quality by addressing both comparison difficulty 
and ordering sensitivity.

\subsection{Additional Experimental Results}
\paragraph{Empirical Validation Across Models.}

To further validate the robustness of hierarchical partitioning and permutation-based inference,
we evaluate both methods across different model scales.

\begin{figure*}[t]
    \centering

    \begin{subfigure}[t]{0.42\linewidth}
        \centering
        \includegraphics[width=\linewidth]
        {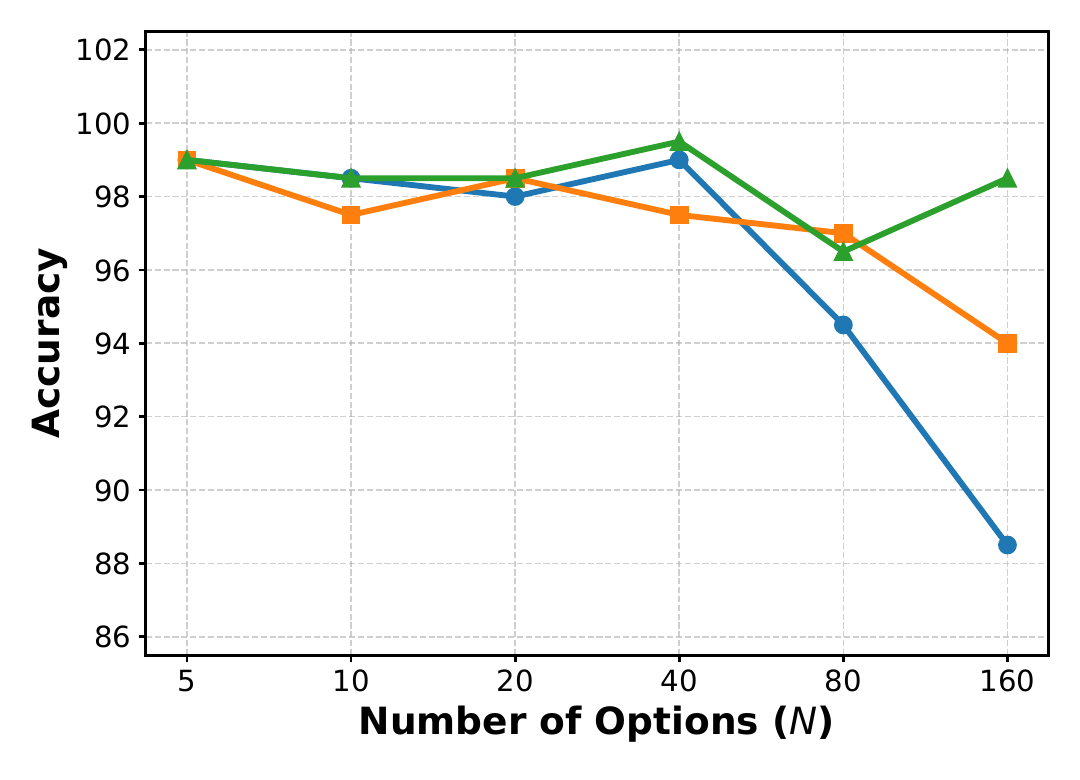}
        \caption{Qwen3-30B}
    \end{subfigure}
    \hfill
    \begin{subfigure}[t]{0.42\linewidth}
        \centering
        \includegraphics[width=\linewidth]
        {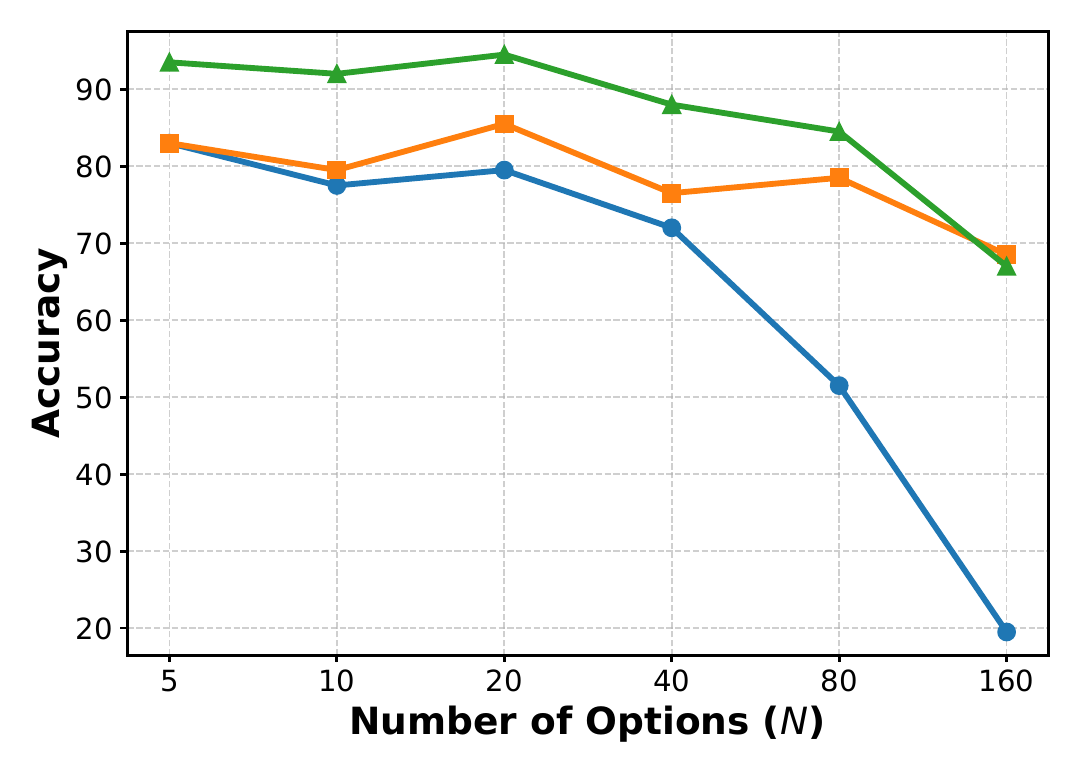}
        \caption{Phi-4-mini}
    \end{subfigure}

    \vspace{0.4em}

    \includegraphics[width=0.45\linewidth]
    {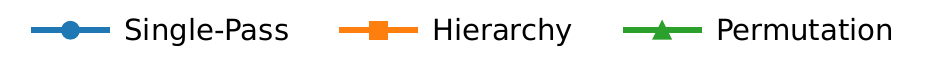}

    \caption{Generalization across model scales.
    Hierarchical grouping and permutation-based inference consistently mitigate large-option degradation across both small and large models.}
    \label{fig:multi_model}
\end{figure*}

As shown in Figure~\ref{fig:multi_model}, both hierarchical grouping and permutation-based inference 
consistently mitigate accuracy degradation as the number of options increases.
This trend holds across models of different capacities, including Qwen3-30B and Phi-4-mini,
indicating that the effectiveness of these strategies is not tied to model scale.

While single-pass inference suffers from severe performance drops at large $N$,
both methods improve robustness by either reducing comparison complexity 
(hierarchical grouping) or alleviating ordering sensitivity (permutation-based inference).

\paragraph{Effect of Group Size and Number of Permutations.}

We further analyze how hyperparameters influence the effectiveness of both methods,
focusing on group size for hierarchical partitioning and the number of shuffled views 
for permutation-based inference.

\begin{figure*}[t]
    \centering

    \begin{subfigure}[t]{0.42\linewidth}
        \centering
        \includegraphics[width=\linewidth]
        {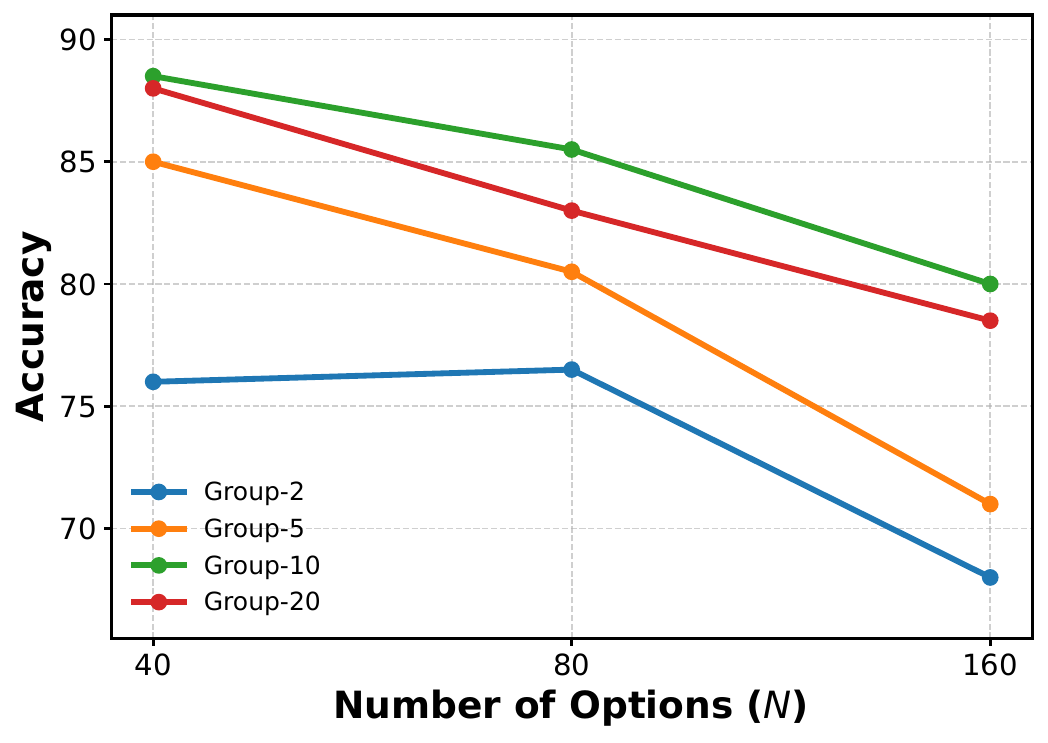}
        \caption{Effect of group size}
    \end{subfigure}
    \hfill
    \begin{subfigure}[t]{0.42\linewidth}
        \centering
        \includegraphics[width=\linewidth]
        {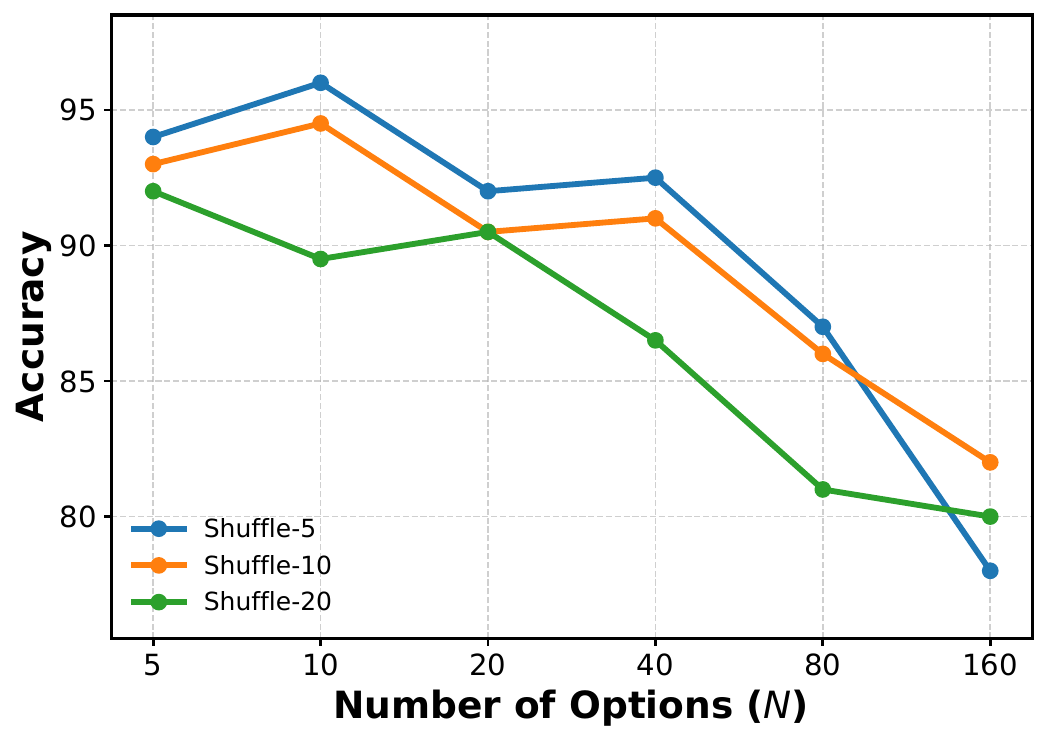}
        \caption{Effect of number of permutations}
    \end{subfigure}

    \caption{Effect of hyperparameters.
    Larger group sizes improve hierarchical partitioning by preserving stronger candidates during local selection, while permutation-based inference benefits from a moderate number of shuffled views but exhibits diminishing returns at larger scales.}
    \label{fig:hyperparam_effect}
\end{figure*}

As shown in Figure~\ref{fig:hyperparam_effect}(a), larger group sizes generally lead to better performance
for hierarchical partitioning across all values of $N$.
In particular, moderate-to-large group sizes (e.g., Group-10 and Group-20) consistently outperform smaller groups,
suggesting that excessively small groups may suffer from insufficient candidate diversity in early stages,
leading to suboptimal local selections.

However, all configurations still exhibit performance degradation as the number of options increases,
indicating that hierarchical partitioning cannot fully eliminate the comparison bottleneck,
but only mitigates it.

Figure~\ref{fig:hyperparam_effect}(b) shows the effect of the number of permutations.
Increasing the number of shuffled views improves robustness at moderate values,
but does not lead to consistent gains at larger $N$.
In fact, excessive permutations can introduce noise in aggregation,
leading to slight performance drops in some cases.

These results highlight that both methods involve trade-offs:
hierarchical partitioning benefits from sufficiently large group sizes to preserve strong candidates,
while permutation-based inference achieves the best performance with a moderate number of diversified views.

\end{document}